\documentclass[preprint,12pt,authoryear]{elsarticle}
\usepackage{color,soul}
\usepackage{booktabs}
\usepackage{amsmath}
\usepackage{multirow}
\usepackage{makecell}
\usepackage{rotating}
\usepackage{xcolor}
\usepackage[capposition=top]{floatrow}
\usepackage{amsmath,amsfonts}
\definecolor{cerulean}{rgb}{0.0,0.48,0.65}
\definecolor{green}{rgb}{0.01, 0.75, 0.24}
\definecolor{Black}{RGB}{0.0, 0.0, 0.0}

\newcommand{\blue}[1]{\textcolor{Black}{#1}}

\newcommand{\shadow}[1]{}

\def\b{\blue}

\def\s{\shadow}

\newcounter{daggerfootnote}
\newcommand*{\daggerfootnote}[1]{%
    \setcounter{daggerfootnote}{\value{footnote}}%
    \renewcommand*{\thefootnote}{\fnsymbol{footnote}}%
    \footnote[2]{#1}%
    \setcounter{footnote}{\value{daggerfootnote}}%
    \renewcommand*{\thefootnote}{\arabic{footnote}}%
    }

\usepackage{amssymb}

\journal{Neural Networks}

\begin{document}

\begin{frontmatter}



\title{Self-Supervised Anomaly Detection:\\ A Survey and Outlook}


\author[inst1,inst2]{Hadi Hojjati\protect\daggerfootnote{Both authors contributed equally to the paper}}
\author[inst1,inst2]{Thi Kieu Khanh Ho$^{\dag}$}
\author[inst1,inst2]{Naregs Armanfard}

\affiliation[inst1]{organization={Department of Electrical and Computer Engineering},
            addressline={McGill University}, 
            city={Montreal},
            state={QC},
            country={Canada}}

\affiliation[inst2]{organization={Mila - Quebec AI Institute},
            city={Montreal}, 
            state={QC},
            country={Canada}}

\begin{abstract}
Anomaly detection (AD) plays a crucial role in various domains, including cybersecurity, finance, and healthcare, by identifying patterns or events that deviate from normal behaviour. In recent years, significant progress has been made in this field due to the remarkable growth of deep learning models. Notably, the advent of self-supervised learning has sparked the development of novel AD algorithms that outperform the existing state-of-the-art approaches by a considerable margin. This paper aims to provide a comprehensive review of the current methodologies in self-supervised anomaly detection. We present technical details of the standard methods and discuss their strengths and drawbacks. We also compare the performance of these models against each other and other state-of-the-art anomaly detection models. Finally, the paper concludes with a discussion of future directions for self-supervised anomaly detection, including the development of more effective and efficient algorithms and the integration of these techniques with other related fields, such as multi-modal learning.
\end{abstract}




\end{frontmatter}


\section{Introduction}
\label{sec:intro}


Anomaly detection (AD) is the task of identifying samples that differ significantly from the majority of data and often signals an irregular, fake, rare, or fraudulent observation \citep{wang2019progress}. Anomaly detection is particularly useful in cases where we cannot define all existing classes during training. This makes AD algorithms applicable to a broad range of applications, including but not limited to intrusion detection in cybersecurity \citep{xin2018machine}, fraud detection \citep{malaiya2018empirical}, acoustic novelty detection \citep{AADCL}, and medical diagnosis \citep{latif2018phonocardiographic}.


In the past, anomaly detection relied on manual inspection of data by experts. However, with the proliferation of sensory systems, the volume of data has surged, making the traditional method impractical. As a result, automatic anomaly detection methods, including machine learning (ML)-based techniques, have gained significant popularity. Over the past few decades, numerous ML-based models have been developed for this purpose. Classical approaches like Kernel Density Estimation (KDE), One-Class Support Vector Machine (OCSVM), and Isolation Forests (IF) have been widely adopted. However, the performance of these algorithms often degrades when applied to higher-dimensional data.
In recent years, deep learning models have shown significant improvements over traditional ML models since they have the capability to learn intricate patterns and representations from vast amounts of data, making them well-suited for anomaly detection. The utilization of deep learning for anomaly detection has yielded high accuracy and robust results, establishing it as a popular choice in various applications \citep{ruff2021unifying, hojjati2021dasvdd}.

\s{
Deep-learning based models for anomaly detection can be broadly classified into three categories: The first category comprises models that utilize deep neural networks to learn a lower-dimensional representation of high-dimensional data. Subsequently, they apply a classical anomaly detection algorithm, such as One-Class Support Vector Machine (OCSVM) \citep{ocsvm}, to the obtained lower-dimensional representation \citep{sabokrou2017deepanomaly}. By mapping the data into a lower-dimensional space, these approaches mitigate the curse of dimensionality issues associated with traditional non-deep learning anomaly detection methods, thereby yielding reasonably accurate detection results. The second group of methods involves using deep neural networks to reconstruct the input data and calculate an anomaly score directly based on the data reconstruction loss. The prevalent network architectures employed in these methods are Autoencoders (AEs) \citep{ae2} and Generative Adversarial Networks (GANs) \citep{schlegl2017unsupervised,schlegl2019f}. The underlying assumption is that a network trained solely on reconstructing normal data will produce a significant reconstruction error when confronted with an anomaly. The third category encompasses algorithms combining both approaches \citep{hojjati2021dasvdd,ruff2018deep}. These methods jointly train a neural network for feature extraction and an anomaly detector on the latent space of the network. The anomaly detector assigns an anomaly score to input data based on the learned representations. By combining feature extraction and anomaly detection in a unified framework, these models aim to enhance detection performance.
Although the above methods use different approaches for AD, the concept remains the same, i.e. normal samples have similar feature distribution in the latent space of the trained network, and abnormal instances are not in line with the ordinary anticipated behaviour of normality\footnote{In the rest of the paper, the term \textit{normal} has no relationship with the normal Gaussian distribution unless specified otherwise.}.
}


Compared to typical deep learning tasks, anomaly detection poses unique challenges due to the characteristics of the data involved. Anomalies are typically rare occurrences or costly events in the real world. Consequently, the training data for anomaly detection is imbalanced, with a majority of normal data and only a small number of anomalies. Moreover, these anomalous samples can be contaminated with noise, further complicating the detection task. Additionally, anomalies cannot be treated as a single class, and a detection system may encounter new types of abnormalities that were not present in the training data. These challenges render a significant portion of deep learning algorithms ineffective for anomaly detection.

In general, deep learning models can be categorized into supervised, semi-supervised, and unsupervised methods. Supervised methods, which rely on labelled data, often achieve high performance. However, as previously mentioned, annotated data is not commonly available for anomaly detection tasks, making semi-supervised and unsupervised models the only practical options. Unfortunately, these algorithms generally do not perform as well as their supervised counterparts. This limitation acts as a significant bottleneck, preventing deep anomaly detection algorithms from surpassing a certain performance threshold.


Recently, there has been a resurgence of hope for anomaly detection algorithms with the emergence of self-supervised learning (SSL).
In SSL, similar to unsupervised learning, the model learns from unlabelled data without external annotation. It learns a generalizable representation from data by solving a supervised proxy task which is often unrelated to the target task but can help the network to learn a better embedding space. Depending on the nature of the data, a diverse set of tasks, such as colorization \citep{larsson2016learning}, mutual information maximization \citep{hjelm2018learning}, and predicting geometric transformations \citep{gidaris2018unsupervised} can be used as the supervised proxy task. These methods showed promising results in various applications, such as speech representation learning \citep{speechSSL}, visual feature learning \citep{VisualSSL}, and healthcare applications \citep{healthSSL}. Even in some cases,  self-supervised algorithms have approached the performance of fully-supervised models \citep{chen2020simple}. 
\s{Additionally, self-supervised models can learn representations that capture complex patterns and relationships in the data, making them effective at detecting subtle anomalies that other methods might miss.}

Motivated by the recent success of SSL, anomaly detection researchers have started to incorporate the idea of self-supervision for developing effective algorithms. Their studies showed that the representation that is learned through self-supervision could be useful for anomaly detection if the anomaly score and the pretext task are defined appropriately \citep{tack2020csi, reiss2021mean}.


As a result, self-supervised algorithms have emerged as the new state-of-the-art in anomaly detection, outperforming other traditional methods. Recently, a wide range of SSL frameworks has been developed for anomaly detection. However, to the best of our knowledge, no paper conducted a comprehensive review of these methods. We aim to fill this gap by thoroughly reviewing and categorizing self-supervised learning approaches in anomaly detection. Our work provides a valuable resource for researchers and practitioners in this field and contributes to advancing state-of-the-art anomaly detection. In short, we can summarize the contribution of our work as follows:


\begin{itemize}
\item We briefly review the current approaches in anomaly detection to locate self-supervised anomaly detection in the context of AD research.
    \item We discuss the current approaches in self-supervised anomaly detection and its application areas.
    \item We divide the existing self-supervised anomaly detection algorithms into two high-level categories based on their requirement of negative samples during training. SSL-based models are different from each other based on their proxy tasks and architecture. Hence, it is essential to categorize these methods to cover all of them. 
    \item \b{For the first time, we extensively cover the self-supervised learning algorithms based on the data type that they are dealing with.}
    \item For each type of method, we describe the techniques and assumptions and highlight their pros and cons. We also discuss the implementation details of some prominent algorithms in each category.
    \item We discuss possible future directions in self-supervised anomaly detection research.
\end{itemize}

\section{Related Works}
\label{sec:section2}
In recent decades, there has been significant research and exploration of the anomaly detection problem across various domains.
Several survey articles attempted to group anomaly detection algorithms into distinctive categories. \citet{surveyHodge} and \citet{surveyagyemang} are two examples of early studies that categorized the existing algorithms and extensively discussed the techniques that are used in each category. 
In another prominent work, \citet{chandola2007outlier} surveyed the existing anomaly detection algorithms and divided them into distinctive categories. In addition to describing the technical details of each method, they identified the underlying assumptions that are implicitly made regarding the anomalies. They also discussed the advantages, disadvantages and computational complexity of each technique. Furthermore, they extensively reviewed the application areas of the methods and highlighted the challenges faced in each domain.

More recently, deep learning methods have inspired researchers in anomaly detection, leading to the development of new algorithms in this domain. As a result, review papers focusing on deep anomaly detection have emerged. \citet{SurveyChalapathy} was one of the first papers that presented a comprehensive review of deep anomaly detection methods. They categorized the existing algorithms based on their underlying assumptions and explained the pros and cons of each approach. \citet{SurveyChalapathy} have also thoroughly explored applications of deep anomaly detection and assessed the effectiveness of each method.
In another similar survey, \citet{SurveyPang} reviewed contemporary deep AD methods. They first discussed the challenges and complexities that anomaly detection faces, and then they categorized the existing deep methods into three high-level categories and eleven fine-grained subcategories. They emphasized how each category addresses challenges and identified key assumptions and intuitions. Notably, they also compiled a list of publicly available codes and datasets for benchmarking.
While most review papers in recent years focused on specific sets of algorithms, \citet{ruff2021unifying} presented an extensive survey of anomaly detection methods, unifying classic shallow methods with recent deep approaches. They highlighted connections and similarities between these two types of algorithms, providing an in-depth description and taxonomy of common practices and challenges in anomaly detection.
In addition to the mentioned studies, several other review papers in this field have been published, focusing on specific domains of application or particular types of methods. For example, the two survey papers \citet{GanSurveyDimattia} and \citet{GanSurveyXia} are dedicated to reviewing the GAN-based anomaly detection methods. They discussed these models' theoretical bases and practical applications and provided a detailed description of existing challenges and future directions in GAN-based anomaly detection. Both of the papers also carried out empirical evaluations to compare the performance of different algorithms. In another study, \citet{SemisupervisedSurvey} empirically evaluated the performance of 29 semi-supervised AD algorithms.

While numerous survey papers have explored various aspects of anomaly detection, there remains a research gap concerning the thorough investigation of self-supervised methods, which have emerged as state-of-the-art in recent years. This paper aims to address this gap and provide a comprehensive analysis of self-supervised anomaly detection papers.

\section{Anomaly Detection: Terminology and Common Practices}

The term \textit{anomaly detection} is commonly used to encompass all algorithms designed to identify samples that deviate from normal patterns. Needless to say, the development of anomaly detection models depends on factors such as the availability of data labels, types of anomalies, and specific applications. Furthermore, there is inconsistency in the nomenclature used in the literature. To ensure clarity and avoid confusion, we first define and describe the relevant terminologies used throughout the paper.

\subsection{Anomaly, Outlier, Novelty, Out-of-Distribution Detection}

\begin{figure*}[!t]
\centering
\includegraphics[width=0.8\linewidth]{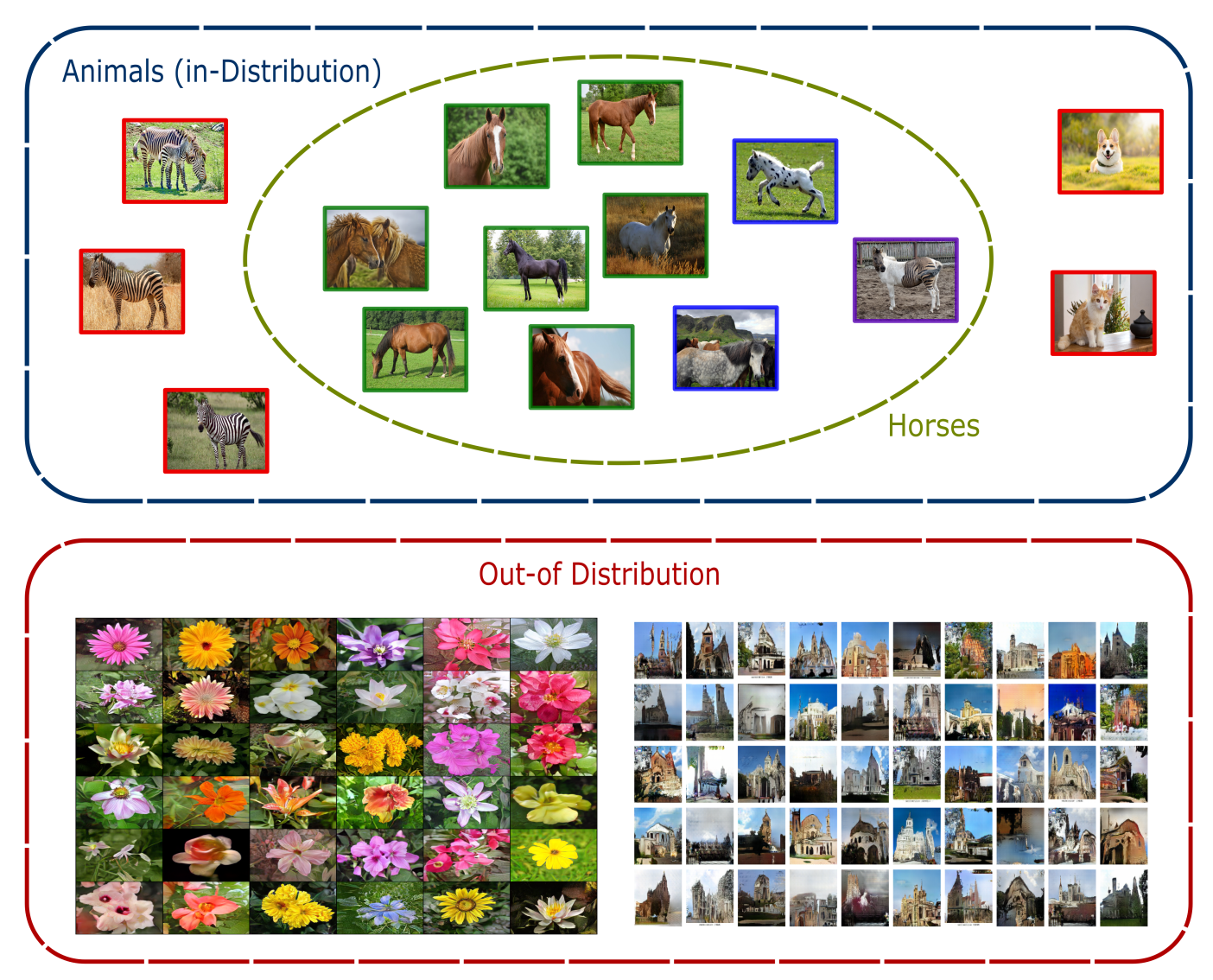}
\caption{Normal samples are shown in green, anomalies in red, outliers in blue and novelties in purple. The dataset of animals is denoted by a light-blue dashed box while a dashed dark-red box shows other out-of-distribution datasets.}
\label{ood}
\end{figure*}

Some studies use the terms \textit{anomaly}, \textit{novelty}, \textit{outlier} and \textit{out-of-distribution} interchangeably, while others distinguish them. Although most of the algorithms for detecting them are similar, their significance and application might differ. In this paper, we adopt the terminology proposed by previous studies and define each task as follows \citep{ruff2021unifying}:

\begin{itemize}

\item\textbf{Anomaly Detection:} \textit{Anomaly detection} can be defined as the task of identifying samples that are drawn from a distribution other than the distribution of normal instances, denoted as $\mathbb{P^+}$. For instance, if we consider $\mathbb{P^+}$ as the distribution of horses, a zebra would be considered an anomaly in the context of anomaly detection.

\item\textbf{Outlier Detection:} An \textit{outlier} is defined as a low-probability sample drawn from the distribution of normal instances, $\mathbb{P^+}$. For instance, in the context of horse detection, a Falabella (a small horse breed) would be considered an outlier among the various horse breeds.

\item\textbf{Novelty Detection} A \textit{novelty} is a sample that is drawn from a new region of a non-stationary distribution of normal samples $\mathbb{P^+}$. These samples are often encountered during the inference phase, but their counterparts were not present in the training data. For instance, a new breed of horses is considered a novelty in the horse detection task.

\item\textbf{Out-of-Distribution Detection} In \textit{out-of-distribution (OOD)} detection, the goal is to identify samples that do not belong to any of the training set classes. This problem, which is also referred to as \textit{open category detection} \citep{opencategory}, is often formulated as a supervised problem where we have the labelled data from $K$ classes during training. We treat all the $K$ classes as normal and aim to identify if a sample does not come from these classes during the inference phase. \s{Recent studies have shown that training a supervised classifier on $K$ classes and using Softmax probabilities for calculating the anomaly scores can yield state-of-the-art performance in the OOD detection task\citep{OSR}}An example of the OOD detection task is using a classifier trained on an animal dataset to detect samples from other datasets, e.g. flowers.

\end{itemize}
Figure~\ref{ood} illustrates an example of normal sample versus anomalies, outliers, novelty and out-of-distribution data.

\subsection{Types of Anomalies}
In the classic anomaly detection literature, anomalies are classified into three categories based on their nature \citep{chandola2007outlier,SurveyPang}:
\begin{itemize}

\item \textbf{Point Anomalies:} A \textit{point anomaly} refers to an individual sample that exhibits an irregularity or deviation from the standard pattern. A single cat image in the dataset of dog images or a fraudulent insurance claim are examples of point anomalies. Most studies in the anomaly detection literature focus on this type of anomaly \citep{SurveyChalapathy}.
    
\item \textbf{Contextual Anomalies:} A \textit{contextual anomaly}, also known as a \textit{conditional anomaly}, is a data point deemed abnormal within a specific context. The context should be defined as a part of the problem formulation. For instance, a value of $120$ $km/h$ is considered an abnormal recording of the speed of a bike, whereas it is not considered an abnormal recording of the speed of a car. The anomaly classification depends on the context in which the data point is evaluated.
    
\item \textbf{Collective Anomalies:} \textit{Collective anomalies}, also called \textit{group anomalies}, are a subset of data points that exhibit collective abnormality when considered in relation to the entire dataset. While each sample within a collective anomaly may not be abnormal, their combined presence indicates an anomaly. For instance, a series of high-value credit card transactions that occur rapidly and consecutively might suggest a stolen credit card, even though each individual transaction might appear normal. The collective behavior or pattern highlights the anomaly in this case.

\end{itemize}

With the emergence of deep anomaly detection methods, recent studies proposed two additional anomaly types to distinguish between the various types of anomalies that deep models aim to detect \citep{ruff2021unifying}:

\begin{itemize}

\item \textbf{Sensory (Low-Level) Anomalies:} \textit{Low-level} or \textit{sensory anomalies} refer to the irregularities that occur in the low-level feature hierarchy, such as textures or edges of an image. An example of a low-level anomaly is a fractured texture. Low-level anomaly detection is helpful in detecting defects and artifacts in industrial applications. The recently introduced \textit{MVTecAD} dataset \citep{MVTECAD} contains numerous examples of sensory anomalies and defects in industrial applications.
    
\item \textbf{Semantic (High-Level) Anomalies:} \textit{High-level} or \textit{semantic anomalies} refer to samples that belong to a different class compared to the normal data. For example, if we train a network to classify cat images as normal samples, any image of an object other than a cat would be considered a semantic anomaly. In this context, the anomaly is determined based on the semantic content or class of the sample rather than low-level features.

\end{itemize}
It is important to note that both sensory and semantic anomalies might overlap with other types of anomalies. However, it is still essential to distinguish between semantic and sensory anomalies to avoid confusion in our discussions throughout the paper.

\subsection{Availability of Data Labels}
To design an appropriate algorithm for anomaly detection, it is crucial to consider the availability of labels. Based on the label availability, AD algorithms can be divided into three settings:

\begin{enumerate}

\item \textbf{Unsupervised Anomaly Detection:} In this setting, which is arguably the most common in anomaly detection, we assume that only unlabeled data is available for training the model \citep{ruff2021unifying,surveyHodge}. In the simplified form of unsupervised learning, we commonly assume that the data is noise-free and its distribution is the same as the normal data, e.g. $\mathbb{P} \equiv \mathbb{P^+}$. If noisy data or undetected anomalies are present in the training dataset, these assumptions are violated, hence the developed models are not robust. A more realistic approach can be to assume that the data distribution $\mathbb{P}$ is a mixture of normal data and anomalies with a pollution rate $\eta \in (0,1)$, e.g. $\mathbb{P} = (1-\eta)\mathbb{P}^+ + \eta \mathbb{P}^-$. In this approach, it is crucial to determine $\eta$ and make a prior assumption about the distribution of anomalies $\mathbb{P^-}$, which may degrade the method generalization. Overall, the unsupervised settings for anomaly detection gained a great interest in learning commonalities of data from a complex and high-dimensional space without the need to access annotated training samples. Note that the self-supervised learning methods, that are the focus of this paper, can be considered as a subgroup of unsupervised learning techniques.

\item \textbf{Semi-supervised Anomaly Detection:}
In this setting, we assume that the training dataset is partially labelled and includes both labelled and unlabeled samples. Semi-supervised algorithms are suitable for scenarios where it is costly to annotate the whole data. This setting is also prevalent in anomaly detection because commonly, both labelled and unlabelled data are present, but labelling the data often requires expert knowledge, or in some cases, such as industrial and biomedical applications, anomalies are costly to occur.
Incorporating a small set of anomaly samples during training could significantly improve the detection accuracy and maximize the robustness of a model \citep{ruff2019deep, min2018ids, kiran2018overview}, especially compared to the unsupervised learning techniques. However, due to the scarce availability of the labelled abnormal samples, a semi-supervised setting is likely prone to overfitting. Therefore, making the correct assumptions about the distribution of anomalies, i.e. $\mathbb{P^-}$, is crucial for accurately incorporating the labelled anomalies in the training process.

It is important to note that Some existing papers refer to the task of \textit{Learning from Positive and Unlabeled examples (LPUE)} as semi-supervised learning \citep{chandola2007outlier}. Note that based on the above definitions, LPUE is an unsupervised learning technique where the entire training data belongs to the normal class. LPUE is commonly used in the literature to benchmark the anomaly detection algorithms using popular datasets, such as CIFAR-10 and MNIST \citep{ruff2018deep, golan2018deep}. In this task, the samples of one class of the dataset are deemed normal and are used during the training, and samples of other classes are considered anomalous \citep{hojjati2021dasvdd}. \textit{One-class AD} is another term which is used for referring to the LPUE task.

\item \textbf{Supervised Anomaly Detection:} In supervised anomaly detection, we assume that the dataset is fully labelled. When anomalies are easily annotated, it is more beneficial to adopt supervised methods\citep{feinman2017detecting, lee2018simple, jumutc2014multi, kim2015deep}.
At this point, it is essential to distinguish between supervised anomaly detection and binary classification problems. One might claim that if the normal and abnormal data are available during the training phase, the problem can be formulated as a supervised binary classification problem and will no longer be an anomaly detection task. However, we should note that, formally speaking, an anomaly is a sample that does not belong to the normal class distribution $\mathbb{P^+}$. The anomaly class includes a broad range of data points that are not accessible/known during the training phase. The common practice anomaly detection is to assume that, in the training phase, there are enough labelled samples from the normal class that can reveal  $\mathbb{P^+}$ while the limited available abnormal samples can only partially reveal $\mathbb{P^-}$. Hence, unlike binary classification, which aims to learn a decision boundary separating the two classes, AD seeks to discover the normal class boundaries. 
Although the supervised settings are more efficient and can achieve higher accuracy, they are rarely used to formulate anomaly detection problems compared to unsupervised and semi-supervised models. This is because, in most real-world applications, it is impossible to describe and have access to all existing anomaly classes.

\end{enumerate}

\section{Self-supervised Learning for Anomaly Detection}
\label{sec:section4}

\begin{figure*}[t]
  \centering
  \includegraphics[width=1\linewidth]{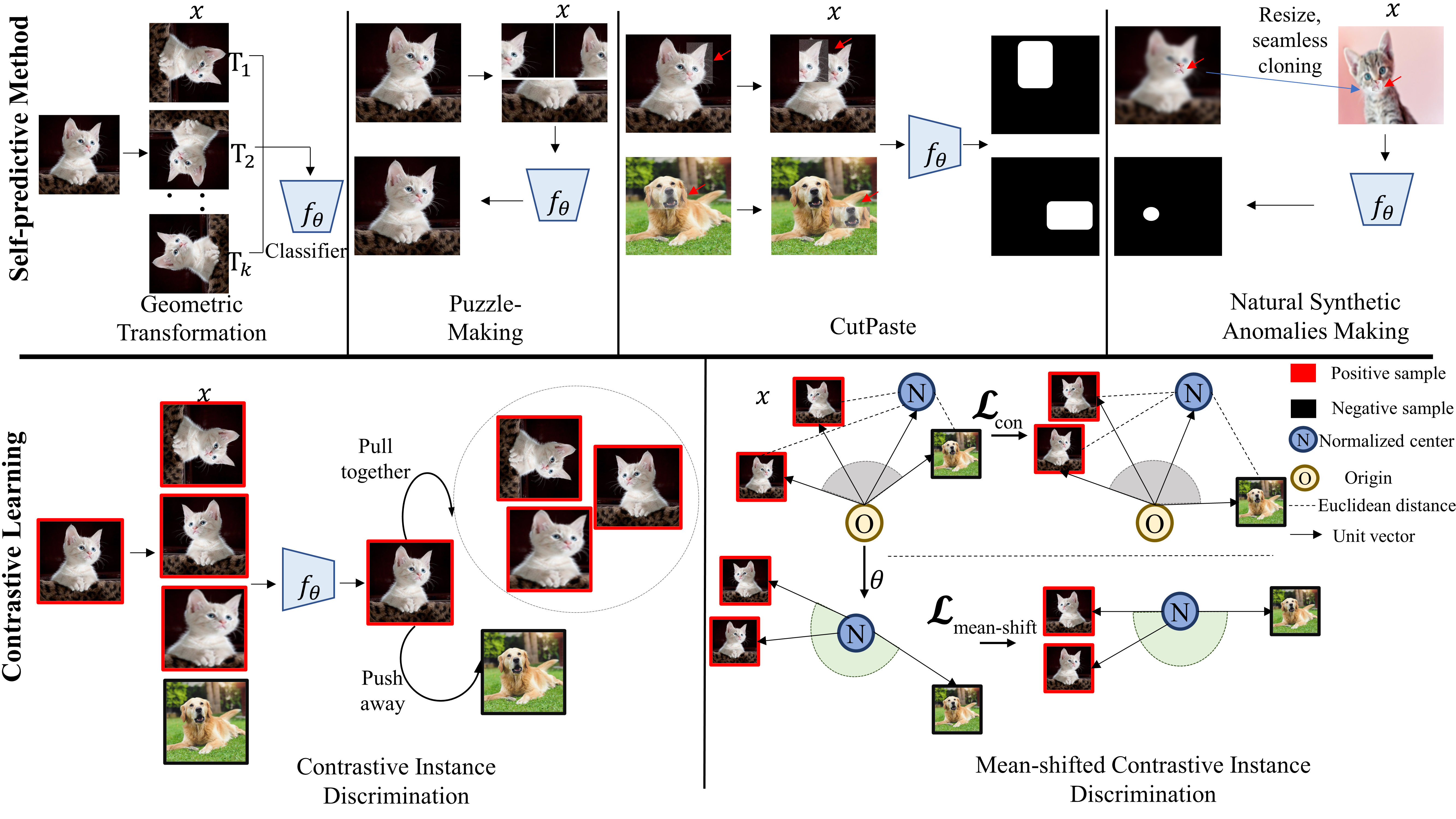}
  \caption{Several examples of pseudo-label generation processes that are associated with two main categories of SSL-AD. $x$ is the pseudo-labeled input and $f_{\theta}$ is the feature extractor.}
  \label{fig:ssl}
\end{figure*}

Self-supervised learning can leverage large amounts of unlabeled data to learn robust representations of normal behaviour, making it a scalable and cost-effective solution for anomaly detection.  In the subsequent sections of this paper, we will delve into the general methodology and contributions of self-supervised anomaly detection papers. Table \ref{tab: methods} and Table \ref{tab: methods_sum} provide a summary of the key aspects of these papers, including the task they aim to solve, the evaluation metrics used, and how they quantify the anomaly score from the representation. In addition, Figure \ref{fig:ssl} illustrates the methodologies employed by each group of methods.

\subsection{Problem Formulation} 
Based on the dataset's nature and the availability of data labels, the anomaly detection task can be formulated differently in past studies. The most common formulation is one-class anomaly detection (aka LPUE) \citep{golan2018deep,Sabokrou_2019_ICCV,chen2020simple}, in which one class of the dataset is trained as normal, while the remaining classes are considered abnormal. An example of this task is taking a class of the CIFAR-10 such as \textit{Cat} as normal, and the rest as anomalies. On the other hand, in \textit{multi-class anomaly detection}, multiple classes in the same datasets are considered normal during training, and one or multiple remaining classes are deemed anomalous \citep{ZHANG2022108234,tack2020csi}.

\subsection{Algorithms} \label{sec:alg}
\s{
Self-supervised algorithms can learn a proper data representation with the help of a defined pretext supervised task from an unlabelled dataset. The pretext task guides the model to learn a generic representation of the data, which can be helpful for downstream tasks such as classification and anomaly detection.}
\s{A wide range of proxy tasks and models are proposed in the literature of self-supervised learning. They include but are not limited to, colorization \citep{larsson2016learning}, maximization of mutual information between low-level and high-level representations \citep{hjelm2018learning}, and predicting geometric transformations \citep{gidaris2018unsupervised}. These methods showed promising results in various tasks, such as speech representation learning \citep{speechSSL}, visual feature learning \citep{VisualSSL}, and healthcare applications \citep{healthSSL}.}

Self-supervised anomaly detection models vary primarily based on the nature of their proxy tasks. The proxy task is designed to guide the model in learning a representation that is specifically suited for anomaly detection, as opposed to a generic representation learned by an unsupervised model. In recent years, contrastive learning methods have emerged as a significant component of self-supervised learning \citep{chen2020simple}. The primary objective of contrastive learning is to develop effective data representations by bringing together different views of the same sample while pushing them apart from other points. To accomplish this, various loss functions have been proposed, such as contrastive loss \citep{conloss} and triplet loss \citep{triplet}. Notably, several variants of contrastive learning models have demonstrated impressive accuracy levels comparable to those of fully-supervised models in specific tasks \citep{chen2020simple}. Anomaly detection is one of the tasks where SSL algorithms have demonstrated remarkable performance levels that were previously unattainable.

\s{Despite their recent success and broad applicability, self-supervised models suffer from several important shortcomings. One of their most significant problems is their computational inefficiency. Compared to a fully-supervised model, they need more time and data to train and get an accuracy comparable to their supervised counterparts.}

%

%
Inspired by earlier works, we categorize the self-supervised AD models based on their pretext task into two groups \citep{nipsssl}:

\begin{itemize}
\item \textbf{Self-predictive Methods:} These algorithms create the pretext task for each individual sample. Commonly, they apply a transformation to the input and try to either predict the applied transformation or reconstruct the original input. These models are effective even if only \textit{positive} samples, \textit{i.e.} in-distribution (IND) samples, are available. As a result, they do not necessarily require samples from other distributions, also known as \textit{negatives}, during training.
    
\item \textbf{Contrastive Methods:} \textit{Contrastive models} define the proxy task on the relationship between pairs of samples. They commonly generate positive views of a sample by applying different geometric transformations. Then, they aim to pull together the positives while pushing them away from the negative ones. In contrastive learning, samples of the current batch other than the anchor sample and its augmentations are considered \textit{negative} while \textit{positive} samples are the ones that are coming from augmentations of the anchor.  Technically, contrastive algorithms can also be considered  self-predictive. In essence, they also need to learn to predict the transformations to associate the same sample's augmentations with each other. However, the immense advancement of contrastive learning in recent years encouraged us to treat them as a stand-alone category.
\end{itemize}
Fig. \ref{fig:ssl} visually illustrates the representation learning process of these two categories. As shown in this figure, unlike self-predictive algorithms, contrastive learning methods incorporate negative samples. This figure also depicts the pseudo-label generation process for different SSL methods. Self-predictive models apply the transformations on positive samples and try to either predict the applied transformation or reconstruct the original input. Contrastive methods, on the other hand, do not explicitly predict the transformations or reconstruct the input and instead aim to distinguish between positive and negative samples. More details on the methods depicted in Fig. \ref{fig:ssl} are presented in Sections \ref{sec:section5} and \ref{sec:cl}.

\b{In the early stages, the primary focus of algorithms was on image and video anomaly detection. This emphasis was primarily due to the fact that self-supervised representation learning and the related proxy tasks were predominantly developed within the computer vision literature. Since a significant number of existing works concentrate on image anomaly detection, and this field is well-established, we first discuss the algorithms that were developed for visual anomaly detection, and subsequently, we will cover the papers that tried to tackle other data types in section 7.}

\subsection{Anomaly Scoring}
Self-supervised models are capable of learning a good feature representation from the input data. However, this representation is not readily useful for anomaly detection. Defining a suitable scoring function to quantify the degree of abnormality from this representation is essential for designing an anomaly detection framework. Previous studies have used a flurry of scoring functions based on the downstream tasks to detect anomalies. For example, two widely used anomaly scores for one-class anomaly detection are normality score and reconstruction error: Normality scores estimate the normality of new samples at the inference time after applying different transformations \citep{sohn2020learning,li2021cutpaste,hendrycks2019using,golan2018deep}. Examples of this type of score include the Dirichlet score \citep{golan2018deep} and rotation score \citep{hendrycks2019using}. Reconstruction error, which is typically measured by the Euclidean distance between the original and the reconstructed input, is another category of scoring functions. The assumption behind using this score is that the reconstructed features of anomalies have higher errors than normal samples \citep{Sabokrou_2019_ICCV,salehi2020puzzleae}. 
For multi-class anomaly detection, scoring functions such as class-wise density estimation (negative Mahalanobis distance) \citep{sehwag2021ssd} and data likelihood criterion \citep{zhang2021selfsupervised} were also used. Finally, for tackling the out-of-distribution detection problem, several other measures, including probability-based measures, rotation score \citep{hendrycks2019using}, Confusion Log Probability (CLP) \citep{Winkens}, Weighting Softmax Probability \citep{Mohseni_Pitale_Yadawa_Wang_2020}, and Mahalanobis distance \citep{sehwag2021ssd} are used in the self-supervised anomaly detection literature.

\subsection{Performance Evaluation}
To evaluate the performance of an anomaly detector, several criteria are used. In practical applications, the cost of false alarms (type I error) and missed-detected anomalies (type II error) are usually different. Most anomaly detectors define the decision function as
\[
    \text{Output}= 
\begin{cases}
    \text{normal}, & \text {if } Score(x) < \zeta \\
    \text{abnormal}, & \text {if } Score(x) \geq \zeta \\
\end{cases},
\]
where $Score(x)$ is the anomaly score for new sample $x$, and the decision threshold $\zeta$ is chosen to minimize the costs corresponding to the type I and II errors and to accommodate other constraints imposed by the environment \citep{field2004minimizing}. However, it is common that the costs and constraints are not stable over time or are not fully specified in various scenarios. As an example, consider a financial fraud detector that receives anomaly alarms to investigate potentially fraudulent activities. A detector can only handle a limited number of alarms, and its job is to maximize the number of anomalies containing these alarms based on the precision metric. Meanwhile, an anomaly alarm being wrongly reported can cause a credit card agency placing a hold on the customer’s credit card. Thus, the goal is to maximize the number of true alarms, given a constraint on the percentage of false alarms by using the recall metric.

Area Under the Receiver Operating Characteristic (ROC) Curve (AUC or simply AUC) is known for its ability to evaluate the model’s performance under a broad range of the decision threshold $\zeta$ \citep{auroc}. The AUROC curve is an indicator for all sets of precision-recall pairs at all possible thresholds. This makes AUC capable of interpreting the performance of models in various scenarios. As shown in Table \ref{tab: methods}, most anomaly detection methods use the AUC metric for evaluation. The random baseline achieves an AUC of 0.5, regardless of the imbalance between normal and abnormal subsets, while an excellent model achieves an AUC close to 1, demonstrating the robustness of the model in distinguishing normal from abnormal classes.

\begin{table*}[htbp]

\caption{Widely-used Self-Supervised Anomaly Detection methods}.
\label{tab: methods}
\begin{center}
\footnotesize
\renewcommand{\arraystretch}{1.2}
\scriptsize
\begin{tabular}{ c|c|c|c|c } 
\hline
  Category&Method & Task & Anomaly Score & Indicator\\ 
\hline \hline
\multirow{ 14}{*}{\makecell{Self\\Predictive}}&\makecell{GOEM \\ \citep{golan2018deep}}  & OCAD 
 & Dirichlet Normality
 & AUC \\
\cline{2-5}

&\makecell{NRE \\ \citep{Sabokrou_2019_ICCV}}& OCAD & Reconstruction Error 
 & EER \\
 \cline{2-5}

&\makecell{SSL-OE \\ \citep{hendrycks2019using}}  & \makecell{OCAD \\ OOD} & \makecell{Rotation Score} 
 & AUC\\
\cline{2-5}

&\makecell{GOAD \\ \citep{bergman2020classification}}   & OCAD &  Softmax Probability & AUC \\

\cline{2-5}

&\makecell{Puzzle-AE \\ \citep{salehi2020puzzleae}}  & OCAD &Error Normalization & AUC \\

\cline{2-5}
&\makecell{CutPaste\\\citep{li2021cutpaste}}  & OCAD & Density Estimator & AUC \\ 
\cline{2-5}

&\makecell{$\mathrm{SLA^2P}$\\\citep{slap}}    & \makecell{OCAD}& Uncertainty Score  & AUC \\
\cline{2-5}
&\makecell{NAF-AL \\ \citep{zhang2021selfsupervised}}   & \makecell{OCAD\\MCAD}& Likelihoods & F1 \\
\cline{2-5}
&\makecell{DAAD \\ \citep{ZHANG2022108234}}  &MCAD& \makecell{Probabilistic scalars \\ with majority voting
}  & \makecell{F1,AUC\\ ACC}\\
\cline{2-5}
&\makecell{Patch-Based \\ \citep{tsai2022multi}}  & OCAD & \makecell{$L_2$ Distance} 
 & AUC \\
\hline
\multirow{ 12}{*}{Contrastive}&\makecell{CLP\\ \citep{Winkens}}  & OOD & CLP Score & AUC \\
\cline{2-5}
&\makecell{CSI\\ \citep{tack2020csi}}& OCAD & \makecell{Cosine similarity, \\ Representation Norm} & AUC \\
\cline{2-5}
&\makecell{SSD \\ \citep{sehwag2021ssd}}  & \makecell{OOD} & \makecell{Mahalanobis distance} & AUC \\
\cline{2-5}
&\makecell{DROC \\ \citep{sohn2020learning}}    & OCAD & Normality score & AUC\\ 
\cline{2-5}
&\makecell{MCL \\ \citep{mcl}}  & MS AD & Mahalanobis distance & AUC \\
\cline{2-5}
&\makecell{NDA \\ \citep{chennegative}}   & ND & \makecell{Reconstruction error} & AUC \\
\cline{2-5}
&\makecell{Spatial CL \\ \citep{kim2022spatial}}  & OCAD & $L_2$ Distance & AUC \\
\cline{2-5}
&\makecell{Self-Distillation \\ \citep{rafiee2022self}}  & OOD & \makecell{Temperature-Weighted\\Nonlinear Score} & AUC \\

\hline
\hline
\end{tabular}
\end{center}
\end{table*}

\begin{table*}[htbp]
\caption{Summary of Widely-used Self-Supervised Anomaly Detection methods}.
\label{tab: methods_sum}
\begin{center}
\scriptsize
\renewcommand{\arraystretch}{1.2}
\begin{tabular}{ c|c } 
\hline
  Method & Summary \\ 
\hline \hline
\makecell{GOEM}  & \makecell{GEOM applies on all the given normal images and encourages learning \\ the features that are  useful for detecting novelties.}\\ 
\hline

\makecell{NRE}& \makecell{Besides learning a reconstruction scheme, AE preserves the local geometric \\ manifold based on NRE that leads to a discriminative neighborhood-guided SSL.}\\ 
\hline

\makecell{SSL-OE}  & \makecell{An auxiliary rotation loss is added to  improve \\ the robustness and uncertainty of deep learning models.}\\ 
\hline

\makecell{GOAD}  & \makecell{GOAD uses affine transforms which are suitable for general data. It tries to predict \\ the applied transforms and uses the output of the classifier to detect anomalies.}\\ 
\hline

\makecell{Puzzle-AE}  & \makecell{U-Net solves the puzzled inputs and the robust \\ adversarial training is used as an automatic shortcut removal.}\\ 
\hline

\makecell{CutPaste}  & \makecell{CutPaste augmentation creates local irregular patterns during training and \\ identifies these local irregularity on unseen real defects at the test time.}\\ 
\hline

\makecell{$\mathrm{SLA^2P}$}    &  \makecell{$\mathrm{SLA^2P}$ designs a discriminative anomaly score by employing feature-level \\ self-supervised learning and adversarial perturbation.}\\ 
\hline
\makecell{NAF-AL}  & \makecell{It employs data transformations in the SSL setting, and learns the data \\ likelihood by Autoregressive Flow-based Active Learning with Marginal Strategy.}\\ 
\hline
\makecell{DAAD}  & \makecell{It includes a classifier and an adversarial training model. It captures \\ different data distributions and makes an evaluation using the majority voting.}\\ 
\hline
\makecell{Patch-Based}  & \makecell{Incorporates relative feature similarity between patches of varying local distances \\  to enhance information extraction from normal images.} \\
\hline
\makecell{CLP}  & \makecell{A simple contrastive training-based approach for OOD detection is proposed. \\CLP captures the similarity of the inlier and outlier dataset(s).}\\ 
\hline
\makecell{CSI}& \makecell{A new detection score is introduced in the training phase that contrasts the \\ sample with distributionally-shifted augmentations of itself.}\\ 
\hline
\makecell{SSD} & \makecell{An outlier detector is based on only unlabeled in-distribution data. SSD uses SSL \\ followed by a Mahalanobis distance in the feature space.}\\ 
\hline
\makecell{DROC}   & \makecell{One-class AD emphasizes the importance of decoupling building \\ classifiers for learning representations.}\\ 
\hline
\makecell{MCL}  & \makecell{MCL can shape dense class-conditional clusters by adding 2 components: class-\\conditional mask and stochastic positive attraction to boost the performance.}\\ 
\hline
\makecell{NDA}  & \makecell{Negative augmentation generates negative samples closer to normal \\  samples and helps separate normal and abnormal points.}\\ 
\hline
\makecell{Spatial CL}  & \makecell{\makecell{Incorporates autoencoder in conjunction with contrastive learning \\ to reproduce the original image from cut-paste augmentation.}}\\ 
\hline
\makecell{Self-Distillation} & \makecell{\makecell{Employs the self-distillation of the in-distribution data, and contrasting against \\ negative examples that are generated through shifting transformations of data.}}\\

\hline
\hline
\end{tabular}
\end{center}
\end{table*}

\section{Self-Predictive Methods in Anomaly Detection}
\label{sec:section5}

Self-Predictive methods can learn data embedding by defining the supervised proxy task on a single sample. This approach focuses on the innate relationship between a sample and its own contents or its augmented views. An example of a self-predictive task is masking a portion of an image and trying to reconstruct it using a neural network \citep{salehi2020puzzleae}.

In most self-predictive approaches, the objective is to predict the label of the applied transformation, such as predicting the degree of rotation of an image. In this case, the anomaly score is commonly defined based on the Softmax probabilities of a supervised classifier. However, the objective of some methods is to reconstruct the original input from its transformed version. Solving the Jigsaw puzzle and denoising autoencoders are examples of this approach. In this case, the reconstruction error of the model is often used as the anomaly score. Geometric transformations were one of the earliest types of transformations that are used for visual representation learning. \citet{doersch2015unsupervised} showed that predicting the relative position of image patches is a helpful pretext task for improving the representation for object detection. In a later work, \citet{gidaris2018unsupervised} used rotation prediction for learning a better representation.

Geometric transformation models first create a self-labelled dataset by applying different geometric transformations to normal samples. The applied transformation is served as the label of each sample. Let $\mathcal{T} = \{T_1,T_2,\dots,T_K\}$ be the set of geometric transformations. The new labelled dataset $S$ can be constructed from the original dataset $\mathcal{D}$ as below:
\begin{equation}
    S := \{(T_j(x),j)|x\in \mathcal{D}, T_j \in \mathcal{T} \},
\end{equation}
where the original data point is shown by $x$. A multi-class network is trained over the dataset $S$ to detect the transformation applied to the sample. During the inference phase, the trained models are applied to the transformed versions of the samples, and the distribution of the Softmax output is used for anomaly detection \citep{golan2018deep}. 
Unlike the Autoencoders and GAN-based methods, the geometric transformation models are discriminative. The intuition behind these models is that the model learns to extract important features of the input by learning to identify the applied geometric transformations. These features can also be helpful for anomaly detection.

The paper by \citet{golan2018deep} was the first work that used geometric transformation learning for anomaly detection. They named their method as GEOM and showed that it can significantly outperform the state-of-the-art in anomaly detection. They showed that their model can beat the top-performing baseline in CIFAR-10 and CatsVSDogs datasets by $32\%$ and $67\%$, respectively.

To calculate the anomaly score of a sample from the Softmax probabilities, \citet{golan2018deep} combined the log-likelihood of the conditional probability of each of the applied transformations:

\begin{equation}
    n_S := \sum_{k=1}^{K} \log{p(y(T_k(x))|T_k)}
\end{equation}
Then, they approximated $p(y(T_k(x))|T_k)$ by a Dirichlet distribution:

\begin{equation}
    n_S = \sum_{k=1}^{K} (\tilde \alpha_k - 1).\log{y(T_k(x))}.
\end{equation}

An important issue of GEOM is that the classifier $p(y(T_k(x))|T_k)$ is only valid for samples the network encountered during the training. For other samples which also includes anomalies, $p(y(T_k(x))|T_k)$ can have a very high variance. To address this problem, \citep{hendrycks2018deep} proposed to use some anomalous samples during the training to ensure that $p(y(T_k(x))|T_k) = \frac{1}{M}$ for anomalies. This method, which is also known as Outlier Exposure (OE), formulates the problem as a supervised task which might not be practical for some real-world applications as they do not have access to anomalies.

Even though self-predictive models showed promising results, their performance is still significantly poorer than fully-supervised models in out-of-distribution detection. However, some recent studies \citep{hendrycks2019using} hinted that using SSL models in conjunction with supervised methods can improve the robustness of the model in different ways. Therefore, even in cases where we have access to anomaly data and labels, using self-supervised proxy tasks can enhance the performance of the anomaly detector. 

A significant downside of geometric models is that they only use transformations that are well-suited for image datasets and cannot be generalized to other data types, e.g. tabular data. To overcome this issue, \citet{bergman2020classification} proposed a method called GOAD. In GOAD, the data is randomly transformed by several affine transformations $\mathcal{T} = \{T_1,T_2,\dots,T_K\}$. Unlike the geometric transformations, affine transforms are not limited to images and can be applied to any data type. Also, we can show that the geometric transformations are special cases of the affine transform, and the GEOM algorithm is a special case of GOAD.
In GOAD, the network learns to map each of the transformations into one hypersphere by minimizing the below triplet loss:
\begin{align} 
L=\sum_i \max{(\|f(T_m(x_i))-c_m\|^2 + s}  \nonumber\\ 
- min_{m'\neq m} \|f(T_m(x_i))-c_{m'}\|^2,0),
\end{align} 
where $f(.)$ is the network, $s$ is a regularizing term for the distance between hyperspheres, and $c_m$ is the hypersphere center corresponding to the $m-$th transformation.

The above objective encourages the network to learn the hyperspheres with low intra-transformation and high inter-transformation variance. This is to provide a feature space, i.e. the last layer of $f(.)$, in which the different transformations are separated. During the inference phase, the test samples are transformed by all transformations and the likelihood of predicting the correct transform is used as the anomaly score.

Although classification-based methods showed significant improvement in semantic anomaly detection on datasets such as CIFAR-10, their performance is poor on real-world datasets such as MVTecAD \citep{salehi2020puzzleae}. This is because these models can learn high-level features of data by learning the patterns which are present both in the original data and its augmented versions, e.g. rotated instances. However, these algorithms might not be well-suited for sensory-level anomaly detection tasks, e.g. detecting cracks in an object. This is because some types of low-level anomalies, such as texture anomalies, are often invariant to the transformations.
To alleviate this issue, several other proxy tasks, that are more suitable for low-level anomaly detection, are proposed. For instance, \citet{salehi2020puzzleae} used the idea of solving the jigsaw puzzle for learning an efficient representation that can be used for pixel-level anomaly detection. Their proposed method, which they named as Puzzle-AE, trains a U-Net autoencoder to reconstruct the puzzled input. The reconstruction objective ensures that the model is sensitive to the pixel-level anomalies, while the pretext task of solving the puzzle enables the network to capture high-level semantic information, as shown in Fig.~\ref{fig:ssl}. They further boosted the performance of their model by incorporating adversarial training.

More recently, \citet{li2021cutpaste} developed a self-supervised method called CutPaste which significantly improves state-of-the-art in defect detection. CutPaste transformation randomly crops a local patch of the image and pastes it back to a different image location. The new augmented dataset is more representative of real anomalies. Thus, the model can be easily trained to identify and localize the local irregularity (shown by the white regions in the black background in Fig.~\ref{fig:ssl}). To detect the augmented samples from the un-transformed ones, the objective of the network is defined as follows:
\begin{equation}
    \mathcal{L}_{CP} = \mathbb{E}_{x \in \mathcal{X}} \{\mathbb{CE}(g(x),0) + \mathbb{CE}(g(\mathrm{CP}(x)),1)\},
\end{equation}
where $CP(.)$ is the CutPaste augmentation, $\mathcal{X}$ is the set of normal data, $\mathbb{CE}(.,.)$ is a cross-entropy loss, and $g$ is a binary classifier that can be parameterized by deep networks.
In order to calculate the anomaly score from the representation, an algorithm like KDE or GDE can be used.

CutPaste can also learn a patch representation and compute the anomaly score of an image patch by cropping a patch before applying CutPaste augmentation. This facilitates localizing the defective area. In this case, the objective loss function is modified as:

\begin{equation}
    \mathbb{E}_{x \in \mathcal{X}} \{\mathbb{CE}(g(c(x)),0) + \mathbb{CE}(g(\mathrm{CP}(c(x))),1)\},
\end{equation}
where $c(x)$ crops a patch at random location $x$.

In another similar work, \citet{schluter2021self}, introduced a new self-supervised task, called Natural Synthetic Anomalies (NSA) to detect and localize anomalies using only normal training data. Their proposed approach creates synthetic anomalies by seamlessly cloning a patch with various sizes from a source image into a destination image. In particular, NSA selects a random rectangular patch in the source image, randomly resizes the patch, blends the patch into the destination location from a different image, and creates a pixel-level mask. The new samples that NSA generates are different in size, shape, texture, location, color, etc. In other words, NSA dynamically produces a wide range of anomalies, which are more realistic approximation of natural anomalies than the samples that CutPaste creates by pasting patches at different locations. An example of NSA is shown in Fig.~\ref{fig:ssl}, where a random patch from a source cat image is seamlessly cloned onto another cat image. The NSA method outperforms the state-of-the-art algorithms on several real-world datasets such as MVTecAD.

\section{Contrastive Methods}
\label{sec:cl}

The primary objective of contrastive self-supervised learning is to learn a feature space or a representation in which the positive samples are closer together and are further away from the negative points. Empirical evidence shows that contrastive learning models such as SimCLR \citep{chen2020simple} and MoCo \citep{he2020momentum} are particularly efficient in computer vision tasks. SimCLR, one of most popular recent contrastive learning algorithms, learns representations by maximizing the agreement between different augmented versions of the same image while repelling them from other samples in the batch. Each image $x_i$ from randomly sampled batch $\mathcal{B}=(\{x_i,y_i\})_{i=1}^N$ is augmented twice, producing an independent pair of views $\{\hat{x}_{2i-1},\hat{x}_{2i}\}$, and augmented batch $\hat{\mathcal{B}}=\{(\hat{x}_i,\hat{y}_i)\}_{i=1}^{2N}$, where the labels of augmented data $\{\hat{y}_{2i-1},\hat{y}_{2i}\}$ are equal to the original label $y_i$. By performing independent transformation $\mathrm{T}$ and $\mathrm{T'}$ drawn from a pre-defined augmentation function pool $\mathcal{T}$, the augmented pair of views $\{\hat{x}_{2i-1}=\mathrm{T}(x_i),\hat{x}_{2i}=\mathrm{T'}(x_i)\}$ are \s{then }generated. Next, $\{\hat{x}_{2i-1},\hat{x}_{2i}\}$ are passed sequentially through an encoder and a projection head to yield latent vectors $\{z_{2i-1},z_{2i}\}$. SimCLR learns the representation by minimizing the following loss for a positive pair of examples $(m,n)$:
\begin{equation}
    l(m,n)=-\log \frac{\exp(sim(z_m,z_n)/\tau)}{\sum_{i=1}^{2N}\mathbf{1}_{\{i \neq m\}}\exp(sim(z_m,z_i)/\tau)}
\end{equation}
where $sim(z_m,z_n)$ represents the cosine similarity between the pair of latent vectors $(z_m,z_n)$, $\mathbf{1}_{i \neq m}$ is an indicator function which is equal to 1 if $i \neq m$ and zero otherwise, and $\tau$ indicates the temperature hyperparameter which determines the degree of repulsion. The final objective is to minimize the contrastive loss, defined in \eqref{eq:simclr}, over all positive pairs in a mini-batch: 
\begin{equation}
\label{eq:simclr}
    \mathcal{L}_{SimCLR} = \frac{1}{2N}\sum_{i=1}^{N} \Big[l(2i-1,2i)+l(2i,2i-1)\Big].
\end{equation}
Contrastive learning models established themselves as powerful representation learning tools. Still, they face crucial challenges for anomaly detection. Most widely-used contrastive learning algorithms, such as SimCLR and MoCo, need negative samples to operate. However, we either only have access to the samples from one class in many anomaly detection tasks, or the distribution of classes is highly imbalanced. In addition, the learned representation is not readily suitable for the anomaly detection task, and we need to define a proper anomaly score.

Despite these challenges, several contrastive anomaly detection models have emerged in the recent years. The CSI method proposed by \citet{tack2020csi} was the first attempt for using contrastive learning in anomaly detection. The CSI method is based on the idea of instance discrimination which considers every data point as a separate class and negative relative to other samples in the dataset \citep{Wu_2018_CVPR}. This idea is proven to be practical in visual representation learning for classification, but its performance in anomaly detection is unexplored \citep{chen2020simple}. They also showed that if specific transformations are used for generating negative samples from a given point, the learned representation can be more appropriate for anomaly detection. These distribution-shifting transformations can be denoted by a set as $\mathcal{S}$. In contrast to SimCLR, which considers augmented samples as positive to each other, CSI attempts to consider them as negative if the augmentation is drawn from $\mathcal{S}$. A significant conclusion of the CSI method is that although using the shifted transformations does not improve and even in some cases hurts the performance of the representation in other downstream tasks such as classification, it can improve the performance for anomaly detection.

If we denote the set of shifting transformations by $\mathcal{S} =\{S_0=I,S_1,...,S_{K-1}\}$ with $I$ being the identity function and $K$ different (either random or deterministic) transformations, the CSI loss can be written as:
 \begin{equation}
     \mathcal{L}_{con-SI}:=\mathcal{L}_{SimCLR}\Big(\bigcup_{S \in \mathcal{S}}\mathcal{B}_{S};\mathcal{T}\Big)
 \end{equation}
 in which $B_S := \{S(x_i)\}_{i=1}^B$. In simpler terms, the $\mathcal{L}_{con-SI}$ is essentially the same as the SimCLR loss, but in the con-SI, the augmented samples are considered negative to each other.
 
 In addition to discriminating each shifted instances, an auxiliary task is added with a Softmax classifier $p_{cls-SI}(y^{S}|x)$ that predicts which shifting transformation $y^S \in \mathcal{S}$ is applied for a given input $x_i$. The classifying shifted instances (cls-SI) loss is defined as below:
 \begin{equation}
     \mathcal{L}_{cls-SI}:=\frac{1}{2B}\frac{1}{K}\sum_{S \in \mathcal{S}} \sum_{\hat{x}_S \in \hat{\mathcal{B}}_S}-\log p_{cls-SI}(y^S = S|\hat{x}_S)
 \end{equation}
 The final loss of CSI is then defined as:
 \begin{equation}
     \mathcal{L}_{CSI}:=\mathcal{L}_{con-SI}+\lambda \mathcal{L}_{cls-SI}
 \end{equation}

The authors of the CSI empirically showed that the norm of the representation $\|z(x)\|$ is indeed a good anomaly score, where $z$ is the representation vector and $\| \cdot \|$ denotes the second norm.  This can be explained intuitively by considering that the contrastive loss increases the norm of the in-distribution samples to maximize the cosine similarity of samples generating from the same anchor. Consequently, during the test time, in-distribution samples are mapped further from the origin of the $z$ space, while the representation of other data points, i.e. anomalies, have a smaller norm hence are closer to the origin. This is an important observation as it helps to solve the problem of defining the anomaly score on a representation that is learned in an unsupervised fashion. The authors also found that the cosine similarity to the nearest training point in $\{x_m\}$ can be another good anomaly score. They defined the score of their model as a combination of these two metrics as below:
\begin{equation}
    s_{con}(x;\{x_m\}) := \max_m{\mathrm{sim}(z(x_m),z_x)}.\|z(x)\|
\end{equation}
where $z_x$ is the representation vector of the test sample $x$ and $z(x_m)$ is the closest representation vector in the training set.

Parallel to \citet{tack2020csi}, \citet{Winkens} developed a contrastive model for detecting out-of-distribution instances. They evaluated their approach on several benchmark OOD tasks and showed that contrastive models are also capable in OOD. The paper's key idea is that a fully supervised model might not be able to capture the patterns that can be useful for out-of-distribution detection. However, using contrastive learning techniques, the model learns high-level and task-agnostic features that can also help detect OODs. When we combine these techniques with the supervised learning techniques, the resulting model can learn more reliable features for both semantic classification and OOD detection.
 
The CSI algorithm shows that the task-agnostic representation learned through contrastive learning is suitable for anomaly detection. However, a task-specific approach can be more suitable for anomaly detection. (The task may be defined as the AD task itself or another downstream task such as data classification.) The contrastive models, such as SimCLR, are quite helpful in learning a representation for individual data points. They can also learn separable clusters for each class without having access to any labels. However, the resulting clusters may have blurry boundaries, and they commonly require fine-tuning for the downstream tasks.

To overcome this obstacle, \citet{mcl} developed a contrastive model which is tailored for anomaly detection. Their model, which is called Masked Contrastive Learning (MCL), modifies the degree of repulsion based on the labels of the data points. In vanilla SimCLR, all other batch samples, regardless of their class label, are considered negative relative to the anchor sample and are repelled with equal magnitude. However, in MCL, the repelling ratio is defined by the following class-conditional mask (CCM):

\begin{equation}
    \mathrm{CCM}(m,n)=
    \begin{cases}
        \alpha & \text{if } \bar y_m = \bar y_n\\
        \frac{1}{\tau} & \text{if } \bar y_m \neq \bar y_n
    \end{cases},
\end{equation}
where $0<\alpha<\frac{1}{\tau}$. Basically, CCM adjusts the temperature $\tau$ for the same labelled views to a smaller value of $\alpha$. This means that if the negative sample has the same class as the anchor, it is repelled with less magnitude compared to other data points. The SimCLR loss function is modified according to this mask as follows:

\begin{equation}
    \mathcal{L}_{CCM} = \frac{1}{2N}\sum_{i=1}^{N} \Big[l_{CCM}(2i-1,2i)+l_{CCM}(2i,2i-1)\Big],
\end{equation}

\begin{equation}
    p_{CCM}(m,n)= \frac{\exp(sim(z_m,z_n)/\tau)}{\sum_{i=1}^{2N}\mathbf{1}_{\{i \neq m\}}\exp(sim(z_m,z_i).\mathrm{CCM}(m,i)},
\end{equation}

\begin{equation}
    l_{CCM}(m,n) = - \log p_{CCM}(m,n),
\end{equation}

Although the proposed mask leads to a finer-grained representation space, the repulsive nature of the loss function may lead to the formation of scattered clusters. To prevent this phenomenon, the MCL algorithm stochastically attracts each sample to the instances with the same class label.

To further improve the MCL model in \citep{mcl}, an auxiliary classifier that predicts the applied transformation is also employed. The masking function is then modified based on the label of sample and its transformations. The repelling ratio is then smaller for the samples that simultaneously have the same class label and transformation labels, compared to the samples with the same class but different transformation labels. A sample with the latter property repels with a smaller magnitude than the negative points.

To score the anomalies in \citep{mcl}, the Mahalanobis distance \citep{MHD}, shown in \eqref{eq:MD}, is employed.

\begin{equation} \label{eq:MD}
    MD(x) = (z_x-\mu)^T\Sigma^{-1}(z_x-\mu),
\end{equation}
where $z_x$ is the representation of $x$, $\mu$ is the sample mean, and $\Sigma$ is the sample covariance of features of the in-distribution training data.
The Mahalanobis distance is a standard metric for scoring anomalies from their representation. It does not require any labelled data that makes it a common choice for many anomaly detection algorithms. In addition to this distance, the score of the auxiliary classifier is used to boost the model's robustness.

In another similar work, \citet{sehwag2021ssd} explored the applicability of contrastive self-supervised learning for out-of-distribution (OOD) and anomaly detection from unlabeled data, and proposed a method called SSD. They also extended their algorithm to work with labelled data in two scenarios: First is the scenario in which it is assumed that there are a few labelled out-of-distribution samples (i.e. a k-shot learning setting where k is set to 1 or 5), and the second scenario is the case in which labels of the in-distribution data are provided during the training phase.

In SSD \citep{sehwag2021ssd}, the SimCLR is used to learn the representation and the Mahalanobis Distance is incorporated to detect anomalies. For the cases where the labelled data is present, the authors suggested using the SupCon loss, defined in \eqref{eq:supcon}, which is a supervised variant of the contrastive loss \citep{supcon}, to have a more effective selection of the positive and negative samples for each image. In SupCon, samples from the same class are treated as positive and other samples as negatives.
\begin{align} \label{eq:supcon}
    \mathcal{L}_{SupCon} = & \nonumber\\
    \frac{1}{2N} \sum_{i=1}^{2N} - & \log{\frac{\frac{1}{2N_{y_m}-1}\sum_{i=1}^{2N}\mathbf{1}(i \neq m)\mathbf{1}(y_i=y_m)e^{u_m^T u_i/\tau}}{\sum_{i=1}^{2N}\mathbf{1}(i \neq m)e^{u_m^T u_i/\tau}}},
\end{align}
where $N_{y_m}$ refers to the number of images with label $y_m$ in the batch, and $u_i=\frac{h(f(x))}{\|h(f(x))\|_2}$ with a projection head $h(\cdot)$ and an encoder $f(\cdot)$.
Using SupCon loss yielded better performance compared to the contrastive loss throughout their experiments for the OOD detection from a labelled dataset. Overall, \citep{sehwag2021ssd} showed that the contrastive approach can outperform other methods in OOD detection in both labelled and unlabeled settings.

Contrastive models are also used in conjunction with one-class models for anomaly detection. One-class classifiers are one of the most widely used models in anomaly detection. They can detect anomalies after learning from a single class of examples. \citep{sohn2020learning} employed a two-stage framework for detecting anomalies using self-supervised learning models. In this framework, an SSL-based neural network is used to learn the representation of the input. A one-class classifier, such as OCSM or KDE, is applied to the learned representation to detect anomalies. The two-stage framework eliminates the need for defining an anomaly score and, as is empirically demonstrated in the paper, it can outperform other state-of-the-art methods.

Despite their promising empirical results, one-class classifiers suffer from a critical problem known as catastrophic collapse. This  phenomenon happens when the network converges to the trivial solution of mapping all the inputs to a single point regardless of the input sample value $x$, i.e. $\phi(x)=c$ where $\phi(\cdot)$ denotes the network output. This trivial solution is obtained when minimizing the center-loss defined as $\mathcal{L} = ||\phi(x)-c||^2$ \citep{reiss2021panda,ruff2018deep}. The features that the network learns in such case are uninformative and cannot be used for distinguishing anomalies from normal data. This issue is also known as ``hypersphere collapse''.

To overcome the hypersphere collapse problem, \citet{reiss2021mean} proposed a new loss function, called Mean-shifted contrastive loss (MSCL). Unlike the conventional contrastive loss, where the angular distance is computed relative to the origin, MSCL measures the angular distance relative to the normalized center of the extracted features. An example of MSCL is shown in Fig.~\ref{fig:ssl}. Formally, for a sample $x$, the mean-shifted representation is defined as:
\begin{equation*}
    \theta(x) = \frac{\phi(x)-c}{\|\phi(x)-c\|},
\end{equation*}
The mean-shifted contrastive loss is then given by:
\begin{align}
    \mathcal{L}_{MSCL}(x',x'') & = \mathcal{L}_{CONS}(\theta(x'),\theta(x'')) \nonumber\\
    & = - \log \frac{exp((\theta(x').\theta(x''))/\tau)}{\sum_{i=1}^{2N}\mathbf{1}[x_i \neq x'].exp((\theta(x').\theta(x_i))/\tau)},
\end{align}
where $\mathcal{L}_{CONS}$ is the typical contrastive loss for a positive pair, shown in SimCLR \citep{chen2020simple}, and $x,x''$ are the two augmentations of the input $x$.

One limitation of the MSCL loss is that it implicitly encourages the network to increase the distance of features from the center. Because of this, normal data lie in a region far away from the center. To solve this issue, the loss function is modified by adding the angular center loss, which shrinks the distance of normal samples from the center. \citet{reiss2021panda} showed that the overall loss, which is a combination of the MSCL and the angular losses, can achieve a better training stability and higher accuracy in anomaly detection than the regular center-loss.

In summary, recent papers suggest that the representation that is learned through self-supervised learning is indeed very useful for anomaly detection. An interesting observation is that even a simple scoring function such as the norm of the representation $\|z\|$ can be used for detecting anomalies from the representations. This can be justified because, in CL-based models, the normal data is spread out on a hypersphere. This property can help to define the anomaly score as the distance of the representation from the center. A smaller distance means a higher probability of the point belonging to the anomaly class.

\b{\section{Self-Supervised Anomaly Detection Beyond Images}}
\label{sec:nonimage}

\begin{table*}[htbp]

\caption{Self-Supervised Anomaly Detection for Non-Image Data}.
\label{tab: nonimage}
\begin{center}
\footnotesize
\renewcommand{\arraystretch}{1.2}
\scriptsize
\begin{tabular}{ c|c|c|c } 
\hline
  \textbf{Data Type} & \textbf{Paper} & \textbf{Type} & \textbf{Idea}\\ 
\hline \hline
\multirow{ 8}{*}{\makecell{\textbf{Audio}}}& \citet{Giri2020}  & Self-Predictive 
 & Machine ID Classification
  \\
\cline{2-4}

&\citet{complex}& Self-Predictive & Machine ID Classification 
  \\
 \cline{2-4}

&\citet{AADCL}  & Contrastive & \makecell{Pitch Shift, Fade In/Out,\\ Time-Stretch, etc.} 
 \\

 \cline{2-4}
&\citet{icasp232}& Contrastive & \makecell{Machine ID Classification\\Contrastive Pretraining}
  \\
 \cline{2-4}

 &\citet{icasp23}& Contrastive & \makecell{Joint Generative/Contrastive \\ Representation Learning} 
  \\
 \cline{2-4}
 
 &\citet{BAI2023103939}& Self-Predictive & \makecell{Time Masking and \\ Machine ID Classification} 
  \\
  
\hline
\multirow{ 11}{*}{\makecell{\textbf{Time-} \\ \textbf{Series}}}& \citet{neuralcontext}& Self-Predictive &  Anomaly Injection \\
\cline{2-4}
& \citet{ho2022self} & Contrastive & \makecell{Graph Contrastive Learning \\ Masked Sensor Reconstruction}\\
\cline{2-4}
& \citet{hadi2023vehicle} & Contrastive & \makecell{Contrastive Learning \\ Between Time Blocks} \\
\cline{2-4}
& \citet{wang2023deep} & Contrastive & \makecell{Joint contrastive and\\ one-class classification}  \\
\cline{2-4}
& \citet{jeong2023anomalybert} & Self-Predictive & Synthetic Anomaly Injection  \\
\cline{2-4}
& \citet{ICKG} & Self-Predictive & Intra-Sample Prediction Task  \\
\cline{2-4}
& \citet{MAD} & Self-Predictive & Masked Data Reconstruction  \\
\cline{2-4}
& \citet{TimeAutoAD} & Contrastive & Pseudo-Negative Generation  \\
\cline{2-4}
& \citet{HUANG2022261}& Self-Predictive &  \makecell{Detection the Downsampling \\ Resolution}  \\

\hline
\multirow{ 16}{*}{\textbf{Graph}}& \citet{liu2021anomaly} & Contrastive & Sub-graph Contrastive Learning  \\
\cline{2-4}
& \citet{zheng2021generative} & Contrastive & \makecell{Sub-graph Contrastive Learning \\ and Node Reconstruction} \\
\cline{2-4}
& \citet{duan2022graph} & Contrastive & \makecell{Graph Views with Node- and \\ Sub-graph-level Contrastive Learning} \\  
\cline{2-4}
& \citet{chen2022gccad} & Contrastive &  \makecell{Node-level Supervised \\ Contrastive Learning}  \\
\cline{2-4}
& \citet{xu2022contrastive} & Contrastive &  \makecell{Graph-level Supervised \\ Contrastive Learning and Reconstruction}  \\
\cline{2-4}
& \citet{zheng2022unsupervised} & Contrastive & \makecell{Graph-level Few-shot \\ Contrastive Learning}  \\
\cline{2-4}
& \citet{huang2022hop} & Self-Predictive & \makecell{Node- and Graph-level based \\ Hop Count Prediction}   \\
\cline{2-4}
& \citet{liu2021Dynanomaly} & Contrastive & \makecell{Edge-level Contrastive \\ Learning in Dynamic Graphs}  \\
\cline{2-4}
& \citet{luo2022deep} & Contrastive & \makecell{Node- and Graph-level Contrastive \\ Learning}  \\ 
\cline{2-4}
&\citet{ho2022self} & Contrastive & \makecell{Node- and Sub-graph-level \\ Contrastive Learning and Reconstruction}   \\

\hline
\multirow{ 3}{*}{\makecell{\textbf{Other}}}& \citet{pmlr-v139-qiu21a} & Self-Predictive &  Trainable Transformations \\
\cline{2-4}
& \citet{manolache-etal-2021-date} & Self-Predictive & \makecell{Text Anomaly Detection}\\
\cline{2-4}
& \citet{shenkar2022anomaly}& Contrastive &  \makecell{Tabular Data Anomaly Detection}  \\

\hline
\hline
\end{tabular}
\end{center}
\end{table*} 

\b{In recent years, there has been a growing interest in extending self-supervised anomaly detection techniques beyond image data. While the majority of early research in anomaly detection focused on image and video data, the need to detect anomalies in various other data types, such as text, audio, and time series, has become increasingly apparent. In this section, we delve into the advancements made in self-supervised anomaly detection methods that specifically target non-image data.}

\b{A crucial aspect of self-supervised learning methods is the selection of data-specific augmentations and proxy tasks. In the context of non-image self-supervised anomaly detection, a primary focus lies in defining a set of augmentations and proxy tasks that are effective for detecting anomalies. Inspired by image anomaly detection models, many algorithms have sought to adapt and extend these techniques for different data types. Table \ref{tab: nonimage} summarizes the important papers in this field. In the following subsections, we explore various data types and their corresponding algorithms, shedding light on their augmentations and proxy tasks.}

\subsection{Audio Anomaly Detection}
\b{Audio data plays a significant role in various applications, including speech recognition, environmental monitoring, and acoustic anomaly detection. The detection of audio anomalies has been a longstanding research challenge. However, more recently, self-supervised methods have emerged as successful approaches for addressing this task. In the realm of audio data, much like in images and videos, the outcomes of augmenting transformations can be evaluated qualitatively. As a consequence, the literature has already established a robust set of positive and negative transformations that have proven effective. These include well-known techniques such as noise injection, pitch shifting, and fade in/fade out, among others. These established transformations have been used in conjunction with the ideas from self-supervised visual anomaly detection to develop new models for acoustic data. Another helpful aspect of audio data is that their spectrogram, which is an essential tool in anomaly detection, can be used as input to computer vision models such as CNNs. As a result, they can are compatible with existing image self-supervised representation learning tools.}

\b{\citet{Giri2020} was one of the first studies that adapted the idea of self-supervised learning for detecting abnormal machine conditions. They have incorporated augmentations such as linearly combining the audio and warping the spectrograms in order to learn a representation which is suitable for anomaly detection. Their research demonstrated that their proposed method surpasses existing baselines by a significant margin.
In another similar work, \citet{complex} introduced an innovative framework for acoustic anomaly detection that incorporates the concept of self-supervision. In this algorithm,  accurately identifying the machine ID associated with a given sound is defined as the proxy task. Additionally, they leveraged phase continuity information and employed the complex spectrum as input to their model. During the inference phase, any data that the model was unable to classify correctly with the corresponding machine ID has been deemed an anomaly. The experimental evaluations conducted in the paper demonstrated that the utilization of a simple proxy task yielded impressive results, significantly enhancing the model's ability to detect anomalies.}

\b{For the first time, \citet{AADCL} introduced a contrastive framework for acoustic anomaly detection. They defined a comprehensive set of transformations, such as time and frequency masking, pitch shift, and noise injection, specifically designed for audio data. These transformations were utilized to create positive and negative pairs for training a contrastive learning algorithm. They have shown that this approach significantly outperforms other existing methods and highlighted the remarkable improvement that could be achieved through contrastive learning in acoustic data. Following this work, \citet{icasp232} proposed a method that combines contrastive learning with the proxy task of machine ID detection to improve accuracy.}

\b{These advancements have shown promise in detecting anomalous sounds, such as abnormal environmental sounds or audio events in surveillance systems.}

\subsection{Time-Series Anomaly Detection}
\b{Time series data arises in a wide range of domains, including finance, manufacturing, and healthcare. Detecting anomalies in time series is crucial for identifying unusual patterns or behaviours. Self-supervised learning techniques have been leveraged to capture temporal dependencies and detect anomalies in time series data.}

\b{Unlike images, videos, and audio data, defining suitable augmentations for time-series data is an exceptionally challenging task that heavily relies on the target application and characteristics of the data. Despite this inherent difficulty, researchers have proposed several ideas in recent years to adapt the self-supervised learning framework to time series. One particularly popular approach, which can be applied to a wide range of time series, involves injecting synthetic anomalies and training the network to distinguish them from positive samples. In an early attempt, \citet{neuralcontext} developed \textit{Neural Contextual Anomaly Detection} (NCAD), which could learn the boundary between normal and abnormal samples by injecting pseudo-negative samples during training. To generate these anomalies, they drew inspiration from \citet{hendrycks2019using}, and replaced segments of the original time series with values obtained from another time series. To further enhance the diversity of the negative set, they also included synthetic point anomalies. A similar concept was employed by \citet{TimeAutoAD} to generate synthetic anomalies and train a representation using contrastive learning, which enables the discrimination between positive and negative samples. Very recently, \citet{jeong2023anomalybert} used the idea of synthetic anomaly injection in conjunction with the self-attention mechanism to detect abnormal sequences with high accuracy.}

\b{Another widely applicable and popular idea is the masking of a segment of the time-series data and training the network to reconstruct it. This concept has been successfully employed in image and audio anomaly detection. Notably, \citet{MAD} demonstrated that this approach could also be effectively utilized for learning efficient representations in time-series data. The underlying assumption behind this idea is that by learning to reconstruct the masked segment, the network will learn the patterns that are present in normal data. In the case of multivariate time series, a possible implementation involves masking the data of one time series and using the data from other entities to reconstruct or predict it \citep{ho2022self,ICKG}. This allows the model to capture the dependencies and relationships between different entities within the time-series data.}

\b{A notable trend in time-series anomaly detection involves leveraging temporal information of the data. This approach aims to capture meaningful patterns and enhance the learning of efficient representations. For example, \citet{HUANG2022261} demonstrated that predicting the downsampling resolution of the data can significantly contribute to learning effective representations from time series. By incorporating the downsampling resolution prediction task, the network is encouraged to understand the underlying temporal structure and capture essential features at different resolutions. This enables the model to develop a comprehensive understanding of the time-series data, leading to improved anomaly detection performance.
Additionally, researchers such as \citet{hadi2023vehicle} have utilized temporal adjacency information to generate positive and negative pairs for training contrastive learning models. This approach enhances the model's ability to capture contextual information and detect anomalies by comparing similar and dissimilar pairs of temporal instances.}

\b{In conclusion, self-supervised learning techniques offer promising avenues for time-series anomaly detection. The injection of synthetic anomalies, along with methods such as contrastive learning and resolution prediction, enables the network to learn efficient representations and distinguish between normal and abnormal sequences.}

\subsection{Graph Anomaly Detection}
\b{Following the great success of SSL in the image/signal/text domains, very recently, SSL has gained significant attention in graph-structured data. A graph is a representation of a network, consisting of nodes that represent entities (e.g., objects, users, sensors) and edges that represent the interactions between entities. These interactions/relationships between nodes are known as structural dependencies and are expressed by the adjacency matrix (aka a square matrix) \citep{liu2022graph}. Each row and column of the matrix is associated with a node in the graph. The non-zero value in the entry of the matrix indicates whether there is an edge between two nodes. Given this unique property, graphs are different from other domains since the samples (nodes) are dependent on each other in the graph, while the samples in images or texts are independent. Due to such dependencies, it is therefore non-trivial to adopt pretext tasks designed for images or texts directly to graphs.}

\b{Many recent SSL methods have provided well-designed pretext tasks based contrastive learning that are applicable for graphs to deal with graph anomaly detection, the task of detecting anomalies (e.g., anomalous nodes, edges, sub-graphs) in \textit{static} graphs. Note that in a static graph, oftentimes seen in social networks, the sets of nodes/edges and their features, as well as the adjacency matrix are fixed. \citep{liu2021anomaly} proposed a local sub-graph-based sampling, which pays attention to the relationship between a node and its neighbors in a static graph, to select constrastive pairs. A pair consists of a node and its neighboring sub-graph. A positive pair composes of a target node and its neighboring sub-graph, while a negative pair consists of a node and its corresponding sub-graph. Note that a target node can be any node in a graph, a selected node in a negative pair is different from the target node selected in a positive pair, hence there is mismatching between the target node and the sub-graph in a negative pair. A constrative-based module is designed to estimate the matching between the target node and sub-graphs in constrastive pairs and would assign the abnormality level for every node.} 

\b{\citep{zheng2021generative} also aimed to compute the level of abnormality of every node in a static graph by designing an effective graph view sampling technique. Given a target node, two positive sub-graphs are sampled, and two negative sub-graphs are sampled randomly and guaranteed that they are different from positive sub-graphs. They designed two pretext tasks, one is to determine the mismatching between the target node and its sub-graphs in contrastive pairs as similar to \citep{liu2021anomaly}, the other is to reconstruct the target node's features based on surrounding nodes in positive sub-graphs. As a result, by taking advantage of multiple pretext tasks, \citep{zheng2021generative} showed better detection performance on anomalous nodes than \citep{liu2021anomaly}.}

\b{Not limited to sub-graph-level sampling, \citep{duan2022graph} showed the effectiveness of combining various contrastive pair sampling strategies. Given the original graph input as the first view, they adopted edge modification to generate the second view of the graph. For each view, they combined node-subgraph, node-node and subgraph-subgraph sampling techniques. The first two techniques can capture sub-graph- and node-level anomalous information in each view, while the latter focuses on more global anomalous information between two views. They showed that a diversity of sampling techniques helps to learn more representative and intrinsic graph embeddings, which could further improve the anomaly detection performance. }

\b{While the above studies are unsupervised graph anomaly detection methods, i.e., no annotated labels are available in the training phase, several studies leveraged prior human knowledge on graph anomalies. For example, \citep{chen2022gccad} took advantage of prior human knowledge, hence, they designed a contrastive loss and trained the model in a supervised manner, i.e., labeled normal and abnormal nodes are respectively treated as positive and negative samples. 
\citep{xu2022contrastive} also used human knowledge for helping the detection performance, but the way of building their contrastive pairs is different from \citep{chen2022gccad}. Given the actual anomalous static graph, they augmented a new graph by a knowledge modeling technique, then fed both original and augmented graphs to a Siamese graph neural network such that both graphs are encoded into the same latent space, making it feasible to contrast original and augmented graphs. After encoding, they designed a constrative loss that is integrated with the human knowledge of anomalies, i.e., the contrastive loss would guide the encoder to differently represent the nodes in the original graph and the nodes in the augmented graph. 
\citep{zheng2022unsupervised} also verified the effectiveness of having prior human knowledge in graph anomaly detection by proposing to use few anomalous samples in the training phase. This technique is known as few-shot supervised learning that could enrich the supervision signals for the model, hence, the detection accuracy could be improved.}

\b{As is seen from the aforementioned studies, most of techniques used the local context of graphs (i.e., the sub-graph knowledge) and adopted contrastive learning, however, \cite{huang2022hop} showed that using only local information is insufficient to effectively detect anomalies. More specifically, they designed a self-predictive framework for hop count (aka the shortest path length between pairs of nodes) prediction task, which considers both local and global information. The intuition behind hop counts based on local and global information is that since node-level anomalies are different normal nodes at both the feature- and adjacency matrix-levels, the distance between an anomalous node and its surrounding nodes should be larger than that between a normal node and its neighboring nodes. Hence, computing hop counts based on both local and global information can be useful to construct an anomaly indicator.}
 
\b{SSL with well-designed pretext tasks has shown a capability to handle complex structural dependencies and detect graph anomalies in static graphs. However, detecting anomalous graph objects raises an even more difficult problem in a \textit{dynamic} graph (aka a graph set), which consists of consecutive temporal graphs indexed in time, hence, the feature sets of nodes/edges and adjacency matrices change overtime. Time-series signals, edge streams in social networks, and videos are some of the examples that can be converted to dynamic graphs \citep{ho2023graph}. Several studies have shown the potential of SSL to detect anomalies in dynamic graphs. For example, \citep{liu2021Dynanomaly} aimed to detect edge-level anomalies at different time steps in an edge stream by designing a dynamic graph transformer-based contrastive learning. Positive edges are sampled from the normal training set while negative edges are randomly sampled based on a random sampling technique and are guaranteed that these negative samples are different from positive samples.}

\b{Other additional examples have demonstrated the ability of SSL in dynamic graphs constructed from different data modalities. For example, \citep{luo2022deep} aimed to detect anomalies in molecular networks, protein networks and social networks. They first constructed dynamic graphs for these networks. Then, they leveraged contrastive learning to capture both node-level and graph-level representations by a dual-graph encoder, and aimed to detect graph-level anomalies. \citep{ho2022self} aimed to effectively construct a graph set for time-series signal data, and then detect node-level and sub-graph-level anomalies in constructed graphs. To do so, they utilized the reconstruction-based and contrastive-based SSL pretext tasks to effectively capture the local sub-graph information in graphs.}

\b{In conclusion, SSL have yielded promising results for detecting anomalous graph objects at the node-, edge-, sub-graph- and graph-levels in both static and dynamic graphs. Using the knowledge of the features sets of nodes/edges, the adjacency matrices, the local and global information in graphs, and more importantly designing a diversity of effective graph augmentation techniques for pretext tasks would significantly improve the method's detection performance.}

\subsection{Anomaly Detection in Other Non-Image Data Types}
\b{
Beyond Graphs, audio, and time series data, self-supervised anomaly detection techniques have also been successfully applied to other data types. In particular, Shenkar et al. (2022) introduced an innovative contrastive learning algorithm specifically designed for tabular data. Their approach involved incorporating the concept of feature masking as a proxy task. During the training process, the model learns to create a mapping that maximizes the mutual information (MI) between the original samples and the masked features. To identify anomalies, the contrastive loss itself is directly used as the anomaly score. The findings of this study demonstrated the efficacy of self-supervised learning in tabular anomaly detection.}

\b{Another area that has recently garnered attention is text anomaly detection using self-supervision. \citet{manolache2021date} introduced a novel proxy task called \textit{Replaced Mask Detection} (RMD), which involves two steps: I) Masking a particular word in the input, and II) Replacing the masked word with an alternative. The model is trained to differentiate between the original and transformed versions of the text. Through extensive analysis, the authors demonstrated that the proposed framework achieved significant improvements in text anomaly detection.}

\b{Self-supervised models have indeed achieved remarkable success in various domains. However, their effectiveness is often dependent on the specific transformations they employ, which can limit their applicability. Fortunately, certain transformations, such as data masking, have proven to be adaptable across different data types. Drawing inspiration from this observation, a dedicated line of research has emerged with the goal of developing self-supervised methods for anomaly detection that can be applied to diverse data types. This research aims to create techniques that leverage self-supervision to detect anomalies effectively and efficiently in various domains, expanding the scope of self-supervised anomaly detection beyond specific data types. The work of \citet{pmlr-v139-qiu21a} is one of the most notable papers in this field. They have introduced the concept of trainable transformations that can be flexibly applied to any data type. The fundamental principle behind their approach involves mapping transformed data into a representation where distinct transformations can be discerned while still preserving the similarity between the transformed and original data. Remarkably, their framework demonstrates the capability to learn domain-specific transformations when applied to diverse datasets, including medical data and cyber-security data. This ability to adapt to different data types underscores the versatility and potential of their method in anomaly detection applications.}

\b{In conclusion, the field of self-supervised anomaly detection has expanded beyond image data, with significant progress made in detecting anomalies in non-image data types. By leveraging self-supervised learning techniques tailored to specific data modalities, researchers have demonstrated promising results in detecting anomalies in text, audio, time series, graphs, and IoT sensor data. These advancements open up new possibilities for anomaly detection in a wide range of applications, contributing to the development of robust and versatile anomaly detection systems.}

\section{Comparative Evaluation and Discussions}
\label{sec:section6}

\b{In this section, we focus on presenting the results reported by self-supervised image anomaly detection papers in a comparative manner to gain valuable insights into their performance. It is important to note that we have chosen to analyze only image data in this section, excluding other data types. This decision was made due to the inherent variations in datasets and backbones used across different studies, which could potentially introduce unfair comparisons. By focusing specifically on image data, we can provide a more meaningful and unbiased evaluation of the self-supervised anomaly detection methods.}

A flurry of datasets is used to benchmark the self-supervised anomaly detection algorithms. CIFAR-10 \citep{cifar}, and MVTecAD \citep{MVTECAD} are two of the most common dataset that recent anomaly detection papers used. CIFAR-10 includes images of ten different objects. To benchmark an AD algorithm on this dataset, we assume that we only have access to the data from one of the classes during the training. During the test time, other classes are considered to be anomalies. 

Table \ref{cifar} presents the result of several state-of-the-art SSL models against the commonly-used shallow and deep baselines for one-class AD on the CIFAR-10 dataset. This task can evaluate the performance of algorithms in semantic (high-level) anomaly detection. It is important to note that for the sake of fair comparison, we included the methods that use the same backbone. Looking at this table, we can readily confirm that the self-supervised approaches can outperform other shallow and deep anomaly detection algorithms by a significant margin. This remarkable improvement leads to the emergence of SSL algorithms as a key category of anomaly detection.

\begin{table*}[t]
\scriptsize
\caption{Performance of Self-Supervised Models on CIFAR-10 against shallow and deep baselines. The bold values denote the highest AUROC ($\%$) result for each class}
\setlength{\tabcolsep}{3.5pt}

\label{cifar}
\begin{center}
\begin{tabular}{@{\hspace{1mm}}c@{\hspace{1mm}}c@{\hspace{1mm}}c@{\hspace{1mm}}c@{\hspace{1mm}}c@{\hspace{1mm}}c@{\hspace{1mm}}c@{\hspace{1mm}}c@{\hspace{1mm}}c@{\hspace{1mm}}c@{\hspace{1mm}}c@{\hspace{1mm}}c@{\hspace{1mm}}c@{\hspace{1mm}}c@{\hspace{1mm}}c@{\hspace{2mm}}c@{\hspace{1mm}}c@{\hspace{0mm}}c@{\hspace{0mm}}}
    \toprule 
     &&\multicolumn{5}{c}{\textbf{Baseline}} 
     & \multicolumn{7}{c}{\textbf{Self-Predictive Method}} 
     & \multicolumn{4}{c}{\textbf{Contrastive Learning}} \\
    \cmidrule(lr){3-7}\cmidrule(lr){8-14}\cmidrule(lr){15-18}
    &\tiny{Class}& \thead{\tiny{KDE}}  & \thead{\tiny{OCSVM}} & \thead{\tiny{DSVDD}} & \thead{\tiny{OCGAN}} & \thead{\tiny{DROCC}} & \thead{\tiny{GEOM}} & \thead{\tiny{RotNet}} & \thead{\tiny{OE}} & \thead{\tiny{GOAD}} & \thead{\tiny{Puzzle}} & \thead{\tiny{SSLOE}} & \thead{\tiny{PANDA}} & \thead{\tiny{CSI}} & \thead{\tiny{SSD}} & \thead{\tiny{NDA}} & \thead{\tiny{MSCL}} \\
    \midrule
    &Plane&61.2&65.6&61.7&75.7&81.7&74.7&71.9&87.6&77.2&78.9&90.4&97.4&89.9&82.7&\textbf{98.5}&97.7\\
    &Car&64.0&40.9&65.9&53.1&76.7&95.7&94.5&93.9&96.7&78.2&99.3&98.4&\textbf{99.1}&98.5&76.5&98.9\\
    &Bird&50.1&65.3&50.8&64.0&66.7&78.1&78.4&78.6&83.3&69.9&93.7&93.9&93.1&84.2&79.6&\textbf{95.8}\\
    &Cat&56.4&50.1&59.1&62.0&67.1&72.4&70.0&79.9&77.7&54.9&88.1&90.6&86.4&84.5&79.1&\textbf{94.5}\\
    &Deer&66.2&75.2&60.9&72.3&73.6&87.8&77.2&81.7&87.8&75.5&97.4&\textbf{97.5}&93.9&84.8&92.4&97.3\\
    &Dog&62.4&51.2&65.7&62.0&74.4&87.8&86.8&85.6&87.8&66.0&94.3&94.4&93.2&90.9&71.7&\textbf{97.1}\\
    &Frog&74.9&71.8&67.7&72.3&74.4&83.4&81.6&93.3&90.0&74.8&97.1&97.5&95.1&91.7&97.5&\textbf{98.4}\\
    &Horse&62.6&51.2&67.3&57.5&71.4&95.5&93.7&87.9&96.1&73.3&98.8&97.5&\textbf{98.7}&95.2&69.1&98.3\\
    &Ship&75.1&67.9&75.9&82.0&80.0&93.3&90.7&92.6&93.8&83.3&\textbf{98.7}&97.6&97.9&92.9&98.5&\textbf{98.7}\\
    &Truck&76.0&48.5&73.1&55.4&76.2&91.3&88.8&92.1&92.0&70.0&\textbf{98.5}&97.4&95.5&94.4&75.2&98.4\\
    \hline
    &\thead{\textit{\tiny{Ave:}}}&64.8&58.8&64.8&65.7&74.2&86.0&83.3&87.3&88.2&72.5&95.6&96.2&94.3&90.0&84.3&\textbf{97.5}\\
    \bottomrule
\end{tabular}
\end{center}
\end{table*}

\begin{figure*}[!t]
\centering
\includegraphics[width=0.9\linewidth]{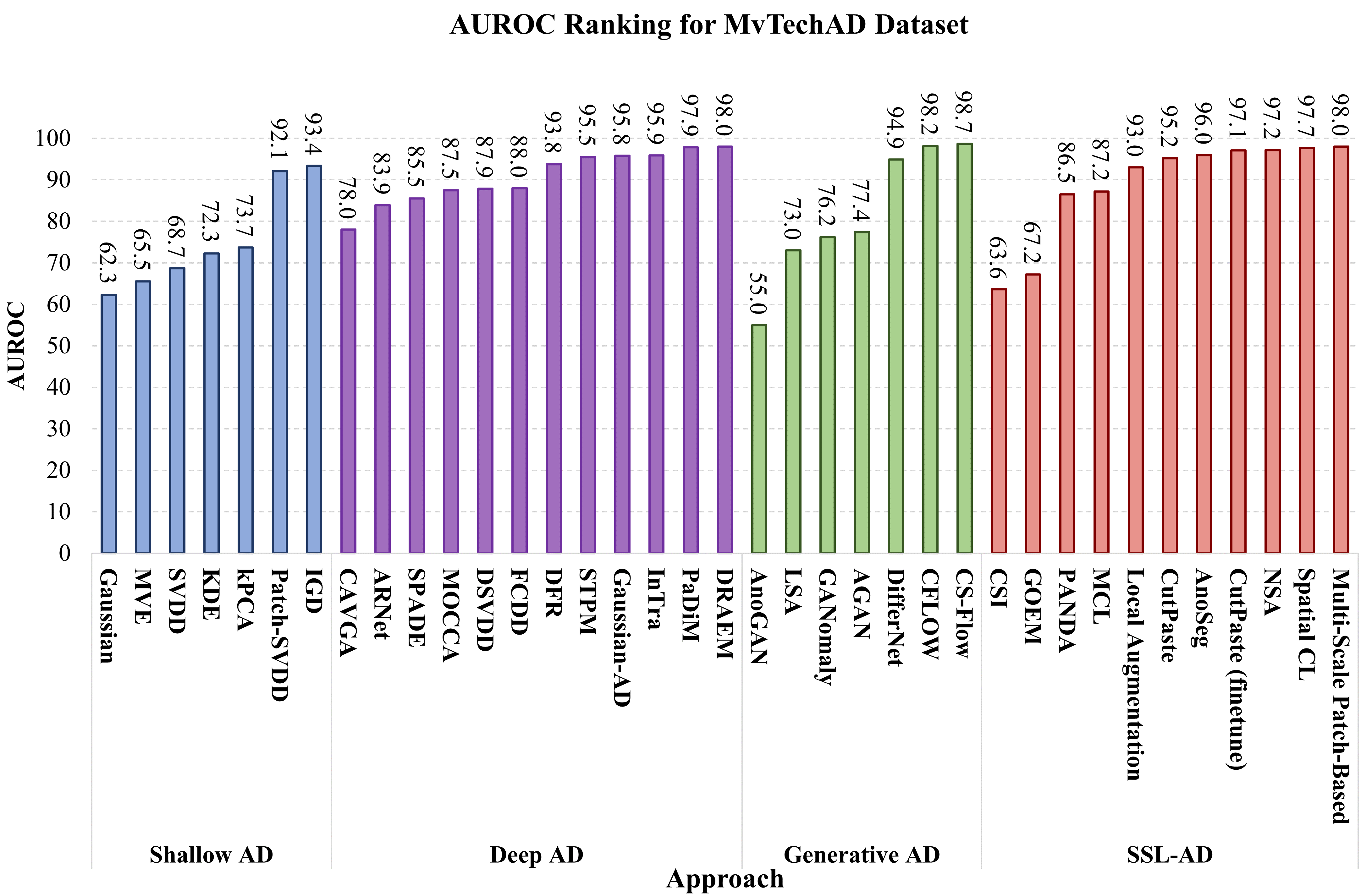}
\caption{Performance of anomaly detection algorithms on the MVTecAD dataset. Each group of algorithm is denoted by a different colour.}
\label{MVTEC}
\end{figure*}

Besides semantic anomaly detection, self-supervised methods show satisfactory performance for defect detection and spotting sensory anomalies \citep{song2021anoseg, tsai2022multi, kim2022spatial}. Fig.~\ref{MVTEC} shows the performance of the self-supervised models on the MVTecAD dataset against other widely-used algorithms including shallow models, deep models and generative models. More specifically, the compared shallow models are Gaussian \citep{ruff2021unifying}, MVE \citep{ruff2021unifying}, SVDD \citep{tax2004support}, KDE \citep{ruff2021unifying}, kPCA \citep{ruff2021unifying}, patch-SVDD \citep{yi2020patch} and IGD \citep{chen2021unsupervised}. The compared deep models are CAVGA \citep{venkataramanan2020attention}, ARNet \citep{fei2020attribute}, SPADE \citep{cohen2020sub}, MOCCA \citep{valerio2020mocca}, DSVDD \citep{ruff2018deep}, FCDD \citep{liznerski2020explainable}, DFR \citep{shi2021unsupervised}, STFPM \citep{wang2021student}, Gaussian-AD \citep{rippel2021modeling}, InTra \citep{pirnay2021inpainting}, PaDiM \citep{defard2021padim} and DREAM \citep{zavrtanik2021draem}. The included generative models in Fig.~\ref{MVTEC} are AnoGAN \citep{schlegl2017unsupervised}, LSA \citep{abati2019latent}, GANomaly \citep{akcay2018ganomaly}, AGAN \citep{ruff2021unifying}, Normalizing Flows-based DifferNet \citep{rudolph2021same}, CFLOW \citep{gudovskiy2022cflow} and CS-Flow \citep{rudolph2022fully}. Looking at the figure, we can infer that SSL-based models can achieve a good performance on this dataset. However, the superiority of self-supervised algorithms over other baselines is less evident in this task than in one-class AD. Also, some algorithms such as GEOM, and CSI, which show state-of-the-art performance on CIFAR-10, achieve a weak accuracy in this anomaly detection task.

The above argument manifests the importance of choosing the right pretext task in self-supervised learning. Methods such as GEOM and RotNet which are based on geometric transformations, and CSI and SSD which are based on contrastive methods, work well for detecting semantic anomalies, but they are not well-suited for defect detection. On the other hand, SSL approaches that are based on pixel-level transformations, such as CutPaste, can achieve good accuracy on the MVTecAD dataset. Choosing the right proxy task, depending on the downstream objective and types of anomalies, is the key to the success of the SSL models. This allows researchers to improve the state-of-the-art by coming up with effective pretext tasks. 

Out-of-distribution detection is another task in which SSL models are widely applied. Table \ref{oodtable} shows the experimental results of some SSL models (shown in the top 10 rows) against a supervised method, shown in the last row of the table. The supervised method is in fact a ResNet-50 network that is trained to classify the data available in CIFAR-10 from the other OOD dataset -- i.e., ResNet-50 is trained as an eleven-way classifier, ten for CIFAR-10 and one for the OOD dataset. To benchmark an OOD algorithm, it is common to train a model on the CIFAR-10 dataset and test the model using another dataset. If the samples of the test datasets are similar to the CIFAR-10 to some extent, the task is called near-OOD detection (e.g. CIFAR-10 vs. CIFAR-100). Otherwise, it is referred to as far-OOD detection (e.g. CIFAR-10 vs. SVHN, or CIFAR-10 vs. LSUN). We observe that SSL can even achieve better performance than the supervised baseline. This manifests that it is not necessary to have access to the data ground truths for the OOD detection task.

\begin{table}[t]
\scriptsize
\caption{Performance of SSL models against a supervise-based method for OOD detection. The bold values denote the highest AUROC ($\%$) result for each OOD dataset}
\setlength{\tabcolsep}{2pt}
\label{oodtable}
\begin{center}
\begin{tabular}{ccccccc}
    \toprule 
    & \multicolumn{3}{c}{IND: CIFAR-10} & \multicolumn{3}{c}{IND: CIFAR-100} \\ 
    \cmidrule(lr){2-4}\cmidrule(lr){5-7}
    \textbf{Method} & \multicolumn{3}{c}{OOD:} & \multicolumn{3}{c}{OOD:} \\
    & \makecell{CIFAR-100} & \makecell{SVHN} & \makecell{LSUN} & \makecell{CIFAR-10} & \makecell{SVHN} & \makecell{LSUN} \\
    \midrule
    RotNet \citep{hendrycks2018deep}&93.3&94.4&\textbf{97.6}&75.7&86.9&\textbf{93.4}\\
    CSI \citep{tack2020csi}&89.2&99.8&90.3&-&-&-\\
    SSL-OE \citep{hendrycks2019using}&93.3&98.4&93.2&-&-&-\\
    CLP \citep{Winkens}&92.9&99.5&-&\textbf{78.9}&95.4&-\\
    SSL-OOD \citep{Mohseni_Pitale_Yadawa_Wang_2020}&93.8&99.2&98.9&77.7&95.8&88.9\\
    MCL \citep{mcl}&90.8&97.9&93.8&-&-&-\\
    MCL-SEI \citep{mcl}&94.0&99.3&96.3&-&-&-\\
    SSD \citep{sehwag2021ssd}&90.6&99.6&96.5&69.6&94.9&79.5\\
    $\mathrm{SSD}_k$ (k=5) \citep{sehwag2021ssd}&93.1&99.7&97.8&78.3&\textbf{99.1}&93.4\\
    SDNS \citep{rafiee2022self}&\textbf{94.2}&\textbf{99.9}&97.5&67.6&97.2&74.6\\
    \hline
    ResNet-50 \citep{sehwag2021ssd}&90.6&99.6&93.8&55.3&94.5&69.4\\
    \bottomrule
    
\end{tabular}
\end{center}
\end{table}

The results reported in Table \ref{oodtable} shows that all the SSL-based methods can achieve an accuracy above $94 \%$ on far-OOD detection (i.e. CIFAR-10 vs. SVHN). It can suggest that the SSL models can learn meaningful features of the dataset. Almost all algorithms perform well in near-OOD detection, and some can even beat the supervised baseline.

\begin{figure*}[t]
  \centering
  \includegraphics[width=1\linewidth]{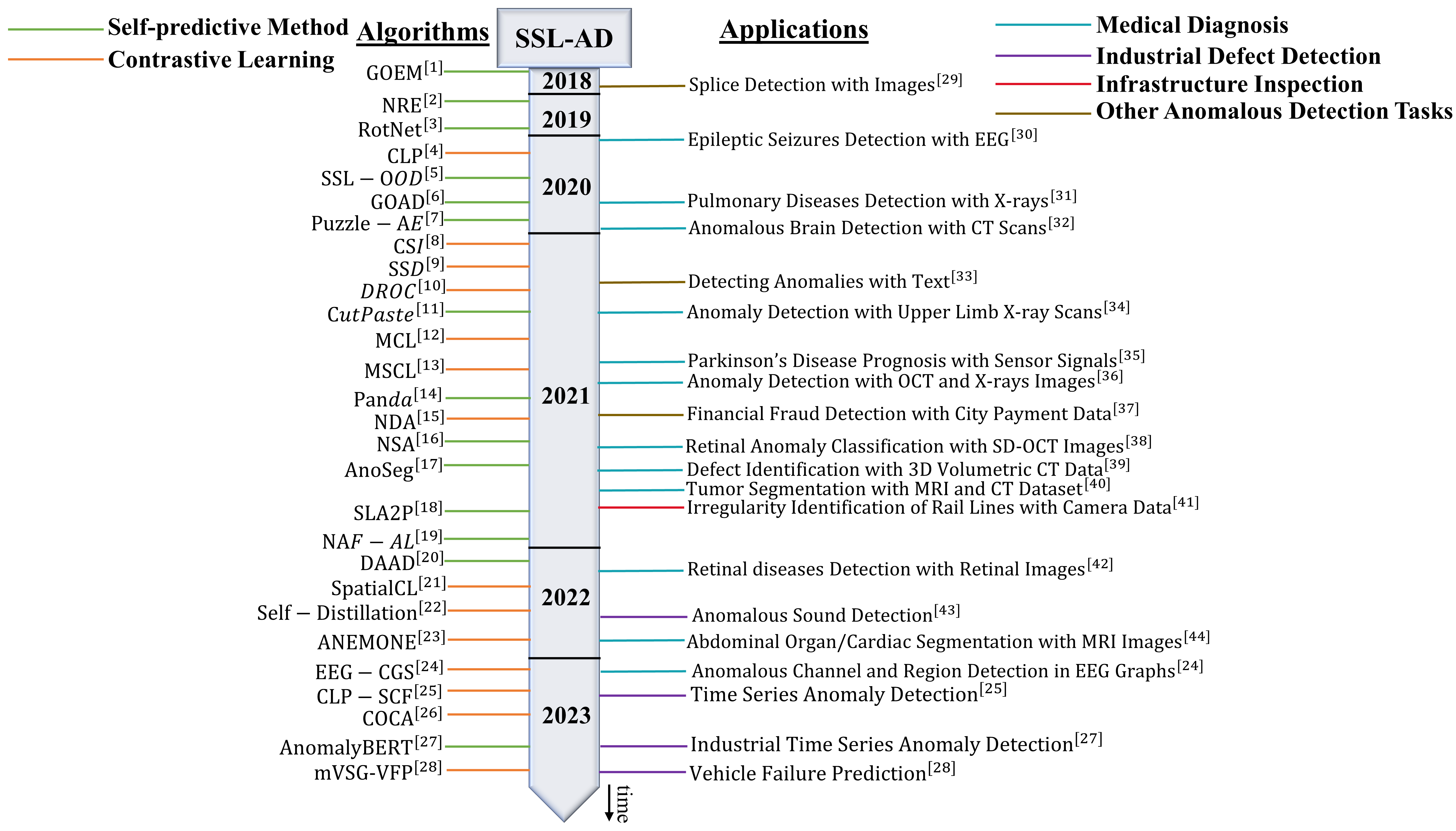}
  \caption{Timeline of Self-Supervised Anomaly Detection Papers. Papers concerning the algorithms are distinguished from the application papers. The category of each algorithm is denoted by a distinctive color.}
  \floatfoot{\scriptsize{$^{[1]}$\citep{golan2018deep}; $^{[2]}$\citep{Sabokrou_2019_ICCV}; $^{[3]}$\citep{hendrycks2018deep}; $^{[4]}$\citep{Winkens}; $^{[5]}$\citep{Mohseni_Pitale_Yadawa_Wang_2020}; $^{[6]}$\citep{bergman2020classification}; $^{[7]}$\citep{salehi2020puzzleae}; $^{[8]}$\citep{tack2020csi}; $^{[9]}$\citep{sehwag2021ssd}; $^{[10]}$\citep{sohn2020learning}; $^{[11]}$\citep{li2021cutpaste}; $^{[12]}$\citep{mcl}; $^{[13]}$\citep{reiss2021mean}; $^{[14]}$\citep{reiss2021panda}; $^{[15]}$\citep{chennegative}; $^{[16]}$\citep{schluter2021self}; $^{[17]}$\citep{song2021anoseg}; $^{[18]}$\citep{slap}; $^{[19]}$\citep{zhang2021selfsupervised}; $^{[20]}$\citep{ZHANG2022108234}; $^{[21]}$\citep{kim2022spatial}; $^{[22]}$\citep{rafiee2022self}; $^{[23]}$\citep{zheng2022unsupervised}; $^{[24]}$\citep{ho2022self}; $^{[25]}$\citep{wang2023deep}; $^{[26]}$\citep{icasp232}; $^{[27]}$\citep{jeong2023anomalybert}; $^{[28]}$\citep{hadi2023vehicle}; $^{[29]}$\citep{huh2018fighting}; $^{[30]}$\citep{xu2020anomaly}; $^{[31]}$\citep{bozorgtabar2020salad}; $^{[32]}$\citep{venkatakrishnan2020self}; $^{[33]}$\citep{manolache-etal-2021-date}; $^{[34]}$\citep{spahr2021self}; $^{[35]}$\citep{jiang2021towards}; $^{[36]}$\citep{zhao2021anomaly}; $^{[37]}$\citep{schreyer2021multi}; $^{[38]}$\citep{park2021self}; $^{[39]}$\citep{cho2021self}; $^{[40]}$\citep{zhang2021self}; $^{[41]}$\citep{jahan2021anomaly}; $^{[42]}$\citep{burlina2022detecting}; $^{[43]}$\citep{BAI2023103939}; $^{[44]}$\citep{hansen2022anomaly}}.}
  \label{fig:overview}
\end{figure*}

\section{Application Domains}

Anomaly detection systems are widely deployed in various domains, such as medicine, industry, infrastructure, social medical, financial security, etc. Despite the fact that self-supervised anomaly detection is a relatively new field, it is now widely employed in practical applications along with other popular methods such as Semi-supervised learning \citep{SemisupervisedSurvey}, and GAN and its variants \citep{GanSurveyXia}.

Self-supervised learning algorithms are commonly used in medical research for detecting irregularities in patients' records. They are successfully employed for detecting epileptic seizures \citep{xu2020anomaly}, pulmonary diseases \citep{bozorgtabar2020salad}, Parkinson disease \citep{jiang2021towards}, and retinal diseases\citep{burlina2022detecting}. In addition, they are applied to different modalities of medical data, including Computed Tomography (CT) scans \citep{venkatakrishnan2020self}, 3D volumetric CT data \citep{cho2021self}, X-ray scans \citep{spahr2021self}, optical coherence tomography (OCT) \citep{zhao2021anomaly}, Spectral Domain - optical coherence tomography images (SD-OCT) \citep{park2021self}, and MRI images \citep{zhang2021self,hansen2022anomaly}.

Self-supervised anomaly detection method are also employed in industrial applications for defect detection, and failure prediction \citep{bahavan2020anomaly, hou2021self}, as well as for monitoring infrastructural facilities \citep{liu2021deepfib,jahan2021anomaly}.

The application of self-supervised AD is not limited to the aforementioned areas. Several fields such as financial fraud detection \citep{schreyer2021multi,wang2021deep}, text anomaly detection \citep{manolache2021date}, and splice detection \citep{huh2018fighting} are also benefited from the SSL algorithms.

Figure \ref{fig:overview} depicts the timeline of papers focusing on self-supervised anomaly detection algorithms and their applications. This figure highlights the rapid growth of this field and its wide applicability in addressing real-world problems.

\section{Future Directions}

Although the self-supervised models have established themselves as state-of-the-art in anomaly detection, there is still much room for improvement in this research field. This section briefly discusses some critical challenges that SSL-based anomaly detectors suffer from and presents some high-level ideas for addressing them.

\subsection{Negative Sampling in Contrastive Models}

In recent years, contrastive models dominated self-supervised AD algorithms. To learn an efficient representation, CL algorithms require accessing negative samples. In the standard setting, it is assumed that other batch samples are negative, even though their class label is the same as that of the query sample. However, if the number of same-class samples increases, the quality of the learned representation degrades. In some anomaly detection tasks, where the training data comprises samples of one class, this negative sampling bias may turn into a big issue. This motivates researchers to design unbiased versions of the contrastive loss \citep{chuang2020debiased}.

Interestingly, previous studies showed that even in the one-class setting, the instance discrimination contrastive learning can lead to a suitable representation for anomalies. This can be because all the training data are spread out on a hypersphere, and the anomalies are mapped to the center of the space, as we discussed in section \ref{sec:cl}.

Following the success of SimCLR, several other contrastive models are developed. These methods can be good candidates for one-class anomaly detection since they can be trained using only positive samples. Some recent models, such as BYOL \citep{byol} and Barlow Twins \citep{barlow}, do not require negative samples during training. To the best of our knowledge, there is no study that evaluates the performance of these models for anomaly detection.

\subsection{Incorporating Labelled Data}

In the most anomaly detection studies, it is assumed that no labeled anomaly is available during the training phase. However, in some applications, we might be able to have a few labelled anomalies. These labelled samples can significantly improve the algorithm if incorporated appropriately. Recently, \citet{sehwag2021ssd} explored the problem of few-shot anomaly detection, where they assume a few labelled anomalies are present. They showed that even a few anomalies can significantly improve the detection accuracy. \citet{zheng2022unsupervised} proposed an extended
algorithm of multi-scale contrastive learning, called ANEMONE, by incorporating it with a handful of ground-truth anomalies. Since the assumption of having access to a few anomaly samples during training time is feasible in many tasks, we believe that models with the capability to incorporate them have a great potential to improve the detection performance. Such methods also have more application in real-life problems.



\subsubsection{Multi-Modal Anomaly Detection}
In many applications, including medical imaging, cybersecurity, and surveillance systems, the datasets contain multiple sources of information or modalities. Detecting anomalies in such cases heavily depends on the quality and relevance of the information contained in each modality and the ability to effectively fuse this information to make a robust decision \citep{mml}. Since self-supervised methods have already established themselves as powerful tools for learning representations, it would be interesting to study their application in multi-modal learning for anomaly detection. To this end, researchers might pursue the direction of designing cross-modal proxy tasks that aid the model to fuse information from different modalities in an efficient manner.

\subsubsection{Efficient Self-Supervised Learning}
Currently, self-supervised models have shown superior performance over traditional algorithms. Yet, they face critical challenges such as their computational cost, which prevents their widespread use in many applications. Future research in self-supervised learning will likely focus on designing computationally effective models that can leverage the vast amounts of unannotated data available for training. Additionally, the use of transfer learning, meta-learning, and federated learning may become more widespread as a way to overcome the limitations of self-supervised algorithms and enable their deployment in resource-constrained environments. Furthermore, research may also investigate the scalability of self-supervised learning to handle large amounts of data and diverse domains, as well as its interpretability and robustness to adversarial attacks.

\section{Conclusion}

In this paper, we discussed the state-of-the-art methods in self-supervised anomaly detection and highlighted the strengths and drawbacks of each approach. We also compared their performance on benchmark datasets and pinpointed their applications. In summary, we can argue that self-supervised models are well suited for tackling the problem of anomaly detection. Yet, there are still a lot of under-explored issues and room for improvement. Still, the significant success of SSL algorithms offers a bright horizon for achieving new milestones in automatic anomaly detection.

\section*{Acknowledgments}
The  authors  wish  to  acknowledge  the financial  support of  the  Natural  Sciences, Engineering  Research  Council of  Canada  (NSERC), Fonds de recherche du Québec (FRQNT), AGE-WELL, and the Department of Electrical and Computer Engineering at McGill University. This  research  was  enabled  in  part  by support provided by Calcul Quebec and Compute Canada.



\bibliographystyle{elsarticle-harv} 
\bibliography{cas-refs}

\begin{thebibliography}{137}
\expandafter\ifx\csname natexlab\endcsname\relax\def\natexlab#1{#1}\fi
\providecommand{\url}[1]{\texttt{#1}}
\providecommand{\href}[2]{#2}
\providecommand{\path}[1]{#1}
\providecommand{\DOIprefix}{doi:}
\providecommand{\ArXivprefix}{arXiv:}
\providecommand{\URLprefix}{URL: }
\providecommand{\Pubmedprefix}{pmid:}
\providecommand{\doi}[1]{\href{http://dx.doi.org/#1}{\path{#1}}}
\providecommand{\Pubmed}[1]{\href{pmid:#1}{\path{#1}}}
\providecommand{\bibinfo}[2]{#2}
\ifx\xfnm\relax \def\xfnm[#1]{\unskip,\space#1}\fi
\bibitem[{Abati et~al.(2019)Abati, Porrello, Calderara and
  Cucchiara}]{abati2019latent}
\bibinfo{author}{Abati, D.}, \bibinfo{author}{Porrello, A.},
  \bibinfo{author}{Calderara, S.}, \bibinfo{author}{Cucchiara, R.},
  \bibinfo{year}{2019}.
\newblock \bibinfo{title}{Latent space autoregression for novelty detection},
  in: \bibinfo{booktitle}{Proceedings of the IEEE/CVF Conference on Computer
  Vision and Pattern Recognition}, pp. \bibinfo{pages}{481--490}.
\bibitem[{Agyemang et~al.(2006)Agyemang, Barker and Alhajj}]{surveyagyemang}
\bibinfo{author}{Agyemang, M.}, \bibinfo{author}{Barker, K.},
  \bibinfo{author}{Alhajj, R.}, \bibinfo{year}{2006}.
\newblock \bibinfo{title}{A comprehensive survey of numeric and symbolic
  outlier mining techniques}.
\newblock \bibinfo{journal}{Intell. Data Anal.} \bibinfo{volume}{10},
  \bibinfo{pages}{521--538}.
\newblock \DOIprefix\doi{10.3233/IDA-2006-10604}.
\bibitem[{Akcay et~al.(2018)Akcay, Atapour-Abarghouei and
  Breckon}]{akcay2018ganomaly}
\bibinfo{author}{Akcay, S.}, \bibinfo{author}{Atapour-Abarghouei, A.},
  \bibinfo{author}{Breckon, T.P.}, \bibinfo{year}{2018}.
\newblock \bibinfo{title}{Ganomaly: Semi-supervised anomaly detection via
  adversarial training}, in: \bibinfo{booktitle}{Asian conference on computer
  vision}, \bibinfo{organization}{Springer}. pp. \bibinfo{pages}{622--637}.
\bibitem[{Azizi et~al.(2021)Azizi, Mustafa, Ryan, Beaver, Freyberg, Deaton,
  Loh, Karthikesalingam, Kornblith, Chen, Natarajan and Norouzi}]{healthSSL}
\bibinfo{author}{Azizi, S.}, \bibinfo{author}{Mustafa, B.},
  \bibinfo{author}{Ryan, F.}, \bibinfo{author}{Beaver, Z.},
  \bibinfo{author}{Freyberg, J.}, \bibinfo{author}{Deaton, J.},
  \bibinfo{author}{Loh, A.}, \bibinfo{author}{Karthikesalingam, A.},
  \bibinfo{author}{Kornblith, S.}, \bibinfo{author}{Chen, T.},
  \bibinfo{author}{Natarajan, V.}, \bibinfo{author}{Norouzi, M.},
  \bibinfo{year}{2021}.
\newblock \bibinfo{title}{Big self-supervised models advance medical image
  classification}, in: \bibinfo{booktitle}{2021 IEEE/CVF International
  Conference on Computer Vision (ICCV)}, pp. \bibinfo{pages}{3458--3468}.
\newblock \DOIprefix\doi{10.1109/ICCV48922.2021.00346}.
\bibitem[{Bahavan et~al.(2020)Bahavan, Suman, Cader, Ranganayake, Seneviratne,
  Maddumage, Seneviratne, Supun, Wijesiri, Dehigaspitiya
  et~al.}]{bahavan2020anomaly}
\bibinfo{author}{Bahavan, N.}, \bibinfo{author}{Suman, N.},
  \bibinfo{author}{Cader, S.}, \bibinfo{author}{Ranganayake, R.},
  \bibinfo{author}{Seneviratne, D.}, \bibinfo{author}{Maddumage, V.},
  \bibinfo{author}{Seneviratne, G.}, \bibinfo{author}{Supun, Y.},
  \bibinfo{author}{Wijesiri, I.}, \bibinfo{author}{Dehigaspitiya, S.}, et~al.,
  \bibinfo{year}{2020}.
\newblock \bibinfo{title}{Anomaly detection using deep reconstruction and
  forecasting for autonomous systems}.
\newblock \bibinfo{journal}{arXiv preprint arXiv:2006.14556} .
\bibitem[{Bai et~al.(2023)Bai, Chen, Wang, Ayub and Yan}]{BAI2023103939}
\bibinfo{author}{Bai, J.}, \bibinfo{author}{Chen, J.}, \bibinfo{author}{Wang,
  M.}, \bibinfo{author}{Ayub, M.S.}, \bibinfo{author}{Yan, Q.},
  \bibinfo{year}{2023}.
\newblock \bibinfo{title}{Ssdpt: Self-supervised dual-path transformer for
  anomalous sound detection}.
\newblock \bibinfo{journal}{Digital Signal Processing} \bibinfo{volume}{135},
  \bibinfo{pages}{103939}.
\newblock \URLprefix
  \url{https://www.sciencedirect.com/science/article/pii/S1051200423000349},
  \DOIprefix\doi{https://doi.org/10.1016/j.dsp.2023.103939}.
\bibitem[{Bergman and Hoshen(2020)}]{bergman2020classification}
\bibinfo{author}{Bergman, L.}, \bibinfo{author}{Hoshen, Y.},
  \bibinfo{year}{2020}.
\newblock \bibinfo{title}{Classification-based anomaly detection for general
  data}, in: \bibinfo{booktitle}{International Conference on Learning
  Representations (ICLR)}.
\bibitem[{Bergmann et~al.(2019)Bergmann, Fauser, Sattlegger and
  Steger}]{MVTECAD}
\bibinfo{author}{Bergmann, P.}, \bibinfo{author}{Fauser, M.},
  \bibinfo{author}{Sattlegger, D.}, \bibinfo{author}{Steger, C.},
  \bibinfo{year}{2019}.
\newblock \bibinfo{title}{Mvtec ad — a comprehensive real-world dataset for
  unsupervised anomaly detection}, in: \bibinfo{booktitle}{2019 IEEE/CVF
  Conference on Computer Vision and Pattern Recognition (CVPR)}, pp.
  \bibinfo{pages}{9584--9592}.
\newblock \DOIprefix\doi{10.1109/CVPR.2019.00982}.
\bibitem[{Bozorgtabar et~al.(2020)Bozorgtabar, Mahapatra, Vray and
  Thiran}]{bozorgtabar2020salad}
\bibinfo{author}{Bozorgtabar, B.}, \bibinfo{author}{Mahapatra, D.},
  \bibinfo{author}{Vray, G.}, \bibinfo{author}{Thiran, J.P.},
  \bibinfo{year}{2020}.
\newblock \bibinfo{title}{Salad: Self-supervised aggregation learning for
  anomaly detection on x-rays}, in: \bibinfo{booktitle}{International
  Conference on Medical Image Computing and Computer-Assisted Intervention},
  \bibinfo{organization}{Springer}. pp. \bibinfo{pages}{468--478}.
\bibitem[{Burlina et~al.(2022)Burlina, Paul, Liu and
  Bressler}]{burlina2022detecting}
\bibinfo{author}{Burlina, P.}, \bibinfo{author}{Paul, W.},
  \bibinfo{author}{Liu, T.A.}, \bibinfo{author}{Bressler, N.M.},
  \bibinfo{year}{2022}.
\newblock \bibinfo{title}{Detecting anomalies in retinal diseases using
  generative, discriminative, and self-supervised deep learning}.
\newblock \bibinfo{journal}{JAMA ophthalmology} \bibinfo{volume}{140},
  \bibinfo{pages}{185--189}.
\bibitem[{Carmona et~al.(2022)Carmona, Aubet, Flunkert and
  Gasthaus}]{neuralcontext}
\bibinfo{author}{Carmona, C.U.}, \bibinfo{author}{Aubet, F.X.},
  \bibinfo{author}{Flunkert, V.}, \bibinfo{author}{Gasthaus, J.},
  \bibinfo{year}{2022}.
\newblock \bibinfo{title}{Neural contextual anomaly detection for time series},
  in: \bibinfo{editor}{Raedt, L.D.} (Ed.), \bibinfo{booktitle}{Proceedings of
  the Thirty-First International Joint Conference on Artificial Intelligence,
  {IJCAI-22}}, \bibinfo{publisher}{International Joint Conferences on
  Artificial Intelligence Organization}. pp. \bibinfo{pages}{2843--2851}.
\newblock \URLprefix \url{https://doi.org/10.24963/ijcai.2022/394},
  \DOIprefix\doi{10.24963/ijcai.2022/394}. \bibinfo{note}{main Track}.
\bibitem[{Chalapathy and Chawla(2019)}]{SurveyChalapathy}
\bibinfo{author}{Chalapathy, R.}, \bibinfo{author}{Chawla, S.},
  \bibinfo{year}{2019}.
\newblock \bibinfo{title}{Deep learning for anomaly detection: A survey}.
\newblock \URLprefix \url{https://arxiv.org/abs/1901.03407},
  \DOIprefix\doi{10.48550/ARXIV.1901.03407}.
\bibitem[{Chandola et~al.(2009)Chandola, Banerjee and
  Kumar}]{chandola2007outlier}
\bibinfo{author}{Chandola, V.}, \bibinfo{author}{Banerjee, A.},
  \bibinfo{author}{Kumar, V.}, \bibinfo{year}{2009}.
\newblock \bibinfo{title}{Outlier detection: A survey}.
\newblock \bibinfo{journal}{ACM Computing Surveys} \bibinfo{volume}{14},
  \bibinfo{pages}{15}.
\bibitem[{Chen et~al.(2022)Chen, Zhang, Zhang, Dong, Song, Zhang, Xu, Kharlamov
  and Tang}]{chen2022gccad}
\bibinfo{author}{Chen, B.}, \bibinfo{author}{Zhang, J.},
  \bibinfo{author}{Zhang, X.}, \bibinfo{author}{Dong, Y.},
  \bibinfo{author}{Song, J.}, \bibinfo{author}{Zhang, P.}, \bibinfo{author}{Xu,
  K.}, \bibinfo{author}{Kharlamov, E.}, \bibinfo{author}{Tang, J.},
  \bibinfo{year}{2022}.
\newblock \bibinfo{title}{Gccad: Graph contrastive learning for anomaly
  detection}.
\newblock \bibinfo{journal}{IEEE Transactions on Knowledge and Data
  Engineering} .
\bibitem[{Chen et~al.(2021a)Chen, Xie, Lin, Qiao, Zhou, Tan, Zhang and
  Ma}]{chennegative}
\bibinfo{author}{Chen, C.}, \bibinfo{author}{Xie, Y.}, \bibinfo{author}{Lin,
  S.}, \bibinfo{author}{Qiao, R.}, \bibinfo{author}{Zhou, J.},
  \bibinfo{author}{Tan, X.}, \bibinfo{author}{Zhang, Y.}, \bibinfo{author}{Ma,
  L.}, \bibinfo{year}{2021}a.
\newblock \bibinfo{title}{Novelty detection via contrastive learning with
  negative data augmentation}, in: \bibinfo{editor}{Zhou, Z.H.} (Ed.),
  \bibinfo{booktitle}{Proceedings of the Thirtieth International Joint
  Conference on Artificial Intelligence, {IJCAI-21}},
  \bibinfo{publisher}{International Joint Conferences on Artificial
  Intelligence Organization}. pp. \bibinfo{pages}{606--614}.
\newblock \URLprefix \url{https://doi.org/10.24963/ijcai.2021/84},
  \DOIprefix\doi{10.24963/ijcai.2021/84}. \bibinfo{note}{main Track}.
\bibitem[{Chen et~al.(2020)Chen, Kornblith, Norouzi and
  Hinton}]{chen2020simple}
\bibinfo{author}{Chen, T.}, \bibinfo{author}{Kornblith, S.},
  \bibinfo{author}{Norouzi, M.}, \bibinfo{author}{Hinton, G.},
  \bibinfo{year}{2020}.
\newblock \bibinfo{title}{A simple framework for contrastive learning of visual
  representations}, in: \bibinfo{booktitle}{International conference on machine
  learning}, \bibinfo{organization}{PMLR}. pp. \bibinfo{pages}{1597--1607}.
\bibitem[{Chen et~al.(2021b)Chen, Tian, Pang and
  Carneiro}]{chen2021unsupervised}
\bibinfo{author}{Chen, Y.}, \bibinfo{author}{Tian, Y.}, \bibinfo{author}{Pang,
  G.}, \bibinfo{author}{Carneiro, G.}, \bibinfo{year}{2021}b.
\newblock \bibinfo{title}{Unsupervised anomaly detection with multi-scale
  interpolated gaussian descriptors}.
\newblock \bibinfo{journal}{arXiv preprint arXiv:2101.10043}
  \bibinfo{volume}{2}.
\bibitem[{Cho et~al.(2021a)Cho, Seol and Lee}]{mcl}
\bibinfo{author}{Cho, H.}, \bibinfo{author}{Seol, J.}, \bibinfo{author}{Lee,
  S.g.}, \bibinfo{year}{2021}a.
\newblock \bibinfo{title}{Masked contrastive learning for anomaly detection},
  in: \bibinfo{editor}{Zhou, Z.H.} (Ed.), \bibinfo{booktitle}{Proceedings of
  the Thirtieth International Joint Conference on Artificial Intelligence,
  {IJCAI-21}}, \bibinfo{publisher}{International Joint Conferences on
  Artificial Intelligence Organization}. pp. \bibinfo{pages}{1434--1441}.
\newblock \URLprefix \url{https://doi.org/10.24963/ijcai.2021/198},
  \DOIprefix\doi{10.24963/ijcai.2021/198}. \bibinfo{note}{main Track}.
\bibitem[{Cho et~al.(2021b)Cho, Kang and Park}]{cho2021self}
\bibinfo{author}{Cho, J.}, \bibinfo{author}{Kang, I.}, \bibinfo{author}{Park,
  J.}, \bibinfo{year}{2021}b.
\newblock \bibinfo{title}{Self-supervised 3d out-of-distribution detection via
  pseudoanomaly generation}, in: \bibinfo{booktitle}{International Conference
  on Medical Image Computing and Computer-Assisted Intervention},
  \bibinfo{organization}{Springer}. pp. \bibinfo{pages}{95--103}.
\bibitem[{Chopra et~al.(2005)Chopra, Hadsell and LeCun}]{conloss}
\bibinfo{author}{Chopra, S.}, \bibinfo{author}{Hadsell, R.},
  \bibinfo{author}{LeCun, Y.}, \bibinfo{year}{2005}.
\newblock \bibinfo{title}{Learning a similarity metric discriminatively, with
  application to face verification}, in: \bibinfo{booktitle}{2005 IEEE Computer
  Society Conference on Computer Vision and Pattern Recognition (CVPR'05)}, pp.
  \bibinfo{pages}{539--546 vol. 1}.
\newblock \DOIprefix\doi{10.1109/CVPR.2005.202}.
\bibitem[{Chuang et~al.(2020)Chuang, Robinson, Lin, Torralba and
  Jegelka}]{chuang2020debiased}
\bibinfo{author}{Chuang, C.Y.}, \bibinfo{author}{Robinson, J.},
  \bibinfo{author}{Lin, Y.C.}, \bibinfo{author}{Torralba, A.},
  \bibinfo{author}{Jegelka, S.}, \bibinfo{year}{2020}.
\newblock \bibinfo{title}{Debiased contrastive learning}.
\newblock \bibinfo{journal}{Advances in Neural Information Processing Systems}
  \bibinfo{volume}{33}.
\bibitem[{Cohen and Hoshen(2020)}]{cohen2020sub}
\bibinfo{author}{Cohen, N.}, \bibinfo{author}{Hoshen, Y.},
  \bibinfo{year}{2020}.
\newblock \bibinfo{title}{Sub-image anomaly detection with deep pyramid
  correspondences}.
\newblock \bibinfo{journal}{arXiv preprint arXiv:2005.02357} .
\bibitem[{Defard et~al.(2021)Defard, Setkov, Loesch and
  Audigier}]{defard2021padim}
\bibinfo{author}{Defard, T.}, \bibinfo{author}{Setkov, A.},
  \bibinfo{author}{Loesch, A.}, \bibinfo{author}{Audigier, R.},
  \bibinfo{year}{2021}.
\newblock \bibinfo{title}{Padim: a patch distribution modeling framework for
  anomaly detection and localization}, in: \bibinfo{booktitle}{International
  Conference on Pattern Recognition}, \bibinfo{organization}{Springer}. pp.
  \bibinfo{pages}{475--489}.
\bibitem[{Di~Mattia et~al.(2019)Di~Mattia, Galeone, De~Simoni and
  Ghelfi}]{GanSurveyDimattia}
\bibinfo{author}{Di~Mattia, F.}, \bibinfo{author}{Galeone, P.},
  \bibinfo{author}{De~Simoni, M.}, \bibinfo{author}{Ghelfi, E.},
  \bibinfo{year}{2019}.
\newblock \bibinfo{title}{A survey on gans for anomaly detection}.
\newblock \URLprefix \url{https://arxiv.org/abs/1906.11632},
  \DOIprefix\doi{10.48550/ARXIV.1906.11632}.
\bibitem[{Doersch et~al.(2015)Doersch, Gupta and
  Efros}]{doersch2015unsupervised}
\bibinfo{author}{Doersch, C.}, \bibinfo{author}{Gupta, A.},
  \bibinfo{author}{Efros, A.A.}, \bibinfo{year}{2015}.
\newblock \bibinfo{title}{Unsupervised visual representation learning by
  context prediction}, in: \bibinfo{booktitle}{Proceedings of the IEEE
  international conference on computer vision}, pp.
  \bibinfo{pages}{1422--1430}.
\bibitem[{Duan et~al.(2022)Duan, Wang, Zhang, Zhu, Hu, Jin, Liu and
  Dong}]{duan2022graph}
\bibinfo{author}{Duan, J.}, \bibinfo{author}{Wang, S.}, \bibinfo{author}{Zhang,
  P.}, \bibinfo{author}{Zhu, E.}, \bibinfo{author}{Hu, J.},
  \bibinfo{author}{Jin, H.}, \bibinfo{author}{Liu, Y.}, \bibinfo{author}{Dong,
  Z.}, \bibinfo{year}{2022}.
\newblock \bibinfo{title}{Graph anomaly detection via multi-scale contrastive
  learning networks with augmented view}.
\newblock \href{http://arxiv.org/abs/2212.00535}{{\tt arXiv:2212.00535}}.
\bibitem[{Fawcett(2006)}]{auroc}
\bibinfo{author}{Fawcett, T.}, \bibinfo{year}{2006}.
\newblock \bibinfo{title}{An introduction to roc analysis}.
\newblock \bibinfo{journal}{Pattern Recogn. Lett.} \bibinfo{volume}{27},
  \bibinfo{pages}{861–874}.
\newblock \URLprefix \url{https://doi.org/10.1016/j.patrec.2005.10.010},
  \DOIprefix\doi{10.1016/j.patrec.2005.10.010}.
\bibitem[{Fei et~al.(2020)Fei, Huang, Jinkun, Li, Zhang and
  Lu}]{fei2020attribute}
\bibinfo{author}{Fei, Y.}, \bibinfo{author}{Huang, C.},
  \bibinfo{author}{Jinkun, C.}, \bibinfo{author}{Li, M.},
  \bibinfo{author}{Zhang, Y.}, \bibinfo{author}{Lu, C.}, \bibinfo{year}{2020}.
\newblock \bibinfo{title}{Attribute restoration framework for anomaly
  detection}.
\newblock \bibinfo{journal}{IEEE Transactions on Multimedia} .
\bibitem[{Feinman et~al.(2017)Feinman, Curtin, Shintre and
  Gardner}]{feinman2017detecting}
\bibinfo{author}{Feinman, R.}, \bibinfo{author}{Curtin, R.R.},
  \bibinfo{author}{Shintre, S.}, \bibinfo{author}{Gardner, A.B.},
  \bibinfo{year}{2017}.
\newblock \bibinfo{title}{Detecting adversarial samples from artifacts}.
\newblock \bibinfo{journal}{arXiv preprint arXiv:1703.00410} .
\bibitem[{Field et~al.(2004)Field, Tyre, Jonz{\'e}n, Rhodes and
  Possingham}]{field2004minimizing}
\bibinfo{author}{Field, S.A.}, \bibinfo{author}{Tyre, A.J.},
  \bibinfo{author}{Jonz{\'e}n, N.}, \bibinfo{author}{Rhodes, J.R.},
  \bibinfo{author}{Possingham, H.P.}, \bibinfo{year}{2004}.
\newblock \bibinfo{title}{Minimizing the cost of environmental management
  decisions by optimizing statistical thresholds}.
\newblock \bibinfo{journal}{Ecology Letters} \bibinfo{volume}{7},
  \bibinfo{pages}{669--675}.
\bibitem[{Fu and Xue(2022)}]{MAD}
\bibinfo{author}{Fu, Y.}, \bibinfo{author}{Xue, F.}, \bibinfo{year}{2022}.
\newblock \bibinfo{title}{Mad: Self-supervised masked anomaly detection task
  for multivariate time series}, in: \bibinfo{booktitle}{2022 International
  Joint Conference on Neural Networks (IJCNN)}, pp. \bibinfo{pages}{1--8}.
\newblock \DOIprefix\doi{10.1109/IJCNN55064.2022.9892218}.
\bibitem[{Gidaris et~al.(2018)Gidaris, Singh and
  Komodakis}]{gidaris2018unsupervised}
\bibinfo{author}{Gidaris, S.}, \bibinfo{author}{Singh, P.},
  \bibinfo{author}{Komodakis, N.}, \bibinfo{year}{2018}.
\newblock \bibinfo{title}{Unsupervised representation learning by predicting
  image rotations}.
\newblock \bibinfo{journal}{arXiv preprint arXiv:1803.07728} .
\bibitem[{Giri et~al.(2020)Giri, Tenneti, Cheng, Helwani, Isik and
  Krishnaswamy}]{Giri2020}
\bibinfo{author}{Giri, R.}, \bibinfo{author}{Tenneti, S.V.},
  \bibinfo{author}{Cheng, F.}, \bibinfo{author}{Helwani, K.},
  \bibinfo{author}{Isik, U.}, \bibinfo{author}{Krishnaswamy, A.},
  \bibinfo{year}{2020}.
\newblock \bibinfo{title}{Self-supervised classification for detecting
  anomalous sounds}, in: \bibinfo{booktitle}{Detection and Classification of
  Acoustic Scenes and Events Workshop 2020}.
\newblock \URLprefix
  \url{https://www.amazon.science/publications/self-supervised-classification-for-detecting-anomalous-sounds}.
\bibitem[{Golan and El-Yaniv(2018)}]{golan2018deep}
\bibinfo{author}{Golan, I.}, \bibinfo{author}{El-Yaniv, R.},
  \bibinfo{year}{2018}.
\newblock \bibinfo{title}{Deep anomaly detection using geometric
  transformations}.
\newblock \bibinfo{journal}{Advances in neural information processing systems}
  \bibinfo{volume}{31}.
\bibitem[{Grill et~al.(2020)Grill, Strub, Altché, Tallec, Richemond,
  Buchatskaya, Doersch, Pires, Guo, Azar, Piot, Kavukcuoglu, Munos and
  Valko}]{byol}
\bibinfo{author}{Grill, J.B.}, \bibinfo{author}{Strub, F.},
  \bibinfo{author}{Altché, F.}, \bibinfo{author}{Tallec, C.},
  \bibinfo{author}{Richemond, P.H.}, \bibinfo{author}{Buchatskaya, E.},
  \bibinfo{author}{Doersch, C.}, \bibinfo{author}{Pires, B.A.},
  \bibinfo{author}{Guo, Z.D.}, \bibinfo{author}{Azar, M.G.},
  \bibinfo{author}{Piot, B.}, \bibinfo{author}{Kavukcuoglu, K.},
  \bibinfo{author}{Munos, R.}, \bibinfo{author}{Valko, M.},
  \bibinfo{year}{2020}.
\newblock \bibinfo{title}{Bootstrap your own latent: A new approach to
  self-supervised learning}.
\newblock \href{http://arxiv.org/abs/2006.07733}{{\tt arXiv:2006.07733}}.
\bibitem[{Guan et~al.(2023)Guan, Xiao, Liu, Zhu and Wang}]{icasp232}
\bibinfo{author}{Guan, J.}, \bibinfo{author}{Xiao, F.}, \bibinfo{author}{Liu,
  Y.}, \bibinfo{author}{Zhu, Q.}, \bibinfo{author}{Wang, W.},
  \bibinfo{year}{2023}.
\newblock \bibinfo{title}{Anomalous sound detection using audio representation
  with machine id based contrastive learning pretraining}, in:
  \bibinfo{booktitle}{ICASSP 2023 - 2023 IEEE International Conference on
  Acoustics, Speech and Signal Processing (ICASSP)}, pp. \bibinfo{pages}{1--5}.
\newblock \DOIprefix\doi{10.1109/ICASSP49357.2023.10096054}.
\bibitem[{Gudovskiy et~al.(2022)Gudovskiy, Ishizaka and
  Kozuka}]{gudovskiy2022cflow}
\bibinfo{author}{Gudovskiy, D.}, \bibinfo{author}{Ishizaka, S.},
  \bibinfo{author}{Kozuka, K.}, \bibinfo{year}{2022}.
\newblock \bibinfo{title}{Cflow-ad: Real-time unsupervised anomaly detection
  with localization via conditional normalizing flows}, in:
  \bibinfo{booktitle}{Proceedings of the IEEE/CVF Winter Conference on
  Applications of Computer Vision}, pp. \bibinfo{pages}{98--107}.
\bibitem[{Hansen et~al.(2022)Hansen, Gautam, Jenssen and
  Kampffmeyer}]{hansen2022anomaly}
\bibinfo{author}{Hansen, S.}, \bibinfo{author}{Gautam, S.},
  \bibinfo{author}{Jenssen, R.}, \bibinfo{author}{Kampffmeyer, M.},
  \bibinfo{year}{2022}.
\newblock \bibinfo{title}{Anomaly detection-inspired few-shot medical image
  segmentation through self-supervision with supervoxels}.
\newblock \bibinfo{journal}{Medical Image Analysis} \bibinfo{volume}{78},
  \bibinfo{pages}{102385}.
\bibitem[{He et~al.(2020)He, Fan, Wu, Xie and Girshick}]{he2020momentum}
\bibinfo{author}{He, K.}, \bibinfo{author}{Fan, H.}, \bibinfo{author}{Wu, Y.},
  \bibinfo{author}{Xie, S.}, \bibinfo{author}{Girshick, R.},
  \bibinfo{year}{2020}.
\newblock \bibinfo{title}{Momentum contrast for unsupervised visual
  representation learning}, in: \bibinfo{booktitle}{Proceedings of the IEEE/CVF
  conference on computer vision and pattern recognition}, pp.
  \bibinfo{pages}{9729--9738}.
\bibitem[{Hendrycks et~al.(2018)Hendrycks, Mazeika and
  Dietterich}]{hendrycks2018deep}
\bibinfo{author}{Hendrycks, D.}, \bibinfo{author}{Mazeika, M.},
  \bibinfo{author}{Dietterich, T.}, \bibinfo{year}{2018}.
\newblock \bibinfo{title}{Deep anomaly detection with outlier exposure}, in:
  \bibinfo{booktitle}{International Conference on Learning Representations}.
\newblock \URLprefix \url{https://openreview.net/forum?id=HyxCxhRcY7}.
\bibitem[{Hendrycks et~al.(2019)Hendrycks, Mazeika, Kadavath and
  Song}]{hendrycks2019using}
\bibinfo{author}{Hendrycks, D.}, \bibinfo{author}{Mazeika, M.},
  \bibinfo{author}{Kadavath, S.}, \bibinfo{author}{Song, D.},
  \bibinfo{year}{2019}.
\newblock \bibinfo{title}{Using self-supervised learning can improve model
  robustness and uncertainty}.
\newblock \bibinfo{journal}{Advances in Neural Information Processing Systems}
  \bibinfo{volume}{32}.
\bibitem[{Hjelm et~al.(2019)Hjelm, Fedorov, Lavoie-Marchildon, Grewal, Bachman,
  Trischler and Bengio}]{hjelm2018learning}
\bibinfo{author}{Hjelm, R.D.}, \bibinfo{author}{Fedorov, A.},
  \bibinfo{author}{Lavoie-Marchildon, S.}, \bibinfo{author}{Grewal, K.},
  \bibinfo{author}{Bachman, P.}, \bibinfo{author}{Trischler, A.},
  \bibinfo{author}{Bengio, Y.}, \bibinfo{year}{2019}.
\newblock \bibinfo{title}{Learning deep representations by mutual information
  estimation and maximization}, in: \bibinfo{booktitle}{International
  Conference on Learning Representations}.
\newblock \URLprefix \url{https://openreview.net/forum?id=Bklr3j0cKX}.
\bibitem[{Ho and Armanfard(2023)}]{ho2022self}
\bibinfo{author}{Ho, T.K.K.}, \bibinfo{author}{Armanfard, N.},
  \bibinfo{year}{2023}.
\newblock \bibinfo{title}{Self-supervised learning for anomalous channel
  detection in eeg graphs: Application to seizure analysis}, in:
  \bibinfo{booktitle}{Proceedings of the AAAI Conference on Artificial
  Intelligence}.
\bibitem[{Ho et~al.(2023)Ho, Karami and Armanfard}]{ho2023graph}
\bibinfo{author}{Ho, T.K.K.}, \bibinfo{author}{Karami, A.},
  \bibinfo{author}{Armanfard, N.}, \bibinfo{year}{2023}.
\newblock \bibinfo{title}{Graph-based time-series anomaly detection: A survey}.
\newblock \bibinfo{journal}{arXiv preprint arXiv:2302.00058} .
\bibitem[{Hodge and Austin(2004)}]{surveyHodge}
\bibinfo{author}{Hodge, V.}, \bibinfo{author}{Austin, J.},
  \bibinfo{year}{2004}.
\newblock \bibinfo{title}{A survey of outlier detection methodologies}.
\newblock \bibinfo{journal}{Artificial Intelligence Review}
  \bibinfo{volume}{22}, \bibinfo{pages}{85--126}.
\newblock \DOIprefix\doi{10.1023/B:AIRE.0000045502.10941.a9}.
\bibitem[{Hojjati and Armanfard(2021)}]{hojjati2021dasvdd}
\bibinfo{author}{Hojjati, H.}, \bibinfo{author}{Armanfard, N.},
  \bibinfo{year}{2021}.
\newblock \bibinfo{title}{Dasvdd: Deep autoencoding support vector data
  descriptor for anomaly detection}, in: \bibinfo{booktitle}{arXiv}.
\newblock \href{http://arxiv.org/abs/2106.05410}{{\tt arXiv:2106.05410}}.
\bibitem[{Hojjati and Armanfard(2022)}]{AADCL}
\bibinfo{author}{Hojjati, H.}, \bibinfo{author}{Armanfard, N.},
  \bibinfo{year}{2022}.
\newblock \bibinfo{title}{Self-supervised acoustic anomaly detection via
  contrastive learning}, in: \bibinfo{booktitle}{ICASSP 2022 - 2022 IEEE
  International Conference on Acoustics, Speech and Signal Processing
  (ICASSP)}.
\bibitem[{Hojjati et~al.(2023)Hojjati, Sadeghi and Armanfard}]{hadi2023vehicle}
\bibinfo{author}{Hojjati, H.}, \bibinfo{author}{Sadeghi, M.},
  \bibinfo{author}{Armanfard, N.}, \bibinfo{year}{2023}.
\newblock \bibinfo{title}{Multivariate time-series anomaly detection with
  temporal self-supervision and graphs: Application to vehicle failure
  prediction}, in: \bibinfo{booktitle}{The European Conference on Machine
  Learning and Principles and Practice of Knowledge Discovery in Databases
  (ECML-PKDD)}.
\bibitem[{Hou et~al.(2021)Hou, Tao and Xu}]{hou2021self}
\bibinfo{author}{Hou, W.}, \bibinfo{author}{Tao, X.}, \bibinfo{author}{Xu, D.},
  \bibinfo{year}{2021}.
\newblock \bibinfo{title}{A self-supervised cnn for particle inspection on
  optical element}.
\newblock \bibinfo{journal}{IEEE Transactions on Instrumentation and
  Measurement} \bibinfo{volume}{70}, \bibinfo{pages}{1--12}.
\bibitem[{Huang et~al.(2022a)Huang, Shen, Yu, Zheng, Huang and
  Ma}]{HUANG2022261}
\bibinfo{author}{Huang, D.}, \bibinfo{author}{Shen, L.}, \bibinfo{author}{Yu,
  Z.}, \bibinfo{author}{Zheng, Z.}, \bibinfo{author}{Huang, M.},
  \bibinfo{author}{Ma, Q.}, \bibinfo{year}{2022}a.
\newblock \bibinfo{title}{Efficient time series anomaly detection by
  multiresolution self-supervised discriminative network}.
\newblock \bibinfo{journal}{Neurocomputing} \bibinfo{volume}{491},
  \bibinfo{pages}{261--272}.
\newblock \URLprefix
  \url{https://www.sciencedirect.com/science/article/pii/S0925231222003435},
  \DOIprefix\doi{https://doi.org/10.1016/j.neucom.2022.03.048}.
\bibitem[{Huang et~al.(2022b)Huang, Pei, Menkovski and
  Pechenizkiy}]{huang2022hop}
\bibinfo{author}{Huang, T.}, \bibinfo{author}{Pei, Y.},
  \bibinfo{author}{Menkovski, V.}, \bibinfo{author}{Pechenizkiy, M.},
  \bibinfo{year}{2022}b.
\newblock \bibinfo{title}{Hop-count based self-supervised anomaly detection on
  attributed networks}, in: \bibinfo{booktitle}{Joint European Conference on
  Machine Learning and Knowledge Discovery in Databases},
  \bibinfo{organization}{Springer}. pp. \bibinfo{pages}{225--241}.
\bibitem[{Huh et~al.(2018)Huh, Liu, Owens and Efros}]{huh2018fighting}
\bibinfo{author}{Huh, M.}, \bibinfo{author}{Liu, A.}, \bibinfo{author}{Owens,
  A.}, \bibinfo{author}{Efros, A.A.}, \bibinfo{year}{2018}.
\newblock \bibinfo{title}{Fighting fake news: Image splice detection via
  learned self-consistency}, in: \bibinfo{booktitle}{Proceedings of the
  European conference on computer vision (ECCV)}, pp.
  \bibinfo{pages}{101--117}.
\bibitem[{Jahan et~al.(2021)Jahan, Umesh and Roth}]{jahan2021anomaly}
\bibinfo{author}{Jahan, K.}, \bibinfo{author}{Umesh, J.P.},
  \bibinfo{author}{Roth, M.}, \bibinfo{year}{2021}.
\newblock \bibinfo{title}{Anomaly detection on the rail lines using semantic
  segmentation and self-supervised learning}, in: \bibinfo{booktitle}{2021 IEEE
  Symposium Series on Computational Intelligence (SSCI)},
  \bibinfo{organization}{IEEE}. pp. \bibinfo{pages}{1--7}.
\bibitem[{Jeong et~al.(2023)Jeong, Yang, Ryu, Park and
  Kang}]{jeong2023anomalybert}
\bibinfo{author}{Jeong, Y.}, \bibinfo{author}{Yang, E.}, \bibinfo{author}{Ryu,
  J.H.}, \bibinfo{author}{Park, I.}, \bibinfo{author}{Kang, M.},
  \bibinfo{year}{2023}.
\newblock \bibinfo{title}{Anomalybert: Self-supervised transformer for time
  series anomaly detection using data degradation scheme}.
\newblock \href{http://arxiv.org/abs/2305.04468}{{\tt arXiv:2305.04468}}.
\bibitem[{Jiang et~al.(2021)Jiang, Lim, Ng, Wang, Chi and
  Miao}]{jiang2021towards}
\bibinfo{author}{Jiang, H.}, \bibinfo{author}{Lim, W.Y.B.},
  \bibinfo{author}{Ng, J.S.}, \bibinfo{author}{Wang, Y.}, \bibinfo{author}{Chi,
  Y.}, \bibinfo{author}{Miao, C.}, \bibinfo{year}{2021}.
\newblock \bibinfo{title}{Towards parkinson’s disease prognosis using
  self-supervised learning and anomaly detection}, in:
  \bibinfo{booktitle}{ICASSP 2021-2021 IEEE International Conference on
  Acoustics, Speech and Signal Processing (ICASSP)},
  \bibinfo{organization}{IEEE}. pp. \bibinfo{pages}{3960--3964}.
\bibitem[{Jiao et~al.(2022)Jiao, Yang, Song and Tao}]{TimeAutoAD}
\bibinfo{author}{Jiao, Y.}, \bibinfo{author}{Yang, K.}, \bibinfo{author}{Song,
  D.}, \bibinfo{author}{Tao, D.}, \bibinfo{year}{2022}.
\newblock \bibinfo{title}{Timeautoad: Autonomous anomaly detection with
  self-supervised contrastive loss for multivariate time series}.
\newblock \bibinfo{journal}{IEEE Transactions on Network Science and
  Engineering} \bibinfo{volume}{9}, \bibinfo{pages}{1604--1619}.
\newblock \DOIprefix\doi{10.1109/TNSE.2022.3148276}.
\bibitem[{Jing and Tian(2021)}]{VisualSSL}
\bibinfo{author}{Jing, L.}, \bibinfo{author}{Tian, Y.}, \bibinfo{year}{2021}.
\newblock \bibinfo{title}{Self-supervised visual feature learning with deep
  neural networks: A survey}.
\newblock \bibinfo{journal}{IEEE Transactions on Pattern Analysis and Machine
  Intelligence} \bibinfo{volume}{43}, \bibinfo{pages}{4037--4058}.
\newblock \DOIprefix\doi{10.1109/TPAMI.2020.2992393}.
\bibitem[{Jumutc and Suykens(2014)}]{jumutc2014multi}
\bibinfo{author}{Jumutc, V.}, \bibinfo{author}{Suykens, J.A.},
  \bibinfo{year}{2014}.
\newblock \bibinfo{title}{Multi-class supervised novelty detection}.
\newblock \bibinfo{journal}{IEEE transactions on pattern analysis and machine
  intelligence} \bibinfo{volume}{36}, \bibinfo{pages}{2510--2523}.
\bibitem[{Khosla et~al.(2020)Khosla, Teterwak, Wang, Sarna, Tian, Isola,
  Maschinot, Liu and Krishnan}]{supcon}
\bibinfo{author}{Khosla, P.}, \bibinfo{author}{Teterwak, P.},
  \bibinfo{author}{Wang, C.}, \bibinfo{author}{Sarna, A.},
  \bibinfo{author}{Tian, Y.}, \bibinfo{author}{Isola, P.},
  \bibinfo{author}{Maschinot, A.}, \bibinfo{author}{Liu, C.},
  \bibinfo{author}{Krishnan, D.}, \bibinfo{year}{2020}.
\newblock \bibinfo{title}{Supervised contrastive learning}, in:
  \bibinfo{editor}{Larochelle, H.}, \bibinfo{editor}{Ranzato, M.},
  \bibinfo{editor}{Hadsell, R.}, \bibinfo{editor}{Balcan, M.},
  \bibinfo{editor}{Lin, H.} (Eds.), \bibinfo{booktitle}{Advances in Neural
  Information Processing Systems}, \bibinfo{publisher}{Curran Associates,
  Inc.}. pp. \bibinfo{pages}{18661--18673}.
\newblock \URLprefix
  \url{https://proceedings.neurips.cc/paper/2020/file/d89a66c7c80a29b1bdbab0f2a1a94af8-Paper.pdf}.
\bibitem[{Kim et~al.(2022)Kim, Jeong, Kim, Chong, Kim and Cho}]{kim2022spatial}
\bibinfo{author}{Kim, D.}, \bibinfo{author}{Jeong, D.}, \bibinfo{author}{Kim,
  H.}, \bibinfo{author}{Chong, K.}, \bibinfo{author}{Kim, S.},
  \bibinfo{author}{Cho, H.}, \bibinfo{year}{2022}.
\newblock \bibinfo{title}{Spatial contrastive learning for anomaly detection
  and localization}.
\newblock \bibinfo{journal}{IEEE Access} \bibinfo{volume}{10},
  \bibinfo{pages}{17366--17376}.
\bibitem[{Kim et~al.(2021)Kim, Ho and Kang}]{complex}
\bibinfo{author}{Kim, M.}, \bibinfo{author}{Ho, M.T.}, \bibinfo{author}{Kang,
  H.G.}, \bibinfo{year}{2021}.
\newblock \bibinfo{title}{Self-supervised complex network for machine sound
  anomaly detection}, in: \bibinfo{booktitle}{2021 29th European Signal
  Processing Conference (EUSIPCO)}, pp. \bibinfo{pages}{586--590}.
\newblock \DOIprefix\doi{10.23919/EUSIPCO54536.2021.9615923}.
\bibitem[{Kim et~al.(2015)Kim, Choi and Lee}]{kim2015deep}
\bibinfo{author}{Kim, S.}, \bibinfo{author}{Choi, Y.}, \bibinfo{author}{Lee,
  M.}, \bibinfo{year}{2015}.
\newblock \bibinfo{title}{Deep learning with support vector data description}.
\newblock \bibinfo{journal}{Neurocomputing} \bibinfo{volume}{165},
  \bibinfo{pages}{111--117}.
\bibitem[{Kiran et~al.(2018)Kiran, Thomas and Parakkal}]{kiran2018overview}
\bibinfo{author}{Kiran, B.R.}, \bibinfo{author}{Thomas, D.M.},
  \bibinfo{author}{Parakkal, R.}, \bibinfo{year}{2018}.
\newblock \bibinfo{title}{An overview of deep learning based methods for
  unsupervised and semi-supervised anomaly detection in videos}.
\newblock \bibinfo{journal}{Journal of Imaging} \bibinfo{volume}{4},
  \bibinfo{pages}{36}.
\bibitem[{Krizhevsky et~al.()Krizhevsky, Nair and Hinton}]{cifar}
\bibinfo{author}{Krizhevsky, A.}, \bibinfo{author}{Nair, V.},
  \bibinfo{author}{Hinton, G.}, .
\newblock \bibinfo{title}{Cifar-10 (canadian institute for advanced research)}.
\newblock \URLprefix \url{http://www.cs.toronto.edu/~kriz/cifar.html}.
\bibitem[{Larsson et~al.(2016)Larsson, Maire and
  Shakhnarovich}]{larsson2016learning}
\bibinfo{author}{Larsson, G.}, \bibinfo{author}{Maire, M.},
  \bibinfo{author}{Shakhnarovich, G.}, \bibinfo{year}{2016}.
\newblock \bibinfo{title}{Learning representations for automatic colorization},
  in: \bibinfo{booktitle}{European Conference on Computer Vision (ECCV)}.
\bibitem[{Latif et~al.(2018)Latif, Usman, Rana and
  Qadir}]{latif2018phonocardiographic}
\bibinfo{author}{Latif, S.}, \bibinfo{author}{Usman, M.},
  \bibinfo{author}{Rana, R.}, \bibinfo{author}{Qadir, J.},
  \bibinfo{year}{2018}.
\newblock \bibinfo{title}{Phonocardiographic sensing using deep learning for
  abnormal heartbeat detection}.
\newblock \bibinfo{journal}{IEEE Sensors Journal} \bibinfo{volume}{18},
  \bibinfo{pages}{9393--9400}.
\bibitem[{Lee et~al.(2018)Lee, Lee, Lee and Shin}]{lee2018simple}
\bibinfo{author}{Lee, K.}, \bibinfo{author}{Lee, K.}, \bibinfo{author}{Lee,
  H.}, \bibinfo{author}{Shin, J.}, \bibinfo{year}{2018}.
\newblock \bibinfo{title}{A simple unified framework for detecting
  out-of-distribution samples and adversarial attacks}.
\newblock \bibinfo{journal}{Advances in neural information processing systems}
  \bibinfo{volume}{31}.
\bibitem[{Li et~al.(2021)Li, Sohn, Yoon and Pfister}]{li2021cutpaste}
\bibinfo{author}{Li, C.L.}, \bibinfo{author}{Sohn, K.}, \bibinfo{author}{Yoon,
  J.}, \bibinfo{author}{Pfister, T.}, \bibinfo{year}{2021}.
\newblock \bibinfo{title}{Cutpaste: Self-supervised learning for anomaly
  detection and localization}, in: \bibinfo{booktitle}{Proceedings of the
  IEEE/CVF Conference on Computer Vision and Pattern Recognition}, pp.
  \bibinfo{pages}{9664--9674}.
\bibitem[{Liu et~al.(2021a)Liu, Xu and Xu}]{liu2021deepfib}
\bibinfo{author}{Liu, M.}, \bibinfo{author}{Xu, Z.}, \bibinfo{author}{Xu, Q.},
  \bibinfo{year}{2021}a.
\newblock \bibinfo{title}{Deepfib: Self-imputation for time series anomaly
  detection}.
\newblock \bibinfo{journal}{arXiv preprint arXiv:2112.06247} .
\bibitem[{Liu et~al.(2018)Liu, Garrepalli, Dietterich, Fern and
  Hendrycks}]{opencategory}
\bibinfo{author}{Liu, S.}, \bibinfo{author}{Garrepalli, R.},
  \bibinfo{author}{Dietterich, T.}, \bibinfo{author}{Fern, A.},
  \bibinfo{author}{Hendrycks, D.}, \bibinfo{year}{2018}.
\newblock \bibinfo{title}{Open category detection with {PAC} guarantees}, in:
  \bibinfo{editor}{Dy, J.}, \bibinfo{editor}{Krause, A.} (Eds.),
  \bibinfo{booktitle}{Proceedings of the 35th International Conference on
  Machine Learning}, \bibinfo{publisher}{PMLR}. pp.
  \bibinfo{pages}{3169--3178}.
\newblock \URLprefix \url{https://proceedings.mlr.press/v80/liu18e.html}.
\bibitem[{Liu et~al.(2022)Liu, Jin, Pan, Zhou, Zheng, Xia and
  Philip}]{liu2022graph}
\bibinfo{author}{Liu, Y.}, \bibinfo{author}{Jin, M.}, \bibinfo{author}{Pan,
  S.}, \bibinfo{author}{Zhou, C.}, \bibinfo{author}{Zheng, Y.},
  \bibinfo{author}{Xia, F.}, \bibinfo{author}{Philip, S.Y.},
  \bibinfo{year}{2022}.
\newblock \bibinfo{title}{Graph self-supervised learning: A survey}.
\newblock \bibinfo{journal}{IEEE Transactions on Knowledge and Data
  Engineering} \bibinfo{volume}{35}, \bibinfo{pages}{5879--5900}.
\bibitem[{Liu et~al.(2021b)Liu, Li, Pan, Gong, Zhou and
  Karypis}]{liu2021anomaly}
\bibinfo{author}{Liu, Y.}, \bibinfo{author}{Li, Z.}, \bibinfo{author}{Pan, S.},
  \bibinfo{author}{Gong, C.}, \bibinfo{author}{Zhou, C.},
  \bibinfo{author}{Karypis, G.}, \bibinfo{year}{2021}b.
\newblock \bibinfo{title}{Anomaly detection on attributed networks via
  contrastive self-supervised learning}.
\newblock \bibinfo{journal}{IEEE transactions on neural networks and learning
  systems} \bibinfo{volume}{33}, \bibinfo{pages}{2378--2392}.
\bibitem[{Liu et~al.(2021c)Liu, Pan, Wang, Xiong, Wang, Chen and
  Lee}]{liu2021Dynanomaly}
\bibinfo{author}{Liu, Y.}, \bibinfo{author}{Pan, S.}, \bibinfo{author}{Wang,
  Y.G.}, \bibinfo{author}{Xiong, F.}, \bibinfo{author}{Wang, L.},
  \bibinfo{author}{Chen, Q.}, \bibinfo{author}{Lee, V.C.},
  \bibinfo{year}{2021}c.
\newblock \bibinfo{title}{Anomaly detection in dynamic graphs via transformer}.
\newblock \bibinfo{journal}{IEEE Transactions on Knowledge and Data
  Engineering} .
\bibitem[{Liznerski et~al.(2020)Liznerski, Ruff, Vandermeulen, Franks, Kloft
  and M{\"u}ller}]{liznerski2020explainable}
\bibinfo{author}{Liznerski, P.}, \bibinfo{author}{Ruff, L.},
  \bibinfo{author}{Vandermeulen, R.A.}, \bibinfo{author}{Franks, B.J.},
  \bibinfo{author}{Kloft, M.}, \bibinfo{author}{M{\"u}ller, K.R.},
  \bibinfo{year}{2020}.
\newblock \bibinfo{title}{Explainable deep one-class classification}.
\newblock \bibinfo{journal}{arXiv preprint arXiv:2007.01760} .
\bibitem[{Luo et~al.(2022)Luo, Wu, Yang, Xue, Peng, Zhou, Chen, Li and
  Sheng}]{luo2022deep}
\bibinfo{author}{Luo, X.}, \bibinfo{author}{Wu, J.}, \bibinfo{author}{Yang,
  J.}, \bibinfo{author}{Xue, S.}, \bibinfo{author}{Peng, H.},
  \bibinfo{author}{Zhou, C.}, \bibinfo{author}{Chen, H.}, \bibinfo{author}{Li,
  Z.}, \bibinfo{author}{Sheng, Q.Z.}, \bibinfo{year}{2022}.
\newblock \bibinfo{title}{Deep graph level anomaly detection with contrastive
  learning}.
\newblock \bibinfo{journal}{Scientific Reports} \bibinfo{volume}{12},
  \bibinfo{pages}{19867}.
\bibitem[{Mahalanobis(1936)}]{MHD}
\bibinfo{author}{Mahalanobis, P.}, \bibinfo{year}{1936}.
\newblock \bibinfo{title}{On the generalised distance in statistics}, in:
  \bibinfo{booktitle}{Proceedings of the National Institute of Sciences of
  India}, pp. \bibinfo{pages}{49--55}.
\bibitem[{Malaiya et~al.(2018)Malaiya, Kwon, Kim, Suh, Kim and
  Kim}]{malaiya2018empirical}
\bibinfo{author}{Malaiya, R.K.}, \bibinfo{author}{Kwon, D.},
  \bibinfo{author}{Kim, J.}, \bibinfo{author}{Suh, S.C.}, \bibinfo{author}{Kim,
  H.}, \bibinfo{author}{Kim, I.}, \bibinfo{year}{2018}.
\newblock \bibinfo{title}{An empirical evaluation of deep learning for network
  anomaly detection}, in: \bibinfo{booktitle}{2018 International Conference on
  Computing, Networking and Communications (ICNC)},
  \bibinfo{organization}{IEEE}. pp. \bibinfo{pages}{893--898}.
\bibitem[{Manolache et~al.(2021a)Manolache, Brad and
  Burceanu}]{manolache-etal-2021-date}
\bibinfo{author}{Manolache, A.}, \bibinfo{author}{Brad, F.},
  \bibinfo{author}{Burceanu, E.}, \bibinfo{year}{2021}a.
\newblock \bibinfo{title}{{DATE}: Detecting anomalies in text via
  self-supervision of transformers}, in: \bibinfo{booktitle}{Proceedings of the
  2021 Conference of the North American Chapter of the Association for
  Computational Linguistics: Human Language Technologies},
  \bibinfo{publisher}{Association for Computational Linguistics},
  \bibinfo{address}{Online}. pp. \bibinfo{pages}{267--277}.
\newblock \URLprefix \url{https://aclanthology.org/2021.naacl-main.25},
  \DOIprefix\doi{10.18653/v1/2021.naacl-main.25}.
\bibitem[{Manolache et~al.(2021b)Manolache, Brad and
  Burceanu}]{manolache2021date}
\bibinfo{author}{Manolache, A.}, \bibinfo{author}{Brad, F.},
  \bibinfo{author}{Burceanu, E.}, \bibinfo{year}{2021}b.
\newblock \bibinfo{title}{Date: Detecting anomalies in text via
  self-supervision of transformers}.
\newblock \bibinfo{journal}{arXiv preprint arXiv:2104.05591} .
\bibitem[{Min et~al.(2018)Min, Long, Liu, Cui, Cai and Ma}]{min2018ids}
\bibinfo{author}{Min, E.}, \bibinfo{author}{Long, J.}, \bibinfo{author}{Liu,
  Q.}, \bibinfo{author}{Cui, J.}, \bibinfo{author}{Cai, Z.},
  \bibinfo{author}{Ma, J.}, \bibinfo{year}{2018}.
\newblock \bibinfo{title}{Su-ids: A semi-supervised and unsupervised framework
  for network intrusion detection}, in: \bibinfo{booktitle}{International
  Conference on Cloud Computing and Security},
  \bibinfo{organization}{Springer}. pp. \bibinfo{pages}{322--334}.
\bibitem[{Mohseni et~al.(2020)Mohseni, Pitale, Yadawa and
  Wang}]{Mohseni_Pitale_Yadawa_Wang_2020}
\bibinfo{author}{Mohseni, S.}, \bibinfo{author}{Pitale, M.},
  \bibinfo{author}{Yadawa, J.}, \bibinfo{author}{Wang, Z.},
  \bibinfo{year}{2020}.
\newblock \bibinfo{title}{Self-supervised learning for generalizable
  out-of-distribution detection}.
\newblock \bibinfo{journal}{Proceedings of the AAAI Conference on Artificial
  Intelligence} \bibinfo{volume}{34}, \bibinfo{pages}{5216--5223}.
\newblock \URLprefix
  \url{https://ojs.aaai.org/index.php/AAAI/article/view/5966},
  \DOIprefix\doi{10.1609/aaai.v34i04.5966}.
\bibitem[{Pang et~al.(2021)Pang, Shen, Cao and Hengel}]{SurveyPang}
\bibinfo{author}{Pang, G.}, \bibinfo{author}{Shen, C.}, \bibinfo{author}{Cao,
  L.}, \bibinfo{author}{Hengel, A.V.D.}, \bibinfo{year}{2021}.
\newblock \bibinfo{title}{Deep learning for anomaly detection: A review}.
\newblock \bibinfo{journal}{ACM Comput. Surv.} \bibinfo{volume}{54}.
\newblock \URLprefix \url{https://doi.org/10.1145/3439950},
  \DOIprefix\doi{10.1145/3439950}.
\bibitem[{Park et~al.(2021)Park, Balint and Hwang}]{park2021self}
\bibinfo{author}{Park, S.}, \bibinfo{author}{Balint, A.},
  \bibinfo{author}{Hwang, H.}, \bibinfo{year}{2021}.
\newblock \bibinfo{title}{Self-supervised medical out-of-distribution using
  u-net vision transformers}, in: \bibinfo{booktitle}{International Conference
  on Medical Image Computing and Computer-Assisted Intervention},
  \bibinfo{organization}{Springer}. pp. \bibinfo{pages}{104--110}.
\bibitem[{Pirnay and Chai(2021)}]{pirnay2021inpainting}
\bibinfo{author}{Pirnay, J.}, \bibinfo{author}{Chai, K.}, \bibinfo{year}{2021}.
\newblock \bibinfo{title}{Inpainting transformer for anomaly detection}.
\newblock \bibinfo{journal}{arXiv preprint arXiv:2104.13897} .
\bibitem[{Qiu et~al.(2021)Qiu, Pfrommer, Kloft, Mandt and
  Rudolph}]{pmlr-v139-qiu21a}
\bibinfo{author}{Qiu, C.}, \bibinfo{author}{Pfrommer, T.},
  \bibinfo{author}{Kloft, M.}, \bibinfo{author}{Mandt, S.},
  \bibinfo{author}{Rudolph, M.}, \bibinfo{year}{2021}.
\newblock \bibinfo{title}{Neural transformation learning for deep anomaly
  detection beyond images}, in: \bibinfo{editor}{Meila, M.},
  \bibinfo{editor}{Zhang, T.} (Eds.), \bibinfo{booktitle}{Proceedings of the
  38th International Conference on Machine Learning},
  \bibinfo{publisher}{PMLR}. pp. \bibinfo{pages}{8703--8714}.
\newblock \URLprefix \url{https://proceedings.mlr.press/v139/qiu21a.html}.
\bibitem[{Rafiee et~al.(2022)Rafiee, Gholamipoorfard, Adaloglou, Jaxy, Ramakers
  and Kollmann}]{rafiee2022self}
\bibinfo{author}{Rafiee, N.}, \bibinfo{author}{Gholamipoorfard, R.},
  \bibinfo{author}{Adaloglou, N.}, \bibinfo{author}{Jaxy, S.},
  \bibinfo{author}{Ramakers, J.}, \bibinfo{author}{Kollmann, M.},
  \bibinfo{year}{2022}.
\newblock \bibinfo{title}{Self-supervised anomaly detection by
  self-distillation and negative sampling}.
\newblock \bibinfo{journal}{arXiv preprint arXiv:2201.06378} .
\bibitem[{Ravanelli et~al.(2020)Ravanelli, Zhong, Pascual, Swietojanski,
  Monteiro, Trmal and Bengio}]{speechSSL}
\bibinfo{author}{Ravanelli, M.}, \bibinfo{author}{Zhong, J.},
  \bibinfo{author}{Pascual, S.}, \bibinfo{author}{Swietojanski, P.},
  \bibinfo{author}{Monteiro, J.}, \bibinfo{author}{Trmal, J.},
  \bibinfo{author}{Bengio, Y.}, \bibinfo{year}{2020}.
\newblock \bibinfo{title}{Multi-task self-supervised learning for robust speech
  recognition}, in: \bibinfo{booktitle}{ICASSP 2020 - 2020 IEEE International
  Conference on Acoustics, Speech and Signal Processing (ICASSP)}, pp.
  \bibinfo{pages}{6989--6993}.
\newblock \DOIprefix\doi{10.1109/ICASSP40776.2020.9053569}.
\bibitem[{Reiss et~al.(2021)Reiss, Cohen, Bergman and Hoshen}]{reiss2021panda}
\bibinfo{author}{Reiss, T.}, \bibinfo{author}{Cohen, N.},
  \bibinfo{author}{Bergman, L.}, \bibinfo{author}{Hoshen, Y.},
  \bibinfo{year}{2021}.
\newblock \bibinfo{title}{Panda: Adapting pretrained features for anomaly
  detection and segmentation}, in: \bibinfo{booktitle}{Proceedings of the
  IEEE/CVF Conference on Computer Vision and Pattern Recognition}, pp.
  \bibinfo{pages}{2806--2814}.
\bibitem[{Reiss and Hoshen(2021)}]{reiss2021mean}
\bibinfo{author}{Reiss, T.}, \bibinfo{author}{Hoshen, Y.},
  \bibinfo{year}{2021}.
\newblock \bibinfo{title}{Mean-shifted contrastive loss for anomaly detection}.
\newblock \bibinfo{journal}{arXiv preprint arXiv:2106.03844} .
\bibitem[{Rippel et~al.(2021)Rippel, Mertens and Merhof}]{rippel2021modeling}
\bibinfo{author}{Rippel, O.}, \bibinfo{author}{Mertens, P.},
  \bibinfo{author}{Merhof, D.}, \bibinfo{year}{2021}.
\newblock \bibinfo{title}{Modeling the distribution of normal data in
  pre-trained deep features for anomaly detection}, in:
  \bibinfo{booktitle}{2020 25th International Conference on Pattern Recognition
  (ICPR)}, \bibinfo{organization}{IEEE}. pp. \bibinfo{pages}{6726--6733}.
\bibitem[{Rudolph et~al.(2021)Rudolph, Wandt and Rosenhahn}]{rudolph2021same}
\bibinfo{author}{Rudolph, M.}, \bibinfo{author}{Wandt, B.},
  \bibinfo{author}{Rosenhahn, B.}, \bibinfo{year}{2021}.
\newblock \bibinfo{title}{Same same but differnet: Semi-supervised defect
  detection with normalizing flows}, in: \bibinfo{booktitle}{Proceedings of the
  IEEE/CVF Winter Conference on Applications of Computer Vision}, pp.
  \bibinfo{pages}{1907--1916}.
\bibitem[{Rudolph et~al.(2022)Rudolph, Wehrbein, Rosenhahn and
  Wandt}]{rudolph2022fully}
\bibinfo{author}{Rudolph, M.}, \bibinfo{author}{Wehrbein, T.},
  \bibinfo{author}{Rosenhahn, B.}, \bibinfo{author}{Wandt, B.},
  \bibinfo{year}{2022}.
\newblock \bibinfo{title}{Fully convolutional cross-scale-flows for image-based
  defect detection}, in: \bibinfo{booktitle}{Proceedings of the IEEE/CVF Winter
  Conference on Applications of Computer Vision}, pp.
  \bibinfo{pages}{1088--1097}.
\bibitem[{Ruff et~al.(2021)Ruff, Kauffmann, Vandermeulen, Montavon, Samek,
  Kloft, Dietterich and M{\"u}ller}]{ruff2021unifying}
\bibinfo{author}{Ruff, L.}, \bibinfo{author}{Kauffmann, J.R.},
  \bibinfo{author}{Vandermeulen, R.A.}, \bibinfo{author}{Montavon, G.},
  \bibinfo{author}{Samek, W.}, \bibinfo{author}{Kloft, M.},
  \bibinfo{author}{Dietterich, T.G.}, \bibinfo{author}{M{\"u}ller, K.R.},
  \bibinfo{year}{2021}.
\newblock \bibinfo{title}{A unifying review of deep and shallow anomaly
  detection}.
\newblock \bibinfo{journal}{Proceedings of the IEEE} .
\bibitem[{Ruff et~al.(2018)Ruff, Vandermeulen, Goernitz, Deecke, Siddiqui,
  Binder, M{\"u}ller and Kloft}]{ruff2018deep}
\bibinfo{author}{Ruff, L.}, \bibinfo{author}{Vandermeulen, R.},
  \bibinfo{author}{Goernitz, N.}, \bibinfo{author}{Deecke, L.},
  \bibinfo{author}{Siddiqui, S.A.}, \bibinfo{author}{Binder, A.},
  \bibinfo{author}{M{\"u}ller, E.}, \bibinfo{author}{Kloft, M.},
  \bibinfo{year}{2018}.
\newblock \bibinfo{title}{Deep one-class classification}, in:
  \bibinfo{booktitle}{International conference on machine learning},
  \bibinfo{organization}{PMLR}. pp. \bibinfo{pages}{4393--4402}.
\bibitem[{Ruff et~al.(2019)Ruff, Vandermeulen, G{\"o}rnitz, Binder, M{\"u}ller,
  M{\"u}ller and Kloft}]{ruff2019deep}
\bibinfo{author}{Ruff, L.}, \bibinfo{author}{Vandermeulen, R.A.},
  \bibinfo{author}{G{\"o}rnitz, N.}, \bibinfo{author}{Binder, A.},
  \bibinfo{author}{M{\"u}ller, E.}, \bibinfo{author}{M{\"u}ller, K.R.},
  \bibinfo{author}{Kloft, M.}, \bibinfo{year}{2019}.
\newblock \bibinfo{title}{Deep semi-supervised anomaly detection}.
\newblock \bibinfo{journal}{arXiv preprint arXiv:1906.02694} .
\bibitem[{Sabokrou et~al.(2019)Sabokrou, Khalooei and
  Adeli}]{Sabokrou_2019_ICCV}
\bibinfo{author}{Sabokrou, M.}, \bibinfo{author}{Khalooei, M.},
  \bibinfo{author}{Adeli, E.}, \bibinfo{year}{2019}.
\newblock \bibinfo{title}{Self-supervised representation learning via
  neighborhood-relational encoding}, in: \bibinfo{booktitle}{Proceedings of the
  IEEE/CVF International Conference on Computer Vision (ICCV)}.
\bibitem[{Salehi et~al.(2020)Salehi, Eftekhar, Sadjadi, Rohban and
  Rabiee}]{salehi2020puzzleae}
\bibinfo{author}{Salehi, M.}, \bibinfo{author}{Eftekhar, A.},
  \bibinfo{author}{Sadjadi, N.}, \bibinfo{author}{Rohban, M.H.},
  \bibinfo{author}{Rabiee, H.R.}, \bibinfo{year}{2020}.
\newblock \bibinfo{title}{Puzzle-ae: Novelty detection in images through
  solving puzzles}.
\newblock \href{http://arxiv.org/abs/2008.12959}{{\tt arXiv:2008.12959}}.
\bibitem[{Schlegl et~al.(2017)Schlegl, Seeb{\"o}ck, Waldstein, Schmidt-Erfurth
  and Langs}]{schlegl2017unsupervised}
\bibinfo{author}{Schlegl, T.}, \bibinfo{author}{Seeb{\"o}ck, P.},
  \bibinfo{author}{Waldstein, S.M.}, \bibinfo{author}{Schmidt-Erfurth, U.},
  \bibinfo{author}{Langs, G.}, \bibinfo{year}{2017}.
\newblock \bibinfo{title}{Unsupervised anomaly detection with generative
  adversarial networks to guide marker discovery}, in:
  \bibinfo{booktitle}{International conference on information processing in
  medical imaging}, \bibinfo{organization}{Springer}. pp.
  \bibinfo{pages}{146--157}.
\bibitem[{Schl{\"u}ter et~al.(2021)Schl{\"u}ter, Tan, Hou and
  Kainz}]{schluter2021self}
\bibinfo{author}{Schl{\"u}ter, H.M.}, \bibinfo{author}{Tan, J.},
  \bibinfo{author}{Hou, B.}, \bibinfo{author}{Kainz, B.}, \bibinfo{year}{2021}.
\newblock \bibinfo{title}{Self-supervised out-of-distribution detection and
  localization with natural synthetic anomalies (nsa)}.
\newblock \bibinfo{journal}{arXiv preprint arXiv:2109.15222} .
\bibitem[{Schreyer et~al.(2021)Schreyer, Sattarov and
  Borth}]{schreyer2021multi}
\bibinfo{author}{Schreyer, M.}, \bibinfo{author}{Sattarov, T.},
  \bibinfo{author}{Borth, D.}, \bibinfo{year}{2021}.
\newblock \bibinfo{title}{Multi-view contrastive self-supervised learning of
  accounting data representations for downstream audit tasks}.
\newblock \bibinfo{journal}{arXiv preprint arXiv:2109.11201} .
\bibitem[{Schroff et~al.(2015)Schroff, Kalenichenko and Philbin}]{triplet}
\bibinfo{author}{Schroff, F.}, \bibinfo{author}{Kalenichenko, D.},
  \bibinfo{author}{Philbin, J.}, \bibinfo{year}{2015}.
\newblock \bibinfo{title}{Facenet: A unified embedding for face recognition and
  clustering}, in: \bibinfo{booktitle}{2015 IEEE Conference on Computer Vision
  and Pattern Recognition (CVPR)}, pp. \bibinfo{pages}{815--823}.
\newblock \DOIprefix\doi{10.1109/CVPR.2015.7298682}.
\bibitem[{Sehwag et~al.(2021)Sehwag, Chiang and Mittal}]{sehwag2021ssd}
\bibinfo{author}{Sehwag, V.}, \bibinfo{author}{Chiang, M.},
  \bibinfo{author}{Mittal, P.}, \bibinfo{year}{2021}.
\newblock \bibinfo{title}{{\{}SSD{\}}: A unified framework for self-supervised
  outlier detection}, in: \bibinfo{booktitle}{International Conference on
  Learning Representations}.
\newblock \URLprefix \url{https://openreview.net/forum?id=v5gjXpmR8J}.
\bibitem[{Shenkar and Wolf(2022)}]{shenkar2022anomaly}
\bibinfo{author}{Shenkar, T.}, \bibinfo{author}{Wolf, L.},
  \bibinfo{year}{2022}.
\newblock \bibinfo{title}{Anomaly detection for tabular data with internal
  contrastive learning}, in: \bibinfo{booktitle}{International Conference on
  Learning Representations}.
\bibitem[{Shi et~al.(2021)Shi, Yang and Qi}]{shi2021unsupervised}
\bibinfo{author}{Shi, Y.}, \bibinfo{author}{Yang, J.}, \bibinfo{author}{Qi,
  Z.}, \bibinfo{year}{2021}.
\newblock \bibinfo{title}{Unsupervised anomaly segmentation via deep feature
  reconstruction}.
\newblock \bibinfo{journal}{Neurocomputing} \bibinfo{volume}{424},
  \bibinfo{pages}{9--22}.
\bibitem[{Sohn et~al.(2020)Sohn, Li, Yoon, Jin and Pfister}]{sohn2020learning}
\bibinfo{author}{Sohn, K.}, \bibinfo{author}{Li, C.L.}, \bibinfo{author}{Yoon,
  J.}, \bibinfo{author}{Jin, M.}, \bibinfo{author}{Pfister, T.},
  \bibinfo{year}{2020}.
\newblock \bibinfo{title}{Learning and evaluating representations for deep
  one-class classification}.
\newblock \bibinfo{journal}{arXiv preprint arXiv:2011.02578} .
\bibitem[{Song et~al.(2021)Song, Kong, Park, Kim and Kang}]{song2021anoseg}
\bibinfo{author}{Song, J.}, \bibinfo{author}{Kong, K.}, \bibinfo{author}{Park,
  Y.I.}, \bibinfo{author}{Kim, S.G.}, \bibinfo{author}{Kang, S.J.},
  \bibinfo{year}{2021}.
\newblock \bibinfo{title}{Anoseg: Anomaly segmentation network using
  self-supervised learning}.
\newblock \bibinfo{journal}{arXiv preprint arXiv:2110.03396} .
\bibitem[{Spahr et~al.(2021)Spahr, Bozorgtabar and Thiran}]{spahr2021self}
\bibinfo{author}{Spahr, A.}, \bibinfo{author}{Bozorgtabar, B.},
  \bibinfo{author}{Thiran, J.P.}, \bibinfo{year}{2021}.
\newblock \bibinfo{title}{Self-taught semi-supervised anomaly detection on
  upper limb x-rays}, in: \bibinfo{booktitle}{2021 IEEE 18th International
  Symposium on Biomedical Imaging (ISBI)}, \bibinfo{organization}{IEEE}. pp.
  \bibinfo{pages}{1632--1636}.
\bibitem[{Tack et~al.(2020)Tack, Mo, Jeong and Shin}]{tack2020csi}
\bibinfo{author}{Tack, J.}, \bibinfo{author}{Mo, S.}, \bibinfo{author}{Jeong,
  J.}, \bibinfo{author}{Shin, J.}, \bibinfo{year}{2020}.
\newblock \bibinfo{title}{Csi: Novelty detection via contrastive learning on
  distributionally shifted instances}.
\newblock \bibinfo{journal}{Advances in neural information processing systems}
  \bibinfo{volume}{33}, \bibinfo{pages}{11839--11852}.
\bibitem[{Tax and Duin(2004)}]{tax2004support}
\bibinfo{author}{Tax, D.M.}, \bibinfo{author}{Duin, R.P.},
  \bibinfo{year}{2004}.
\newblock \bibinfo{title}{Support vector data description}.
\newblock \bibinfo{journal}{Machine learning} \bibinfo{volume}{54},
  \bibinfo{pages}{45--66}.
\bibitem[{Tsai et~al.(2022)Tsai, Wu and Lai}]{tsai2022multi}
\bibinfo{author}{Tsai, C.C.}, \bibinfo{author}{Wu, T.H.}, \bibinfo{author}{Lai,
  S.H.}, \bibinfo{year}{2022}.
\newblock \bibinfo{title}{Multi-scale patch-based representation learning for
  image anomaly detection and segmentation}, in:
  \bibinfo{booktitle}{Proceedings of the IEEE/CVF Winter Conference on
  Applications of Computer Vision}, pp. \bibinfo{pages}{3992--4000}.
\bibitem[{Valerio~Massoli et~al.(2020)Valerio~Massoli, Falchi, Kantarci, Akti,
  Kemal~Ekenel and Amato}]{valerio2020mocca}
\bibinfo{author}{Valerio~Massoli, F.}, \bibinfo{author}{Falchi, F.},
  \bibinfo{author}{Kantarci, A.}, \bibinfo{author}{Akti, {\c{S}}.},
  \bibinfo{author}{Kemal~Ekenel, H.}, \bibinfo{author}{Amato, G.},
  \bibinfo{year}{2020}.
\newblock \bibinfo{title}{Mocca: Multi-layer one-class classification for
  anomaly detection}.
\newblock \bibinfo{journal}{arXiv e-prints} , \bibinfo{pages}{arXiv--2012}.
\bibitem[{Venkatakrishnan et~al.(2020)Venkatakrishnan, Kim, Eisawy, Pfister and
  Navab}]{venkatakrishnan2020self}
\bibinfo{author}{Venkatakrishnan, A.R.}, \bibinfo{author}{Kim, S.T.},
  \bibinfo{author}{Eisawy, R.}, \bibinfo{author}{Pfister, F.},
  \bibinfo{author}{Navab, N.}, \bibinfo{year}{2020}.
\newblock \bibinfo{title}{Self-supervised out-of-distribution detection in
  brain ct scans}.
\newblock \bibinfo{journal}{arXiv preprint arXiv:2011.05428} .
\bibitem[{Venkataramanan et~al.(2020)Venkataramanan, Peng, Singh and
  Mahalanobis}]{venkataramanan2020attention}
\bibinfo{author}{Venkataramanan, S.}, \bibinfo{author}{Peng, K.C.},
  \bibinfo{author}{Singh, R.V.}, \bibinfo{author}{Mahalanobis, A.},
  \bibinfo{year}{2020}.
\newblock \bibinfo{title}{Attention guided anomaly localization in images}, in:
  \bibinfo{booktitle}{European Conference on Computer Vision},
  \bibinfo{organization}{Springer}. pp. \bibinfo{pages}{485--503}.
\bibitem[{Villa-Perez et~al.(2021)Villa-Perez, Alvarez-Carmona,
  Loyola-Gonzalez, Medina-Perez, Velazco-Rossell and
  Choo}]{SemisupervisedSurvey}
\bibinfo{author}{Villa-Perez, M.E.}, \bibinfo{author}{Alvarez-Carmona, M.A.},
  \bibinfo{author}{Loyola-Gonzalez, O.}, \bibinfo{author}{Medina-Perez, M.A.},
  \bibinfo{author}{Velazco-Rossell, J.C.}, \bibinfo{author}{Choo, K.K.R.},
  \bibinfo{year}{2021}.
\newblock \bibinfo{title}{Semi-supervised anomaly detection algorithms: A
  comparative summary and future research directions}.
\newblock \bibinfo{journal}{Knowledge-Based Systems} \bibinfo{volume}{218},
  \bibinfo{pages}{106878}.
\newblock \URLprefix
  \url{https://www.sciencedirect.com/science/article/pii/S0950705121001416},
  \DOIprefix\doi{https://doi.org/10.1016/j.knosys.2021.106878}.
\bibitem[{Wang et~al.(2021a)Wang, Dou, Chen, Chen, Liu and
  Philip}]{wang2021deep}
\bibinfo{author}{Wang, C.}, \bibinfo{author}{Dou, Y.}, \bibinfo{author}{Chen,
  M.}, \bibinfo{author}{Chen, J.}, \bibinfo{author}{Liu, Z.},
  \bibinfo{author}{Philip, S.Y.}, \bibinfo{year}{2021}a.
\newblock \bibinfo{title}{Deep fraud detection on non-attributed graph}, in:
  \bibinfo{booktitle}{2021 IEEE International Conference on Big Data (Big
  Data)}, \bibinfo{organization}{IEEE}. pp. \bibinfo{pages}{5470--5473}.
\bibitem[{Wang et~al.(2021b)Wang, Han, Ding and Huang}]{wang2021student}
\bibinfo{author}{Wang, G.}, \bibinfo{author}{Han, S.}, \bibinfo{author}{Ding,
  E.}, \bibinfo{author}{Huang, D.}, \bibinfo{year}{2021}b.
\newblock \bibinfo{title}{Student-teacher feature pyramid matching for
  unsupervised anomaly detection}.
\newblock \bibinfo{journal}{arXiv preprint arXiv:2103.04257} .
\bibitem[{Wang et~al.(2019)Wang, Bah and Hammad}]{wang2019progress}
\bibinfo{author}{Wang, H.}, \bibinfo{author}{Bah, M.J.},
  \bibinfo{author}{Hammad, M.}, \bibinfo{year}{2019}.
\newblock \bibinfo{title}{Progress in outlier detection techniques: A survey}.
\newblock \bibinfo{journal}{Ieee Access} \bibinfo{volume}{7},
  \bibinfo{pages}{107964--108000}.
\bibitem[{Wang et~al.(2023)Wang, Liu, Mou, Gao, Guo, Liu, Wo and
  Liu}]{wang2023deep}
\bibinfo{author}{Wang, R.}, \bibinfo{author}{Liu, C.}, \bibinfo{author}{Mou,
  X.}, \bibinfo{author}{Gao, K.}, \bibinfo{author}{Guo, X.},
  \bibinfo{author}{Liu, P.}, \bibinfo{author}{Wo, T.}, \bibinfo{author}{Liu,
  X.}, \bibinfo{year}{2023}.
\newblock \bibinfo{title}{Deep contrastive one-class time series anomaly
  detection}.
\newblock \href{http://arxiv.org/abs/2207.01472}{{\tt arXiv:2207.01472}}.
\bibitem[{Wang et~al.(2021c)Wang, Qin, Wei, Xu, Bai and Fu}]{slap}
\bibinfo{author}{Wang, Y.}, \bibinfo{author}{Qin, C.}, \bibinfo{author}{Wei,
  R.}, \bibinfo{author}{Xu, Y.}, \bibinfo{author}{Bai, Y.},
  \bibinfo{author}{Fu, Y.}, \bibinfo{year}{2021}c.
\newblock \bibinfo{title}{Sla$^2$p: Self-supervised anomaly detection with
  adversarial perturbation}.
\newblock \URLprefix \url{https://arxiv.org/abs/2111.12896},
  \DOIprefix\doi{10.48550/ARXIV.2111.12896}.
\bibitem[{Weng and Kim(2021)}]{nipsssl}
\bibinfo{author}{Weng, L.}, \bibinfo{author}{Kim, J.W.}, \bibinfo{year}{2021}.
\newblock \bibinfo{title}{Tutorial: Self-supervised learning}, in:
  \bibinfo{editor}{Canziani, A.}, \bibinfo{editor}{Grant, E.} (Eds.),
  \bibinfo{booktitle}{Advances in Neural Information Processing Systems}.
\newblock \URLprefix \url{https://nips.cc/virtual/2021/tutorial/21895}.
\bibitem[{Winkens et~al.(2020)Winkens, Bunel, Roy, Stanforth, Natarajan,
  Ledsam, MacWilliams, Kohli, Karthikesalingam, Kohl, Cemgil, Eslami and
  Ronneberger}]{Winkens}
\bibinfo{author}{Winkens, J.}, \bibinfo{author}{Bunel, R.},
  \bibinfo{author}{Roy, A.G.}, \bibinfo{author}{Stanforth, R.},
  \bibinfo{author}{Natarajan, V.}, \bibinfo{author}{Ledsam, J.R.},
  \bibinfo{author}{MacWilliams, P.}, \bibinfo{author}{Kohli, P.},
  \bibinfo{author}{Karthikesalingam, A.}, \bibinfo{author}{Kohl, S.},
  \bibinfo{author}{Cemgil, T.}, \bibinfo{author}{Eslami, S.M.A.},
  \bibinfo{author}{Ronneberger, O.}, \bibinfo{year}{2020}.
\newblock \bibinfo{title}{Contrastive training for improved out-of-distribution
  detection}.
\newblock \URLprefix \url{https://arxiv.org/abs/2007.05566},
  \DOIprefix\doi{10.48550/ARXIV.2007.05566}.
\bibitem[{Wu et~al.(2018)Wu, Xiong, Yu and Lin}]{Wu_2018_CVPR}
\bibinfo{author}{Wu, Z.}, \bibinfo{author}{Xiong, Y.}, \bibinfo{author}{Yu,
  S.X.}, \bibinfo{author}{Lin, D.}, \bibinfo{year}{2018}.
\newblock \bibinfo{title}{Unsupervised feature learning via non-parametric
  instance discrimination}, in: \bibinfo{booktitle}{Proceedings of the IEEE
  Conference on Computer Vision and Pattern Recognition (CVPR)}.
\bibitem[{Xia et~al.(2022)Xia, Pan, Li, He, Ma, Zhang and Ding}]{GanSurveyXia}
\bibinfo{author}{Xia, X.}, \bibinfo{author}{Pan, X.}, \bibinfo{author}{Li, N.},
  \bibinfo{author}{He, X.}, \bibinfo{author}{Ma, L.}, \bibinfo{author}{Zhang,
  X.}, \bibinfo{author}{Ding, N.}, \bibinfo{year}{2022}.
\newblock \bibinfo{title}{Gan-based anomaly detection: A review}.
\newblock \bibinfo{journal}{Neurocomputing} \URLprefix
  \url{https://www.sciencedirect.com/science/article/pii/S0925231221019482},
  \DOIprefix\doi{https://doi.org/10.1016/j.neucom.2021.12.093}.
\bibitem[{Xin et~al.(2018)Xin, Kong, Liu, Chen, Li, Zhu, Gao, Hou and
  Wang}]{xin2018machine}
\bibinfo{author}{Xin, Y.}, \bibinfo{author}{Kong, L.}, \bibinfo{author}{Liu,
  Z.}, \bibinfo{author}{Chen, Y.}, \bibinfo{author}{Li, Y.},
  \bibinfo{author}{Zhu, H.}, \bibinfo{author}{Gao, M.}, \bibinfo{author}{Hou,
  H.}, \bibinfo{author}{Wang, C.}, \bibinfo{year}{2018}.
\newblock \bibinfo{title}{Machine learning and deep learning methods for
  cybersecurity}.
\newblock \bibinfo{journal}{Ieee access} \bibinfo{volume}{6},
  \bibinfo{pages}{35365--35381}.
\bibitem[{Xu et~al.(2020)Xu, Zheng, Mao, Wang and Zheng}]{xu2020anomaly}
\bibinfo{author}{Xu, J.}, \bibinfo{author}{Zheng, Y.}, \bibinfo{author}{Mao,
  Y.}, \bibinfo{author}{Wang, R.}, \bibinfo{author}{Zheng, W.S.},
  \bibinfo{year}{2020}.
\newblock \bibinfo{title}{Anomaly detection on electroencephalography with
  self-supervised learning}, in: \bibinfo{booktitle}{2020 IEEE International
  Conference on Bioinformatics and Biomedicine (BIBM)},
  \bibinfo{organization}{IEEE}. pp. \bibinfo{pages}{363--368}.
\bibitem[{Xu et~al.(2022)Xu, Huang, Zhao, Dong and Li}]{xu2022contrastive}
\bibinfo{author}{Xu, Z.}, \bibinfo{author}{Huang, X.}, \bibinfo{author}{Zhao,
  Y.}, \bibinfo{author}{Dong, Y.}, \bibinfo{author}{Li, J.},
  \bibinfo{year}{2022}.
\newblock \bibinfo{title}{Contrastive attributed network anomaly detection with
  data augmentation}, in: \bibinfo{booktitle}{Advances in Knowledge Discovery
  and Data Mining: 26th Pacific-Asia Conference, PAKDD 2022, Chengdu, China,
  May 16--19, 2022, Proceedings, Part II}, \bibinfo{organization}{Springer}.
  pp. \bibinfo{pages}{444--457}.
\bibitem[{Yi and Yoon(2020)}]{yi2020patch}
\bibinfo{author}{Yi, J.}, \bibinfo{author}{Yoon, S.}, \bibinfo{year}{2020}.
\newblock \bibinfo{title}{Patch svdd: Patch-level svdd for anomaly detection
  and segmentation}, in: \bibinfo{booktitle}{Proceedings of the Asian
  Conference on Computer Vision}.
\bibitem[{Zavrtanik et~al.(2021)Zavrtanik, Kristan and
  Sko{\v{c}}aj}]{zavrtanik2021draem}
\bibinfo{author}{Zavrtanik, V.}, \bibinfo{author}{Kristan, M.},
  \bibinfo{author}{Sko{\v{c}}aj, D.}, \bibinfo{year}{2021}.
\newblock \bibinfo{title}{Draem-a discriminatively trained reconstruction
  embedding for surface anomaly detection}, in: \bibinfo{booktitle}{Proceedings
  of the IEEE/CVF International Conference on Computer Vision}, pp.
  \bibinfo{pages}{8330--8339}.
\bibitem[{Zbontar et~al.(2021)Zbontar, Jing, Misra, LeCun and Deny}]{barlow}
\bibinfo{author}{Zbontar, J.}, \bibinfo{author}{Jing, L.},
  \bibinfo{author}{Misra, I.}, \bibinfo{author}{LeCun, Y.},
  \bibinfo{author}{Deny, S.}, \bibinfo{year}{2021}.
\newblock \bibinfo{title}{Barlow twins: Self-supervised learning via redundancy
  reduction}, in: \bibinfo{editor}{Meila, M.}, \bibinfo{editor}{Zhang, T.}
  (Eds.), \bibinfo{booktitle}{Proceedings of the 38th International Conference
  on Machine Learning}, \bibinfo{publisher}{PMLR}. pp.
  \bibinfo{pages}{12310--12320}.
\newblock \URLprefix \url{https://proceedings.mlr.press/v139/zbontar21a.html}.
\bibitem[{Zeng et~al.(2023)Zeng, Song, Zhuo, Zhou, Li, Xue, Dai and
  McLoughlin}]{icasp23}
\bibinfo{author}{Zeng, X.M.}, \bibinfo{author}{Song, Y.},
  \bibinfo{author}{Zhuo, Z.}, \bibinfo{author}{Zhou, Y.}, \bibinfo{author}{Li,
  Y.H.}, \bibinfo{author}{Xue, H.}, \bibinfo{author}{Dai, L.R.},
  \bibinfo{author}{McLoughlin, I.}, \bibinfo{year}{2023}.
\newblock \bibinfo{title}{Joint generative-contrastive representation learning
  for anomalous sound detection}, in: \bibinfo{booktitle}{ICASSP 2023 - 2023
  IEEE International Conference on Acoustics, Speech and Signal Processing
  (ICASSP)}, pp. \bibinfo{pages}{1--5}.
\newblock \DOIprefix\doi{10.1109/ICASSP49357.2023.10095568}.
\bibitem[{Zhang et~al.(2021a)Zhang, Saleeby, Feldhausen, Bi, Plotkowski and
  Womble}]{zhang2021selfsupervised}
\bibinfo{author}{Zhang, J.}, \bibinfo{author}{Saleeby, K.},
  \bibinfo{author}{Feldhausen, T.}, \bibinfo{author}{Bi, S.},
  \bibinfo{author}{Plotkowski, A.}, \bibinfo{author}{Womble, D.},
  \bibinfo{year}{2021}a.
\newblock \bibinfo{title}{Self-supervised anomaly detection via neural
  autoregressive flows with active learning}, in: \bibinfo{booktitle}{NeurIPS
  2021 Workshop on Deep Generative Models and Downstream Applications}.
\newblock \URLprefix \url{https://openreview.net/forum?id=LdWEo5mri6}.
\bibitem[{Zhang et~al.(2022a)Zhang, Mu, Zhang, Liu, Zong and
  Li}]{ZHANG2022108234}
\bibinfo{author}{Zhang, X.}, \bibinfo{author}{Mu, J.}, \bibinfo{author}{Zhang,
  X.}, \bibinfo{author}{Liu, H.}, \bibinfo{author}{Zong, L.},
  \bibinfo{author}{Li, Y.}, \bibinfo{year}{2022}a.
\newblock \bibinfo{title}{Deep anomaly detection with self-supervised learning
  and adversarial training}.
\newblock \bibinfo{journal}{Pattern Recognition} \bibinfo{volume}{121},
  \bibinfo{pages}{108234}.
\newblock \URLprefix
  \url{https://www.sciencedirect.com/science/article/pii/S0031320321004155},
  \DOIprefix\doi{https://doi.org/10.1016/j.patcog.2021.108234}.
\bibitem[{Zhang et~al.(2021b)Zhang, Xie, Huang, Zhang and Wang}]{zhang2021self}
\bibinfo{author}{Zhang, X.}, \bibinfo{author}{Xie, W.}, \bibinfo{author}{Huang,
  C.}, \bibinfo{author}{Zhang, Y.}, \bibinfo{author}{Wang, Y.},
  \bibinfo{year}{2021}b.
\newblock \bibinfo{title}{Self-supervised tumor segmentation through layer
  decomposition}.
\newblock \bibinfo{journal}{arXiv preprint arXiv:2109.03230} .
\bibitem[{Zhang et~al.(2022b)Zhang, Zhao, Cai, Feng, Miao, Guan, Tao and
  Cao}]{ICKG}
\bibinfo{author}{Zhang, Z.}, \bibinfo{author}{Zhao, L.}, \bibinfo{author}{Cai,
  D.}, \bibinfo{author}{Feng, S.}, \bibinfo{author}{Miao, J.},
  \bibinfo{author}{Guan, Y.}, \bibinfo{author}{Tao, H.}, \bibinfo{author}{Cao,
  J.}, \bibinfo{year}{2022}b.
\newblock \bibinfo{title}{Time series anomaly detection for smart grids via
  multiple self-supervised tasks learning}, in: \bibinfo{booktitle}{2022 IEEE
  International Conference on Knowledge Graph (ICKG)}, \bibinfo{publisher}{IEEE
  Computer Society}, \bibinfo{address}{Los Alamitos, CA, USA}. pp.
  \bibinfo{pages}{392--397}.
\newblock \URLprefix
  \url{https://doi.ieeecomputersociety.org/10.1109/ICKG55886.2022.00057},
  \DOIprefix\doi{10.1109/ICKG55886.2022.00057}.
\bibitem[{Zhao et~al.(2021)Zhao, Li, He, Ma, Fang, Li and
  Zheng}]{zhao2021anomaly}
\bibinfo{author}{Zhao, H.}, \bibinfo{author}{Li, Y.}, \bibinfo{author}{He, N.},
  \bibinfo{author}{Ma, K.}, \bibinfo{author}{Fang, L.}, \bibinfo{author}{Li,
  H.}, \bibinfo{author}{Zheng, Y.}, \bibinfo{year}{2021}.
\newblock \bibinfo{title}{Anomaly detection for medical images using
  self-supervised and translation-consistent features}.
\newblock \bibinfo{journal}{IEEE Transactions on Medical Imaging}
  \bibinfo{volume}{40}, \bibinfo{pages}{3641--3651}.
\bibitem[{Zheng et~al.(2021)Zheng, Jin, Liu, Chi, Phan and
  Chen}]{zheng2021generative}
\bibinfo{author}{Zheng, Y.}, \bibinfo{author}{Jin, M.}, \bibinfo{author}{Liu,
  Y.}, \bibinfo{author}{Chi, L.}, \bibinfo{author}{Phan, K.T.},
  \bibinfo{author}{Chen, Y.P.P.}, \bibinfo{year}{2021}.
\newblock \bibinfo{title}{Generative and contrastive self-supervised learning
  for graph anomaly detection}.
\newblock \bibinfo{journal}{IEEE Transactions on Knowledge and Data
  Engineering} .
\bibitem[{Zheng et~al.(2022)Zheng, Jin, Liu, Chi, Phan, Pan and
  Chen}]{zheng2022unsupervised}
\bibinfo{author}{Zheng, Y.}, \bibinfo{author}{Jin, M.}, \bibinfo{author}{Liu,
  Y.}, \bibinfo{author}{Chi, L.}, \bibinfo{author}{Phan, K.T.},
  \bibinfo{author}{Pan, S.}, \bibinfo{author}{Chen, Y.P.P.},
  \bibinfo{year}{2022}.
\newblock \bibinfo{title}{From unsupervised to few-shot graph anomaly
  detection: A multi-scale contrastive learning approach}.
\newblock \bibinfo{journal}{arXiv preprint arXiv:2202.05525} .

\end{thebibliography}





\end{document}